\documentclass{article}
\usepackage[margin=1in]{geometry}
\usepackage{titlesec}
\titleformat*{\section}{\large\bfseries}
\titleformat*{\subsection}{\bfseries}
\titleformat*{\subsubsection}{\itshape}
\usepackage[title]{appendix}

\usepackage{authblk}

\usepackage{amsmath, amssymb, siunitx}
\usepackage{bm, bbm}
\DeclareMathAlphabet{\mathbbm}{U}{bbold}{m}{n}

\usepackage{graphicx}
\usepackage{float}
\usepackage{booktabs}
\usepackage[font=small, skip=4pt]{caption}
\usepackage[skip=4pt]{subcaption}
\usepackage[table]{xcolor}

\usepackage{enumitem}
\setlist[itemize]{nosep}

\usepackage[colorlinks, allcolors=blue, pdfstartview=]{hyperref}
\usepackage[nameinlink, capitalize]{cleveref}
\usepackage{natbib}
\title{Deep neural networks for choice analysis: Enhancing behavioral regularity with gradient regularization}
\author[a]{Siqi Feng}
\author[b]{Rui Yao}
\author[c]{Stephane Hess}
\author[d]{Ricardo A. Daziano}
\author[e]{Timothy Brathwaite}
\author[e]{Joan Walker}
\author[f]{Shenhao Wang\thanks{Corresponding author. E-mail: \href{mailto:shenhaowang@ufl.edu}{shenhaowang@ufl.edu}.}}

\affil[a]{Department of Systems Engineering, Cornell University}
\affil[b]{School of Architecture, Civil and Environmental Engineering, École Polytechnique Fédérale de Lausanne}
\affil[c]{Institute for Transport Studies and Choice Modelling Centre, University of Leeds}
\affil[d]{School of Civil and Environmental Engineering, Cornell University}
\affil[e]{Department of Civil and Environmental Engineering, University of California, Berkeley}
\affil[f]{Department of Urban and Regional Planning, University of Florida}
\date{}

\begin{document}
	\maketitle
	\vspace*{-2em}
	
	\begin{abstract}
		\noindent Deep neural networks (DNNs) have been increasingly applied in travel demand modeling because of their automatic feature learning, high predictive performance, and economic interpretability. Nevertheless, DNNs frequently present behaviorally irregular patterns, significantly limiting their practical potentials and theoretical validity in travel behavior modeling. This study proposes strong and weak behavioral regularities as novel metrics to evaluate the monotonicity of individual demand functions (known as the ``law of demand''), and further designs a constrained optimization framework with six gradient regularizers to enhance DNNs' behavioral regularity. The empirical benefits of this framework are illustrated by applying these regularizers to travel survey data from Chicago and London, which enables us to examine the trade-off between predictive power and behavioral regularity for large versus small sample scenarios and in-domain versus out-of-domain generalizations. The results demonstrate that, unlike models with strong behavioral foundations such as the multinomial logit, the benchmark DNNs cannot guarantee behavioral regularity. However, after applying gradient regularization, we increase DNNs' behavioral regularity by around 6 percentage points while retaining their relatively high predictive power. In the small sample scenario, gradient regularization is more effective than in the large sample scenario, simultaneously improving behavioral regularity by about 20 percentage points and log-likelihood by around 1.7\%. Compared with the in-domain generalization of DNNs, gradient regularization works more effectively in out-of-domain generalization: it drastically improves the behavioral regularity of poorly performing benchmark DNNs by around 65 percentage points, highlighting the criticality of behavioral regularization for improving model transferability and applications in forecasting. Moreover, the proposed optimization framework is applicable to other neural network--based choice models such as TasteNets. Future studies could use behavioral regularity as a metric along with log-likelihood, prediction accuracy, and $F_1$ score when evaluating travel demand models, and investigate other methods to further enhance behavioral regularity when adopting complex machine learning models.
	\end{abstract}
	
	
	\clearpage
	\section{Introduction}
	
	Deep neural networks (DNNs) have revolutionized fields such as computer vision and natural language processing, which in turn support technologies such as self-driving cars and large language models \citep{van2023chatgpt, lecun2015deep}. DNNs have also been applied in economics \citep{zheng2023deep} to interpret and predict individual choice behavior \citep{wang2020deeparch, wang2020deep}. It is in this latter area that DNNs offer a contrast with the conventionally used discrete choice models (DCMs), which are typically based on random utility maximization and the assumption that travelers choose the alternative with the highest random utility in the choice set \citep{ben1985discrete}. One drawback of this traditional modeling paradigm is the time-consuming trial-and-error process for an ``optimal'' specification of the model, in particular the utility function that represents economic preferences \citep{van2022choice}. Additionally, the decisions made in this process are often subjective. By contrast, DNNs are capable of automatic feature learning, i.e., the specification of a DNN-based choice model is automatically learned from the input data, which avoids the specification search process and reduces the level of subjectivity. The high prediction accuracy of DNNs is a result of their complex model structure, which can capture intricate behavioral relationships and provide new insights beyond those of conventional DCMs.
	
	Traditional choice modelers often see DNNs as ``black-box'' models, although DNNs actually contain complete economic information for choice analysis \citep{wang2020deep}. However, existing DNNs often exhibit behaviorally irregular patterns because the demand functions in DNNs are not guaranteed to decrease monotonically with generalized costs. The ``law of demand'' in economics indicates an inverse relationship between generalized costs and the aggregate demand. While DCMs such as random utility models (RUMs) do not impose specific prior directionality, the specification search conducted by an analyst would prevent models that lead to counter-intuitive results. However, non-monotonic patterns have been detected empirically in DNNs' predictions, even with model ensembles \citep{wang2020deeparch, wang2020deep, xia2023random}, as the analyst has less control. This fact might be a drawback of the nonlinear structure of DNNs, allowing them to flexibly fit the data but making it challenging to constrain the gradient's direction with limited data samples. The issue often deteriorates in out-of-domain applications, i.e., applying a trained DNN to a test set with unseen distributions \citep{quinonero2008dataset}. In fact, the out-of-domain generalizability of DNNs has attracted rising interests in several computer science subfields, including domain adaption \citep{wang2018deep} and transfer learning \citep{pan2009survey}.
	
	To improve the monotonicity of DNNs, we propose to regularize the loss function in training, which has been shown in computer science to enhance the robustness of DNNs \citep{lyu2015unified, ross2018improving}. Nevertheless, this approach has rarely been considered in previous DNN-based choice models, or used for enhancing behavioral regularity \citep[cf.][]{wang2020deep, zheng2021equality}. In this paper, we address the issue of behavioral irregularity by first defining strong and weak behavioral regularity metrics based on the monotonicity of demand functions, and further designing and implementing a constrained optimization framework that regularizes the input gradient in order to explicitly constrain the gradient's direction. We then design experiments to examine the performance of behaviorally regularized DNNs in terms of behavioral regularity and predictive performance, differentiating between in-domain and out-of-domain generalizations. We also consider the effect of sample size, which is crucial in practice because large samples are costly for travel surveys. Based on two travel survey datasets from Chicago and London, our experiments compare across the standard DNN and TasteNet \citep{han2022neural} architectures using five evaluation metrics, including log-likelihood, prediction accuracy, $F_1$ score, strong behavioral regularity, and weak behavioral regularity. The multinomial logit (MNL) is chosen as a benchmark model due to its concise expression and high behavioral regularity. For TasteNets, we also compare the performance of our gradient regularizers (as soft constraints) and the hard constraints applied in \citet{han2022neural}. The results show that by using appropriate gradient regularization, both DNNs and TasteNets can achieve high regularity without sacrificing their predictive power, which makes these models competitive in real-world applications and demonstrates the generality of our gradient regularization framework.
	
	The rest of this paper is organized as follows. \cref{2} briefly reviews the literature about the behavioral irregularity issue of DNNs with possible solutions. \cref{3} introduces the theory, formulates the problem, and develops a solution framework based on gradient regularization. \cref{4} sets up the mode choice experiment, while \cref{5} illustrates and analyzes the empirical results. Finally, \cref{6} concludes the study and looks ahead to future research. To facilitate future research, we uploaded this work to the following GitHub repository: \url{https://github.com/siqi-feng/DNN-behavioral-regularity}.
	
	\section{Literature review}
	\label{2}
	
	The economic choice behavior of humans generally follows the law of demand in economics, which states the inverse relationship between price and quantity demanded. This indicates a monotonic change in market demand due to the change in consumers' purchasing power, including price and income changes \citep{chiappori1985distribution, hardle1991empirical, hildenbrand1983law, quah2000monotonicity}. The transportation field has also observed the negative influence of travel costs on travel demand \citep{mcfadden1974measurement, souche2010measuring, yao2005study}, based on which demand management policies such as road pricing \citep{may1992road, yang1997traffic} were developed. Although such market rationality is widely recognized, individual choice behavior might be irrational \citep{becker1962irrational, knez1985individual}. \citet{lichtenstein1971reversals} studied preference reversal in decision making, which is a typical counterexample of individual rationality. Studies in bounded rationality theory \citep{simon1957models, di2016boundedly, watling2018stochastic} and prospect theory \citep{kahneman1979prospect, tversky1992advances} also relax the strict monotonicity assumption in demand modeling.
	
	The law of demand is generally followed by the design of random utility models. In the MNL model, for example, an increase in the travel cost of an alternative would be expected to decrease its systematic utility, thus decreasing its choice probability by design. If the initial model includes estimates with counter-intuitive signs, such as positive cost estimates, an analyst can readily spot them and then refine the model specification or deal with data issues. Once all signs are as expected, the monotonic relationship is guaranteed. By contrast, this is not the case in DNNs because of the complex nonlinear model structure, especially when the number of hidden layers increases. For example, \citet{xia2023random} observed non-monotonic demand predictions with increasing generalized costs in a mode choice experiment with DNNs, which suggests the need to investigate the monotonicity of DNNs to improve their behavioral regularity. Although shallow neural networks (NNs) might reduce the risk of non-monotonic behavior of DNNs \citep{alwosheel2019computer, zhao2020prediction}, this might come at the cost of reduced modeling flexibility and universal approximation power. Alternatively, \citet{han2022neural} and \citet{sifringer2020enhancing} proposed to use DNNs only for learning latent representation in the utility function, while resorting to the DCM framework to ensure model monotonicity. For example, TasteNet, the neural-embedded DCM proposed by \citet{han2022neural}, assumes a linear model specification with parameters learned by an NN. In this paper, TasteNets are considered as a reference to the standard DNN architecture. On the other hand, depending on model design, these hybrid DNN models might still produce irregular predictions \citep{wang2021theory, wong2021reslogit}. Moreover, hybrid DNN models are a compromise for regularity since they again require the subjective process of model specification. To fully utilize the capability of DNNs, previous studies have attempted to migrate the non-monotonic issue through model ensemble. However, irregular patterns might still be observed after averaging over multiple trainings \citep{wang2020deeparch, wang2020deep, xia2023random}. One promising direction is to directly integrate domain-specific knowledge into the design and training of DNNs, such as incorporating monotonicity constraints in model training \citep{haj2023incorporating}. Nevertheless, there is no consensus on how to measure or improve the model regularity of DNNs within the choice modeling field. This paper contributes to the development of a behavioral regularity measure and a novel regularization framework.
	
	As discussed in computer science applications, the regularity of DNNs can be improved by employing either hard or soft constraints. The first category enforces monotonicity by model construction, e.g., constraining the positiveness of weights in hidden layers \citep{daniels2010monotone, dugas2009incorporating, sill1997monotonic} through non-negativity constraints \citep{lawson1995solving}, restricting the derivatives to be positive \citep{neumann2013reliable, wehenkel2019unconstrained}, down-weighting samples that violate monotonicity \citep{archer1993application}, and incorporating deep lattice networks for learning monotonic functions \citep{you2017deep}. The second category achieves monotonicity by regularization, i.e., by augmenting a regularization term in the loss function to jointly improve model monotonicity. In the choice modeling domain, recent studies have also incorporated hard constraints on model monotonicity based on the modeler's prior assumptions to improve intepretability, such as in lattice networks \citep{kim2024new} and TasteNets \citep{han2022neural}. Regularization is firmly rooted in constrained optimization, including the Lagrangian method as an example \citep{boyd2004convex}. It has been widely applied as a local method to penalize constraint violations. For example, \citet{sill1996monotonicity} penalized squared deviations in monotonicity for virtual pairs of input variables, while \citet{gupta2019incorporate} proposed a pointwise loss that embeds prior knowledge about monotonicity. Moreover, gradient regularization has also been used to enhance model robustness against adversarial examples \citep{lyu2015unified, ross2018improving}, e.g., penalizing the squared $L_2$ norm of the gradient of the loss with respect to (w.r.t.)\ inputs \citep{drucker1991double, ororbia2017unifying}, penalizing the squared Frobenius norm of the Jacobian matrix of probabilities \citep{sokolic2017robust} and utilities \citep{jakubovitz2018improving} w.r.t.\ inputs. Note that regularizing the gradient norms might be less effective to improve monotonicity than regularizing the gradient's direction like in \citet{haj2023incorporating}. Inspired by the aforementioned regularization methods, we will design our own approaches in the demand modeling context.
	
	\section{Methodology}
	\label{3}
	
	\subsection{DNNs for choice analysis}
	The discrete choice problem is cast as a supervised classification task in DNN-based choice analysis. Assuming there are in total $D$ explanatory variables $\{x_1, \ldots, x_D\}$ for all alternatives, the attribute vector of individual $n$ can be written as $\mathbf{x}_n = [x_{n1},\ldots,x_{nD}]^\top$. Then, a DNN model predicts the probability of $n$ choosing $i$ out of $J$ alternatives, i.e., $P_{ni}: \mathbb{R}^D\to (0,1)$ and $\sum_{i=1}^J P_{ni} = 1$. The observed choice vector $\mathbf{y}_n\in\{0,1\}^J$ of $n$ is used for DNN training, where $y_{ni} = 1$ if alternative $i$ is chosen, and $y_{ni} = 0$ otherwise. Similar to conventional RUMs, DNN models aim to find specifications with high predictive power and behavioral regularity.
	
	In contrast to RUMs with handcrafted specifications, DNNs automatically learn specifications through their unique representation learning capability. Particularly, utility vector $\mathbf{V}_{n} = [V_{n1}, \ldots, V_{nJ}]^\top$ is specified through a series of transformations, termed as layers $\{f_1, \ldots, f_H\}$, where $H$ denote the total number of layers in a DNN. Each layer $f_h$ contains a learnable parameter matrix $W_h$, a bias vector $\mathbf{b}_h$, and an activation function $\varphi(\cdot)$ (e.g., the rectified linear unit, ReLU) to transform $\mathbf{x}_n$. Specifically, each layer transformation can be written as
	\begin{equation}
		f_h(\mathbf{x}_n) = \varphi(W_h \mathbf{x}_n + \mathbf{b}_h)
	\end{equation}
	and the utility vector $\mathbf{V}_n$ is computed in a composite form:
	\begin{equation}
		\mathbf{V}_n = \left(f_H\circ f_{H-1}\circ\cdots\circ f_2\circ f_1\right)(\mathbf{x}_n)
		\label{eq:DNN_utility_function}
	\end{equation}
	Finally, a softmax classification layer (i.e., the logistic function) outputs the choice probability of $i$ as
	\begin{equation}
		P_{ni} = \frac{e^{V_{ni}}}{\sum_{j=1}^J e^{V_{nj}}}
		\label{eq:softmax}
	\end{equation}
	
\begin{figure}[!htb]
		\centering
		\includegraphics[width=.55\linewidth]{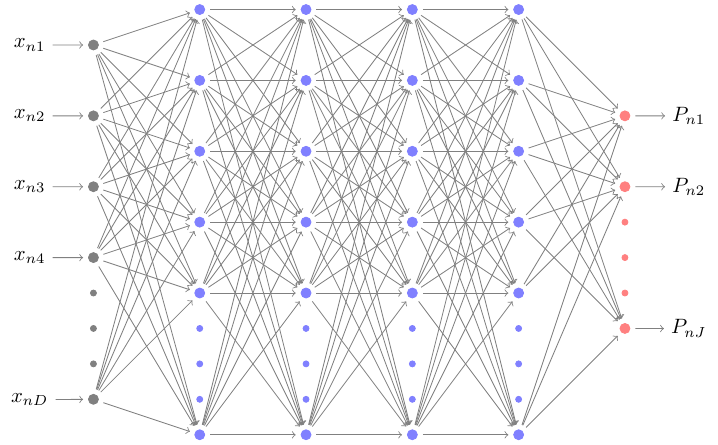}
		\caption{A feedforward DNN structure with four hidden layers and one classification layer.}
		\label{fig:dnn}
	\end{figure}

	The DNN structure generalizes the classical linear MNL model. If an NN is specified with a single output layer (i.e., without any hidden layer) and an identity activation function, the utility function in \cref{eq:DNN_utility_function} would collapse to
	\begin{equation}
		\mathbf{V}_n = f(\mathbf{x}_n) = W \mathbf{x}_n + \mathbf{b}
		\label{eq:mnl}
	\end{equation}
	where $W\in \mathbb{R}^{J\times D}$ can be interpreted as parameters and $\mathbf{b}\in \mathbb{R}^{J\times 1}$ as alternative-specific constants. Although closely related to the MNL model, DNNs allow for flexible model specifications through multi-layer nonlinear transformations. We illustrate in \cref{fig:dnn} a feedforward DNN structure with four hidden layers and one classification layer for a choice modeling problem with $D$ attributes for $J$ alternatives.
	
	\subsection{Behavioral regularity metrics}
	In this study, we propose a novel metric to evaluate behavioral regularity, which measures the monotonicity of aggregate choice probability functions. The proposed metric essentially measures the monotonicity consistency between the model and prior knowledge on the correct signs of parameter estimates, commonly used in the subjective process of selecting plausible specifications for RUMs. The behavioral regularity metric of alternative $i$ w.r.t.\ a cost variable $x_d$ (with value $x_{nd}$ for individual $n$) is defined as
	\begin{equation}
		B_{id} = \iint \mathbbm{1}\left\{\frac{\partial P_i(\mathbf{x}_z)}{\partial x_d} < \varepsilon \right\} \rho(x_d, z) dx_d dz
		\label{eq:theoretical_reg}
	\end{equation}
	where $z$ represents a sociodemographic factor of a population group; $P_i(\mathbf{x}_z)$ maps the individual's attributes $\mathbf{x}_z$, including both individual-specific sociodemographics mapped from factor $z$, and alternative-specific cost variables, to the individual's probability of choosing $i$; $\mathbbm{1}\{\cdot\}$ is an indicator function that equals 1 if ${\partial P_i(\mathbf{x}_z)}/{\partial x_d} < \varepsilon$, and 0 otherwise; $\rho(x_d, z)$ denotes the joint density of attributes $x_d$ and the population factor $z$; and $\varepsilon$ represents the modeler's prior assumptions on the monotonicity of $P_i$ w.r.t.\ $x_d$:
	\begin{itemize}
		\item $\varepsilon = 0$, termed as \emph{strong} regularity, requires $P_i$ to be strictly decreasing w.r.t.\ $x_d$. The formulation assumes that all individuals across population groups reduce their choice probability of alternative $i$ with larger costs in $x_d$.
		\item $\varepsilon > 0$, termed as \emph{weak} regularity, relaxes the strict monotonicity assumption and allows $P_i$ to be non-decreasing w.r.t.\ $x_d$. The formulation assumes that some population groups do not respond (with zero derivatives) to $x_d$, implying that the behavioral regularity constraint becomes weaker.
	\end{itemize}
	As an illustration, a classical linear MNL model with a negative parameter w.r.t.\ $x_d$ yields $B_{id} = 1$, which implies that all individual behaviors are consistent with the demand monotonicity assumption and well captured by the model. We also note that \cref{eq:theoretical_reg} can be interpreted as the cumulative distribution of behavioral regularity with $\varepsilon$ over the whole population and the domain of $x_d$.
	
	The population-level behavioral regularity metric in \cref{eq:theoretical_reg} can be approximated by the mean behavioral regularity across individuals:
	\begin{equation}
		B_{id} \approx \frac{1}{N} \sum_{n=1}^N \mathbbm{1}\left\{\frac{\Delta P_{ni}}{\Delta x_{nd}} < \varepsilon \right\}
		\label{eq:emperical_reg}
	\end{equation}
	where $N$ is the sample size, and the partial derivative is computed with finite differences. The proposed empirical regularity metric in \cref{eq:emperical_reg} is the sample analog of the exact metric in \cref{eq:theoretical_reg}. By the Glivenko--Cantelli theorem, \cref{eq:emperical_reg} converges in probability to \cref{eq:theoretical_reg}, the cumulative distribution function for the mean population behavioral regularity with $\varepsilon$.
	
	Our behavioral regularity metrics can be extended to incorporate taste heterogeneity across population groups and even individuals within each group by distinguishing $\varepsilon$ w.r.t.\ different groups, i.e., $\varepsilon$ can be further specified as $\varepsilon_z$ to reflect the group-specific thresholds. Meanwhile, our metrics only require aggregate regularity rather than individual regularity, inspired by classical economics discussions: market rationality is a fundamental law, whereas individual behaviors might present more diverse and irrational patterns \citep{becker1962irrational}.
	
	\subsection{Achieving behavioral regularity by constrained optimization}
	
	\subsubsection{Unconstrained likelihood maximization}
	DNN-based choice models can be estimated using the likelihood maximization framework. Given a set of hyperparameters and the softmax activation function, likelihood maximization and cross-entropy minimization are mathematically equivalent, i.e., an unconstrained DNN learns parameters $W$ through minimizing the cross-entropy $L$:
	\begin{equation}
		\min_W L(W) = \min_W \frac{1}{N} \sum_{n=1}^N \sum_{i=1}^J -y_{ni}\log P_i(\mathbf{x}_n; W)
		\label{eq:unconstrained}
	\end{equation}
	
	The unconstrained formulation in \cref{eq:unconstrained} is sufficient for the estimation of conventional RUMs, since they often satisfy convexity conditions because of linear-in-parameters specifications. In a linear MNL model, for example, choice probability $P_{ni}$ increases monotonically with utility $V_{ni}$ according to \cref{eq:softmax}. The linear specification in \cref{eq:mnl} induces monotonicity of utility w.r.t.\ cost variables. Therefore, if individuals indeed perceive higher utility with lower costs, the optimization would result in negative parameter estimates, and the behavioral monotonicity is clearly satisfied by evoking the chain rule. The multi-layer nonlinear transformations in DNNs allow for approximation of arbitrary functions, but these complex transformations might lead to non-monotonic choice probability functions, especially when the network is deep. In this case, unconstrained likelihood maximization can no longer guarantee a behaviorally regular model.
	
	\subsubsection{Constrained likelihood maximization}
	To address the irregularity issue of DNNs, we introduce a set of behavioral regularity constraints into the optimization problem, which yields
	\begin{align}
		\min_W \quad & L(W) \\
		\text{s.t.} \quad & R(\mathbf{x}_n; W) \le 0, \quad n = 1, \ldots, N
		\label{eq:constraints}
	\end{align}
	where $R$ constrains the attributes $\mathbf{x}_n$ of an individual $n$, defined as $R:\mathbb{R}^{J\times D} \to \mathbb{R}$, where dimension $J$ is the number of alternatives and $D$ is the number of attributes. Therefore, the behavioral regularity constraints are imposed at the individual level to enhance the aggregate regularity in $B_{id}$. Specific behavioral regularity constraints will be designed in the next subsection.
	
	Training DNNs with constraints is challenging. We tackle this problem by treating the \emph{hard} constraints in \cref{eq:constraints} as \emph{soft} constraints, motivated by the Lagrangian relaxation method. Given a hyperparameter $\lambda$, we consider the following optimization problem:
	\begin{equation}
		\min_W L(W) + \lambda \sum_{n=1}^N R(\mathbf{x}_n; W)
		\label{eq:loss_sum_reg}
	\end{equation}
	where $\lambda$ controls the strength of the behavioral regularity constraint and can be interpreted as a Lagrangian multiplier for constrained optimization. We note that the relaxation formulation in \cref{eq:loss_sum_reg} is similar to the regularization methods that are commonly applied in machine learning for model sparsity, while our motivation is to improve the behavioral regularity of the DNN choice models.
	
	Compared with the hard constraint formulation, soft regularization can flexibly accommodate the various degrees of validity in our behavioral regularity assumptions. Similar to the motivation for the weak regularity metric, our approach allows each individual $n$ to somewhat violate the preset constraint $R(\mathbf{x}_n)$, thus accommodating the potentially irregular behavior of certain individuals. As a result, hyperparameter $\lambda$ provides insight into the consistency between behavioral regularity assumptions and the actual behavior of studied individuals. If a larger $\lambda$ is required to achieve higher predictive performance, it might imply that the actual behavior is inconsistent with prior assumptions, thus providing extra insight into the validity of behavioral regularity constraints. Although the gradient regularization method does not guarantee strong behavioral regularity, it provides insights into model misspecification or the inconsistency between modeling assumptions and data.
	
	\subsection{Gradient regularization}
	We design gradient regularizers to improve the behavioral regularity of DNN-based choice models. Specifically, we constrain the demand feedback on generalized costs by the gradient's direction (i.e., signs of the parameter estimates) and magnitude.
	
	For individual $n$, the Jacobian matrix (gradient) of demand vector $\mathbf{P} = [P_1, \ldots, P_J]^\top$ w.r.t.\ explanatory variables $\{x_1, \ldots, x_D\}$ can be written as
	\begin{equation}
		\nabla\mathbf{P}(\mathbf{x}_n) =
		\begin{bmatrix}
			\frac{\partial P_1}{\partial x_1}(\mathbf{x}_n) & \cdots & \frac{\partial P_1}{\partial x_D}(\mathbf{x}_n) \\
			\vdots & \ddots & \vdots \\
			\frac{\partial P_J}{\partial x_1}(\mathbf{x}_n) & \cdots & \frac{\partial P_J}{\partial x_D}(\mathbf{x}_n)
		\end{bmatrix}
	\end{equation}
	which includes three types of partial derivatives:
	\begin{itemize}
		\item direct derivatives, e.g., the probability of driving w.r.t.\ driving cost;
		\item cross derivatives, e.g., the probability of driving w.r.t.\ public transit time; and
		\item sociodemographic derivatives, e.g., the probability of driving w.r.t.\ the age of traveler.
	\end{itemize}
	
	The constrained likelihood maximization framework in \cref{eq:loss_sum_reg} allows us to impose gradient constraints to the three types of partial derivatives in the Jacobian matrix. Since behavioral regularity is reflected by the gradient's direction, we introduce a mask matrix for individual $n$:
	\begin{equation}
		\Psi(\mathbf{x}_n) =
		\begin{bmatrix}
			\mathbbm{1}\left\{\frac{\partial P_1}{\partial x_1}(\mathbf{x}_n) \notin \mathbb{S}_{11}\right\} & \cdots & \mathbbm{1}\left\{\frac{\partial P_1}{\partial x_D}(\mathbf{x}_n) \notin \mathbb{S}_{1D}\right\} \\
			\vdots & \ddots & \vdots \\
			\mathbbm{1}\left\{\frac{\partial P_J}{\partial x_1}(\mathbf{x}_n) \notin \mathbb{S}_{J1}\right\} & \cdots & \mathbbm{1}\left\{\frac{\partial P_J}{\partial x_D}(\mathbf{x}_n) \notin \mathbb{S}_{JD}\right\}
		\end{bmatrix}
	\end{equation}
	where $\mathbbm{1}\{\cdot\}$ is an indicator function that equals 1 if $\partial P_i(\mathbf{x}_n)/\partial x_d \notin \mathbb{S}_{id}$, and 0 otherwise; and set $\mathbb{S}_{id}$ defines the expected sign of the partial derivative. Combining the mask matrix $\Psi(\mathbf{x}_n)$ and the Jacobian matrix $\nabla\mathbf{P}(\mathbf{x}_n)$, we define the \emph{sum-based} gradient regularization using the Frobenius inner product:\footnote{For real matrices, we have $\langle A, B\rangle_F = \sum_{i,j} A_{ij} B_{ij}$.}
	\begin{equation}
		R_\sigma(\mathbf{x}_n) = \langle \Psi(\mathbf{x}_n), \nabla\mathbf{P}(\mathbf{x}_n)\rangle_F
	\end{equation}
	
	This sum-based approach flexibly accommodates different prior assumptions on the signs of the derivatives. Set $\mathbb{S}_{id}$ can take negative values ($\mathbb{S}_{id} = \mathbb{R}^-$), positive values ($\mathbb{S}_{id} = \mathbb{R}^+$), or any real values ($\mathbb{S}_{id} = \mathbb{R}$), depending on the prior assumption on attribute $x_d$'s effect on demand $P_i(\mathbf{x}_n)$. For example, by imposing $\mathbb{S} = \mathbb{R}^-$ on the direct derivatives, they are expected to be negative and penalized if non-negative. On the other hand, when there is no prior assumption regarding a derivative, we allow all possible signs by taking $\mathbb{S} = \mathbb{R}$. Despite such flexibility, we only impose negative constraints on the direct derivatives throughout our empirical experiments, which is the least controversial among all possibilities.
	
	As an alternative to the sum-based approach, we could also regularize the gradient's magnitude, implying that demands are not expected to change drastically with small cost perturbations. Using the same notations, we define the \emph{norm-based} regularization as
	\begin{equation}
		R_\nu(\mathbf{x}_n) = \left\|\nabla\mathbf{P}(\mathbf{x}_n)\right\|_F^2 = \langle \nabla\mathbf{P}(\mathbf{x}_n), \nabla\mathbf{P}(\mathbf{x}_n)\rangle_F
	\end{equation}
	
	This norm-based approach is relatively common in the computer science literature \citep{drucker1991double, jakubovitz2018improving}, thus serving as a benchmark regularization method for our empirical experiments. Although smoothness is a relatively common assumption from a pure mathematical perspective, it is not founded on strong behavioral regularity beliefs due to possible threshold effects of pricing.
	
	The regularization terms proposed above are termed as probability gradient regularizers (PGRs) because they exploit the analytical relationship between demand monotonicity and probability gradients $\nabla\mathbf{P}(\mathbf{x}_n)$. Due to the computational chain among utilities, choice probabilities, and log-likelihoods, it is also possible to replace the probability gradients by utility and log-likelihood gradients. The two alternative regularizers are defined as
	\begin{itemize}
		\item Utility gradient regularizers (UGRs): According to the softmax function in \cref{eq:softmax}, choice probability $P_i(\mathbf{x}_n)$ increases monotonically with utility $V_i(\mathbf{x}_n)$. Consequently, demand monotonicity can be retained by regularizing the utility monotonicity w.r.t.\ generalized costs and evoking the chain rule. Therefore, we construct UGRs by replacing $\mathbf{P}(\mathbf{x}_n)$ with $\mathbf{V}(\mathbf{x}_n)$ in derivation.
		\item Log-likelihood gradient regularizers (LGRs): We define the individual- and alternative-specific log-likelihood as $l_i(\mathbf{x}_n) = -y_{ni}\log P_i(\mathbf{x}_n)$. Since logarithmic transformation is monotonic, demand monotonicity can be retained by regularizing the log-likelihood monotonicity w.r.t.\ generalized costs. Thus we construct LGRs by replacing $\mathbf{P}(\mathbf{x}_n)$ with $\mathbf{l}(\mathbf{x}_n)$ in derivation.
	\end{itemize}
	
	In brief, by combining sum- and norm-based regularization with probability, utility, and log-likelihood gradients, we have designed six gradient regularizers. They are hereafter referred to as sum-PGR, sum-UGR, sum-LGR, norm-PGR, norm-UGR, and norm-LGR, all of which will be tested thoroughly in our empirical experiments.
	
	When hidden layers are not present, norm-UGR reduces to $L_2$ regularization  because the gradient of the utility is simply the DNN parameters $W$. On the other hand, the proposed gradient regularizers differ from commonly used sparsity regularizers (e.g., $L_1$ and $L_2$ norms) if any hidden layer is present. This is because the derivatives of our gradient regularizers become nonseparable and nonlinear in DNN parameters $W$, as opposed to $L_2$ norms whose derivatives would be separable and linear in $W$. This difference also implies that our gradient regularizers would still be effective under various averaging scheme employed in algorithms like Adam, making the proposed approaches more robust to the choice of algorithms.
	
	\section{Setup of experiments}
	\label{4}
	
	\subsection{Datasets}
	Our experiments use two datasets from Chicago and London, with distinguished car- and transit-dependent travel patterns, to examine the gradient regularizers and the behavioral regularity metrics. The first dataset was collected by the Chicago Metropolitan Agency for Planning (CMAP) in the My Daily Travel Survey in 2018--2019.\footnote{See \url{https://www.cmap.illinois.gov/data/transportation/travel-survey}.} After preliminary cleaning, the full dataset retains 26,099 trips made by four travel modes: driving, walking, train (public transit), and cycling. Driving accounts for 70\% of these trips, whereas the proportion of cycling trips is negligible. Hence the walking and cycling modes were merged into a single active mode to create a more balanced dataset. Based on the spatial information of each trip in terms of origin and destination, we compiled level of service data by utilizing Google Directions API to collect the travel time of each mode, where active times were calculated by averaging walking and cycling times. Train costs were provided by the dataset, while driving costs were computed by summing the money paid to toll plazas en route and parking lots. The $K$-nearest neighbors algorithm was applied to impute the missing data, especially for driving and train costs. This study uses in total ten explanatory variables: two continuous alternative attributes (time of each mode, and costs by car and train), three discrete sociodemographics (age, household size, and number of cars in the household), and five indicator variables of sociodemographics (higher education, males, one-person households, one-car households, and high-income households). The basic statistics of the full CMAP dataset are summarized in \cref{tab:stat_full}.
	
	The second dataset is the London Travel Demand Survey \citep[LTDS;][]{hillel2018recreating, wang2020deep}, which has 81,086 trips made by four travel modes: walking, cycling, public transit, and driving. The proportion of cycling trips is again negligible, hence an active mode is created with travel time defined as the average of walking and cycling times. Meanwhile, the public transit time is defined as the sum of access time, in-vehicle time, and transfer time. We use in total nine explanatory variables: two continuous alternative attributes (time of each mode, and costs by car and public transit), two discrete sociodemographics (number of cars in the household, and number of public transit transfers), and five indicator variables of sociodemographics (youths, seniors, males, driving license, and one-car households). The basic statistics of the full LTDS dataset are summarized in \cref{tab:stat_full_london}.
	
	The two datasets were further reprocessed to create three smaller datasets for each to examine the effects of large versus small sample sizes, and in-domain versus out-of-domain generalizations. The first dataset, named as \emph{10K-Random}, incorporates 10,000 trips with 70\% randomly sampled for training, 10\% for validation, and 20\% for testing. This dataset is considered as the benchmark for the ideal modeling scenario with a sufficiently large sample size. The second dataset, named as \emph{1K-Random}, contains 1,000 trips with 80\% randomly sampled for training and 20\% for validation. To avoid random variation due to small sample sizes, another 500 trips were randomly sampled for testing. The 1K-Random dataset aims to simulate a classical choice modeling scenario with only limited samples available. By comparing the results between these two datasets, we could evaluate how predictive performance and behavioral regularity vary with sample sizes. The third dataset, named as \emph{10K-Sorted}, employs a different strategy for data splitting, where the 10,000 trips are sorted by driving cost, while the upper 20\% were used for testing, and the lower 80\% were further randomly sampled for training (70\%) and validation (10\%). As shown in \cref{sec:appendix_a.2}, the distributions of variables are quite different between the training and test sets, with significantly higher mean and standard deviations in the test set. This train--test split scheme simulates the testing carried out on more expensive trips. Such out-of-domain generalizability is not only of theoretical interests, but also highly relevant in practice because it investigates model transferability, i.e., how the models perform in a target context distinct from their source context. One rationale of the sampling scheme for 10K-Sorted datasets is that it resembles cross-city policy learning. Local governments regularly seek to implement transportation policies (e.g., congestion charging) that originate in other cities. The out-of-domain generalization can simulate the data and modeling challenges in such cross-city policy learning.
	
	\subsection{Experimental design}
	Our experiments use the training set for model training, the validation set for hyperparameter searching, and the test set for model evaluation and comparison. Using the training set of 10K-Random (CMAP), we show in \cref{sec:appendix_b} that both Adam \citep{kingma2014adam} and AdamW \citep{loshchilov2017decoupled} are empirically suitable for our experiments, while standard stochastic gradient descent (SGD) converges much slower and results in unreasonable individual demand functions. The difference between Adam and AdamW lies in the implementation of weight decay as another type of regularization, which is not included in the experiments, thus teasing out the effects of gradient regularization from other factors. The training set was divided into 10 batches for model training, with a learning rate of \num{e-3}. To ensure convergence, we train each model until the validation loss in consecutive iterations reaches an optimum.
	
	Using the validation sets, we selected the optimal regularization strength $\lambda$ for each of the gradient regularizers by overall model performance, thus balancing predictive power and behavioral regularity. The DNN architecture was chosen with four hidden layers and 100 neurons per layer after random search. As the range of $\lambda$ depends on the dataset, model class, and gradient regularizer, we only show an example of the hyperparameter space in \cref{tab:hyperparameter_space}, where we took $\lambda$ values from \num{e-4} to 100 in a logarithmic scale to fully demonstrate the effects of $\lambda$ and select the optimum. It is expected that the DNNs with extremely small $\lambda$'s approximate the benchmark DNN, while those with large $\lambda$'s sacrifice predictive power for behavioral regularity.
	
	\begin{table}[!htb]
		\centering\small
		\caption{Hyperparameter space for DNNs with sum-XGR (10K-Random, CMAP).}
		\label{tab:hyperparameter_space}
		\begin{tabular}{ll}
			\toprule
			Hyperparameter & Values \\
			\midrule
			Depth (number of hidden layers) & 3, \textbf{4}, 5, 6 \\
			Width (number of neurons per layer) & 50, \textbf{100}, 150 \\
			Regularization strength & \num{e-4}, \num{e-3}, \textbf{0.01}, 0.1, 1, 10, 100 \\
			\bottomrule
		\end{tabular}
	\end{table}
	
	Lastly, all models are evaluated by their performance in the test sets. Particularly, to mitigate model randomness, we analyze the ensemble performance by averaging the results of 10 model replications. \cref{5} will focus on comparing model performance using the test sets, while the training and validation performance are reported in \cref{sec:appendix_c}. The models are evaluated by five metrics: log-likelihood, prediction accuracy, $F_1$ score, strong behavioral regularity, and weak behavioral regularity. The first three metrics focus on predictive power, measuring how well a model fits the observed outputs. Among the three metrics, log-likelihood is the most important one because of its probabilistic nature, its wide adoption in the field of discrete choice analysis, and its solid theoretical foundation for model convergence. Prediction accuracy and $F_1$ score are also adopted because the former is the most common metric in machine learning and the latter tackles the potential evaluation problem in imbalanced datasets. In addition to these predictive metrics, we also evaluate the models using strong and weak behavioral regularities based on \cref{eq:emperical_reg}. To empirically compute the two behavioral metrics, we set parameter $\varepsilon$ to a small negative number for strong regularity and a small positive number for weak regularity. Despite the theoretical threshold $\varepsilon = 0$ for strong regularity, we set the value slightly lower than zero to enhance numerical stability and distinguish between the two metrics.\footnote{As a result, the empirical strong regularity of MNL presented in \cref{5} could be slightly lower than 100\%.} We conducted a sensitivity test in \cref{sec:appendix_e} to discuss the impact of $\varepsilon$ on the metric values. \cref{5} will fully demonstrate the trade-off between predictive and behavioral metrics by adjusting the strength of regularization.
	
	\subsection{Models}
	Three types of models are compared, including MNL models from the DCM family, standard DNNs, and TasteNets, a hybrid model combining features of DCMs and NNs \citep{han2022neural}. The proposed gradient regularizers are implemented on both DNNs and TasteNets. The MNL models are estimated with \texttt{PyLogit}, and DNNs and TasteNets are implemented with \texttt{PyTorch}. The following linear-in-parameters utility function is specified for the MNL models:
	\begin{align}
		\text{Driving: }V_{n1} &= \beta_{t1} t_{n1} + \beta_{c1} c_{n1} \\
		\text{Public transit: }V_{n2} &= \alpha_2 + \bm{\gamma}_2 \mathbf{z}_n + \beta_{t2} t_{n2} + \beta_{c2} c_{n2} \\
		\text{Active mode: }V_{n3} &= \alpha_3 + \bm{\gamma}_3 \mathbf{z}_n + \beta_{t3} t_{n3}
	\end{align}
	where $t_{ni}$ is the travel time of individual $n$ by alternative $i = \{1,2,3\}$, $c_{ni}$ is the travel cost of $n$ by $i = \{1,2\}$, $\mathbf{z}_n$ is a set of variables specific to $n$, and $\mathbf{w} = \{\bm{\alpha}, \bm{\beta}, \bm{\gamma}\}$ is the set of parameters to be estimated. Our experiments focus on evaluating the effectiveness of the proposed gradient regularization in improving DNNs' behavioral regularity and prediction power, while the MNL models are only benchmarks to demonstrate their inherent behavioral regularity.
	
	For TasteNets, we follow the model specification as in \citet{han2022neural}. Specifically, this implementation maps individual characteristics into individual-specific taste parameters for alternative attributes via a feedforward NN with one hidden layer:
	\begin{equation}
		V_{ni} = T(\mathbf{z}_n; W)^\top \mathbf{x}_{ni} = \bm{\beta}_{ni}^\top \mathbf{x}_{ni}
		\label{eq:TasteNet}
	\end{equation}
	where $T$ represents the NN architecture, $W$ is a set of weights, and $\bm{\beta}_{ni}$ is a vector of taste parameters specific to individual $n$ and alternative $i$. \citet{han2022neural} suggested imposing hard constraints on taste parameters to enforce behavioral regularity, given that \cref{eq:TasteNet} is linear in taste parameters. For example, let $b_{ni}$ denote the NN-embedded taste parameter for travel cost, we can enforce the sign constraint by setting $-\mathrm{ReLU}(-b_{ni})$ or $-\exp(-b_{ni})$ to ensure behaviorally regular TasteNets. We also compare the performance of soft and hard constraints on TasteNets in \cref{5}.
	
	\section{Results}
	\label{5}
	
	In this section, we present the results of our empirical work in three stages. \cref{sec:model_performance} compares the behaviorally regularized DNNs and TasteNets with benchmark models, including their counterparts without regularization and MNL models, regarding predictive power and behavioral regularity metrics. Specifically, we design the large sample (10K-Random), small sample (1K-Random), and out-of-domain generalization (10K-Sorted) scenarios. We note that although sharing certain similarity, out-of-sample generalization (i.e., testing on new samples, like the 10K- and 1K-Random datasets) and out-of-domain generalization (i.e., testing on new distributions, like the 10K-Sorted datasets) are two different concepts. In particular, the former assumes that the training and test sets follow the same statistical pattern, whereas the latter refers to unforeseen distribution shift such as cross-city policy transfer \citep{liu2021towards}. \cref{sec:tradeoff} further investigates how the strength of regularization influences the trade-off between predictive power and behavioral regularity in each scenario. Finally, our empirical findings are summarized in \cref{sec:summary_findings}.
	
	\subsection{Enhancing model performance with gradient regularization}
	\label{sec:model_performance}
	
	\subsubsection{Large sample scenario}
	\label{sec:performance_10k_random}
\begin{figure}[!htb]
		\centering
		\begin{subfigure}{.325\linewidth}
			\includegraphics[width=\linewidth]{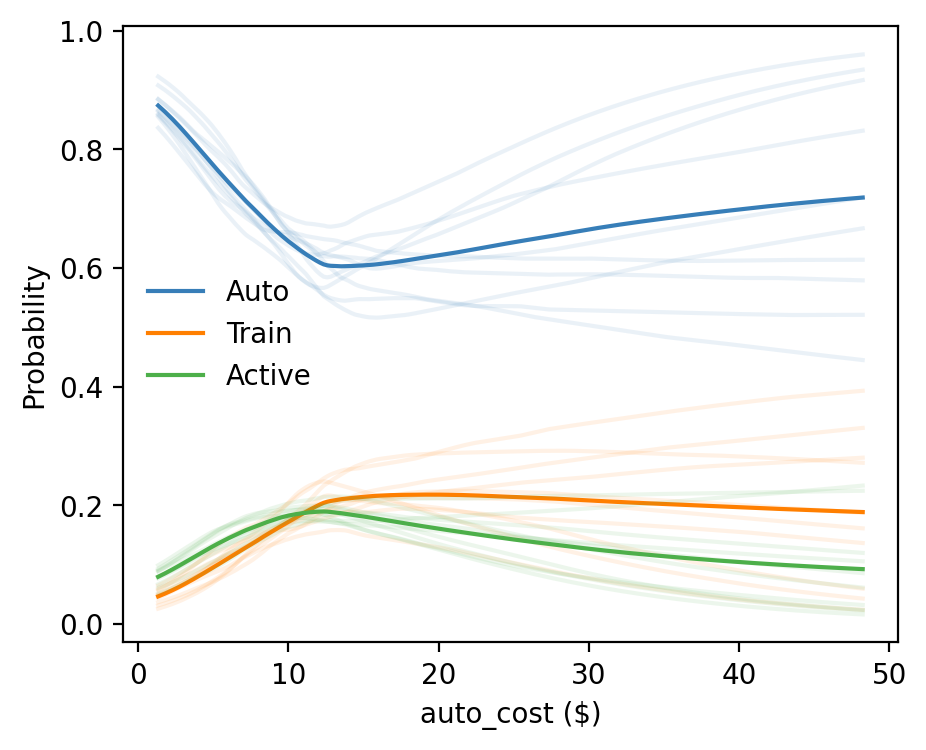}
			\subcaption{DNN}
		\end{subfigure}
		\begin{subfigure}{.325\linewidth}
			\includegraphics[width=\linewidth]{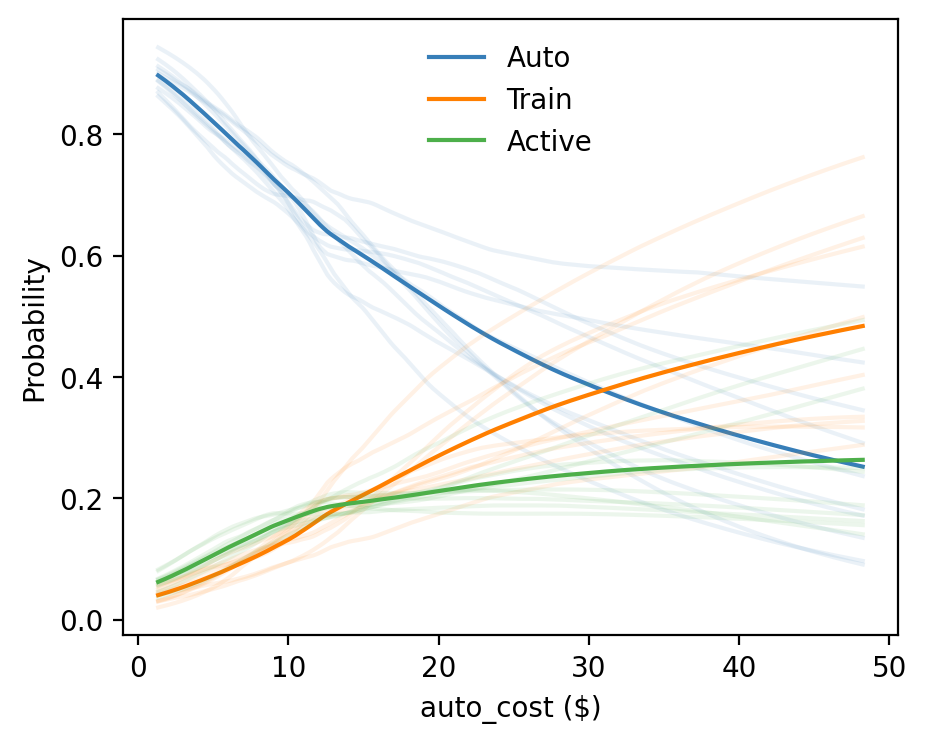}
			\subcaption{DNN, sum-PGR}
		\end{subfigure}
		\begin{subfigure}{.325\linewidth}
			\includegraphics[width=\linewidth]{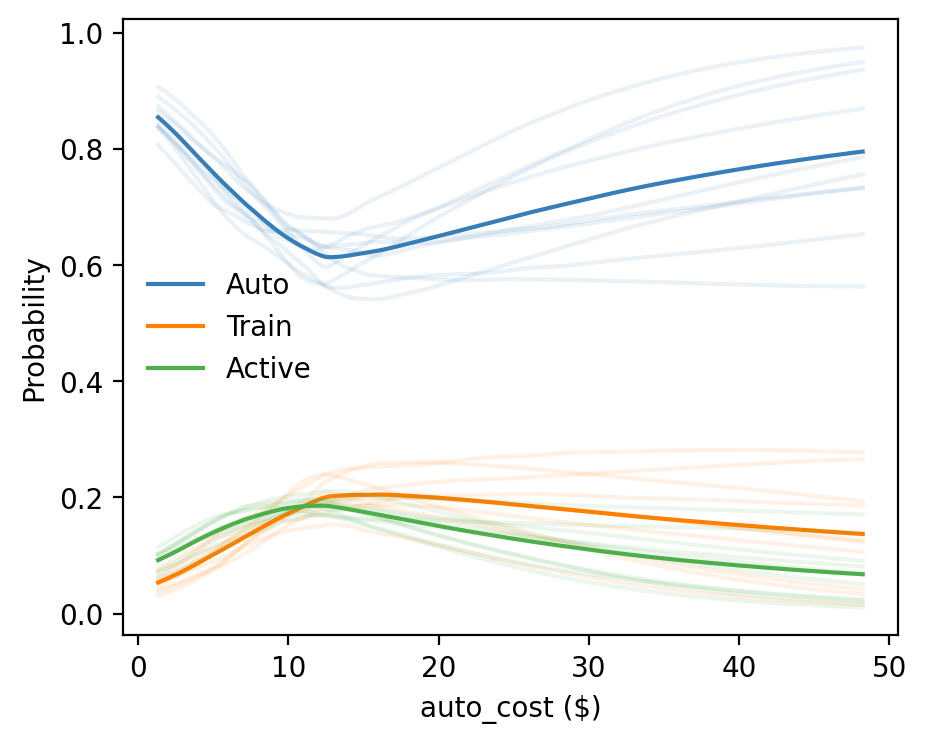}
			\subcaption{DNN, norm-PGR}
		\end{subfigure}
		\par\smallskip
		\begin{subfigure}{.325\linewidth}
			\includegraphics[width=\linewidth]{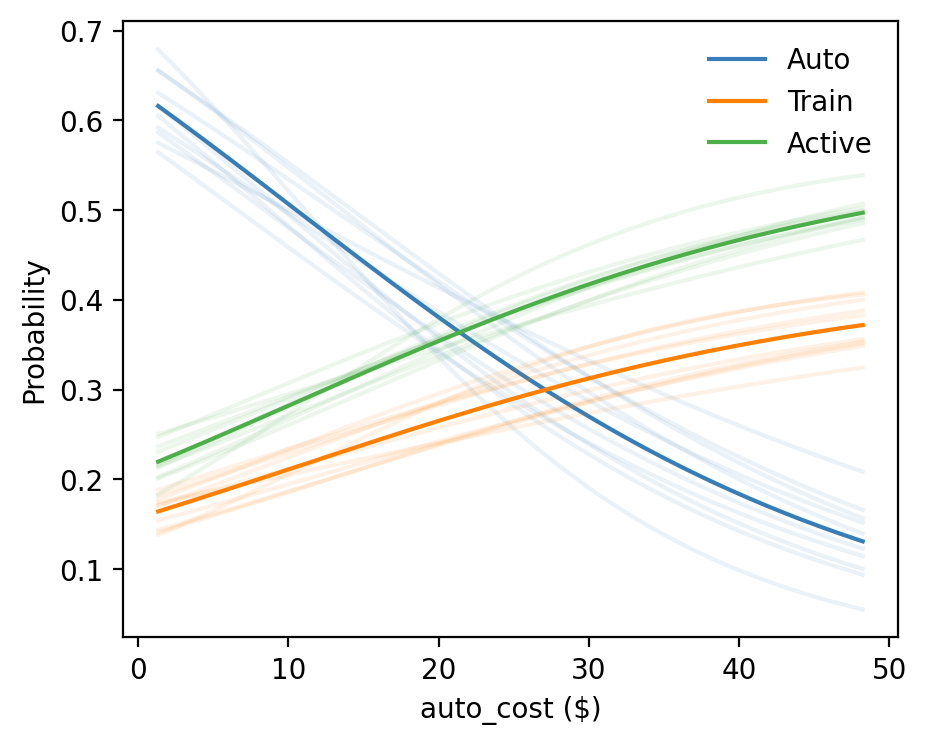}
			\subcaption{TasteNet}
		\end{subfigure}
		\begin{subfigure}{.325\linewidth}
			\includegraphics[width=\linewidth]{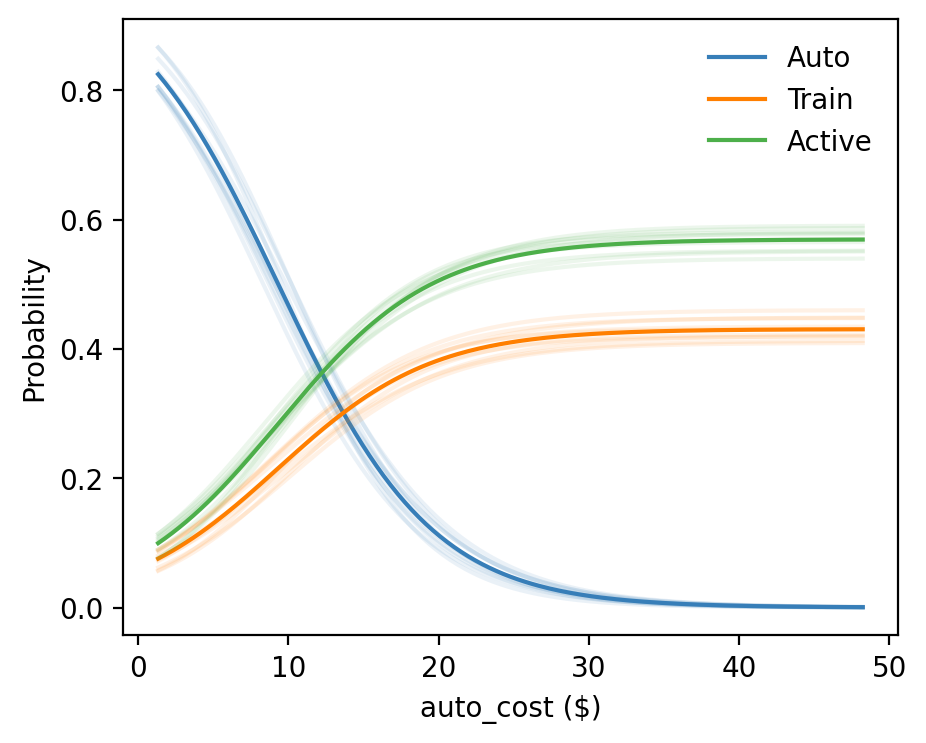}
			\subcaption{TasteNet, Exp}
		\end{subfigure}
		\begin{subfigure}{.325\linewidth}
			\includegraphics[width=\linewidth]{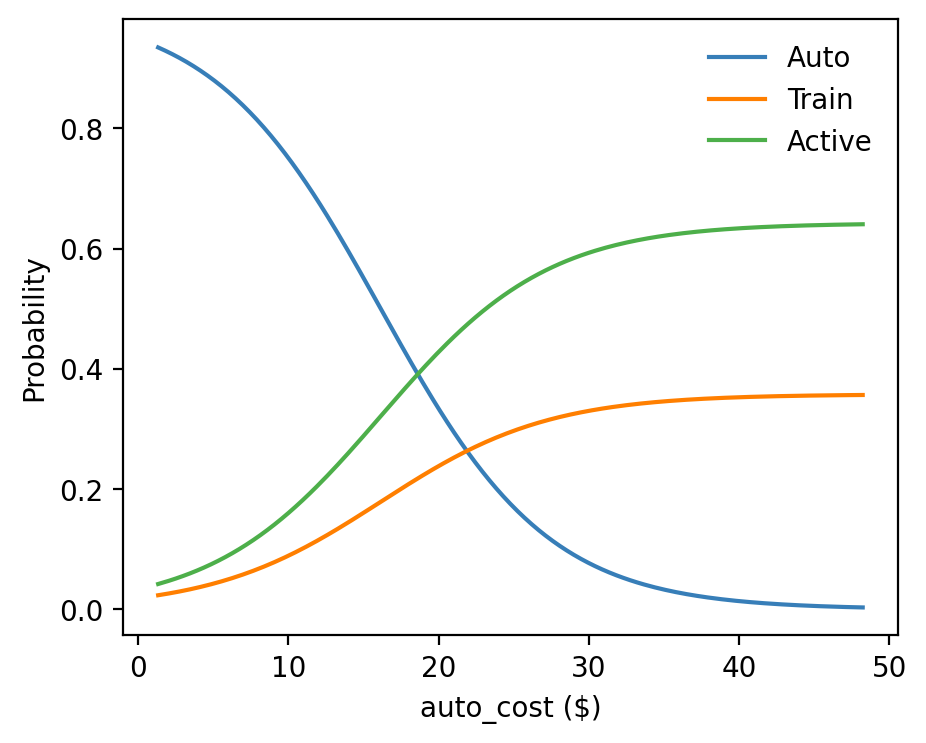}
			\subcaption{MNL}
		\end{subfigure}
		\caption{Individual demands as functions of driving costs (10K-Random, CMAP).}
		\label{fig:sub_10k}
	\end{figure}

	Using two 10K-Random datasets from Chicago and London, we evaluate the regularized DNNs by five metrics in the test sets: log-likelihood, accuracy, and $F_1$ score that capture the models' predictive power, as well as strong and weak regularities that describe their behavioral regularity. \cref{tab:10k_random} summarizes the performance of DNNs and TasteNets with optimal regularization strengths, TasteNets with hard constraints, alongside the DNN, TasteNet, and MNL benchmarks. The training and validation performance of the models are summarized in \cref{sec:appendix_c.1}. In each row, the best and second best metrics are marked in bold and underlined, respectively. For DNNs and TasteNets, each metric is averaged across ten trained model replications, with standard deviations shown in parentheses. To illustrate demand monotonicity, we plot the individual demand functions of the three alternatives for selected models in \cref{fig:sub_10k}, where light and dark curves represent the results of training replications and ensembles, respectively. \cref{fig:sub_10k} uses an ``average individual'' as the market representative, and varies the driving cost while keeping all other variables constant. Furthermore, \cref{sec:appendix_d} illustrates the impact of regularization strength on individual demands. By examining the large sample and in-domain scenarios (10K-Random), we have three major empirical findings. 
	
	\begin{table}[!htb]
		\caption{Model performance in the test sets of 10K-Random.}
		\label{tab:10k_random}
		\resizebox{\linewidth}{!}{
			\begin{tabular}{l|cccc|cccccc|c}
				\toprule
				Metric: & \multicolumn{4}{c|}{DNN} & \multicolumn{6}{c|}{TasteNet} & RUM \\
				Mean (SD) & No GR & PGR & UGR & LGR & No GR & PGR & UGR & LGR & ReLU & Exp & MNL \\
				\midrule
				\multicolumn{12}{l}{Panel 1: CMAP data, sum-XGR} \\
				\midrule
				Log-likelihood & $-1351.9$ & $\mathbf{-1344.3}$ & $-1350.2$ & $\underline{-1347.4}$ & $-1438.3$ & $-1438.4$ & $-1439.0$ & $-1438.4$ & $-1463.0$ & $-1439.7$ & $-1426.3$ \\
				& (4.697) & (5.521) & (5.803) & (5.110) & (5.834) & (6.027) & (5.992) & (6.099) & (20.05) & (6.073) & (0) \\
				Accuracy & \underline{0.729} & \textbf{0.730} & \underline{0.729} & \underline{0.729} & 0.713 & 0.713 & 0.713 & 0.713 & 0.712 & 0.717 & 0.718 \\
				& (0.003) & (0.002) & (0.002) & (0.002) & (0.003) & (0.002) & (0.003) & (0.002) & (0.002) & (0.003) & (0) \\
				$F_1$ score & 0.691 & \textbf{0.698} & 0.694 & \underline{0.696} & 0.654 & 0.654 & 0.654 & 0.654 & 0.650 & 0.664 & 0.669 \\
				& (0.005) & (0.004) & (0.004) & (0.004) & (0.006) & (0.005) & (0.005) & (0.005) & (0.002) & (0.004) & (0) \\
				Strong regularity & 0.888 & 0.990 & 0.982 & 0.991 & \underline{0.998} & \textbf{0.999} & \textbf{0.999} & \textbf{0.999} & 0.426 & \textbf{0.999} & \underline{0.998} \\
				& (0.066) & (0.003) & (0.012) & (0.003) & (0.002) & (0.001) & (0.001) & (0.001) & (0.320) & (0.001) & (0) \\
				Weak regularity & 0.922 & \underline{0.999} & 0.996 & \underline{0.999} & \underline{0.999} & \textbf{1.000} & \textbf{1.000} & \textbf{1.000} & \textbf{1.000} & \textbf{1.000} & \textbf{1.000} \\
				& (0.061) & (0.001) & (0.006) & (0.001) & (0.002) & (0.001) & (0.001) & (0.001) & (0.000) & (0.000) & (0) \\
				\midrule
				\multicolumn{12}{l}{Panel 2: CMAP data, norm-XGR} \\
				\midrule
				Log-likelihood & $\mathbf{-1351.9}$ & $\underline{-1353.9}$ & $-1362.0$ & $-1354.0$ & $-1438.3$ & $-1439.1$ & $-1469.5$ & $-1440.8$ & $-1463.0$ & $-1439.7$ & $-1426.3$ \\
				& (4.697) & (5.018) & (3.110) & (4.451) & (5.834) & (5.836) & (6.287) & (5.804) & (20.05) & (6.073) & (0) \\
				Accuracy & \textbf{0.729} & \textbf{0.729} & 0.725 & \underline{0.727} & 0.713 & 0.713 & 0.710 & 0.712 & 0.712 & 0.717 & 0.718 \\
				& (0.003) & (0.004) & (0.004) & (0.003) & (0.003) & (0.003) & (0.003) & (0.003) & (0.002) & (0.003) & (0) \\
				$F_1$ score & \textbf{0.691} & \underline{0.688} & 0.677 & 0.683 & 0.654 & 0.653 & 0.643 & 0.652 & 0.650 & 0.664 & 0.669 \\
				& (0.005) & (0.007) & (0.009) & (0.007) & (0.006) & (0.005) & (0.005) & (0.005) & (0.002) & (0.004) & (0) \\
				Strong regularity & 0.888 & 0.857 & 0.706 & 0.815 & \underline{0.998} & \underline{0.998} & 0.929 & 0.997 & 0.426 & \textbf{0.999} & \underline{0.998} \\
				& (0.066) & (0.069) & (0.051) & (0.068) & (0.002) & (0.002) & (0.030) & (0.004) & (0.320) & (0.001) & (0) \\
				Weak regularity & 0.922 & 0.893 & 0.756 & 0.851 & \underline{0.999} & \underline{0.999} & 0.941 & 0.998 & \textbf{1.000} & \textbf{1.000} & \textbf{1.000} \\
				& (0.061) & (0.067) & (0.051) & (0.069) & (0.002) & (0.002) & (0.026) & (0.003) & (0.000) & (0.000) & (0) \\
				\midrule
				\multicolumn{12}{l}{Panel 3: LTDS data, sum-XGR} \\
				\midrule
				Log-likelihood & $\underline{-1292.0}$ & $\mathbf{-1288.6}$ & $\mathbf{-1288.6}$ & $-1305.0$ & $-1305.8$ & $-1308.6$ & $-1317.6$ & $-1308.3$ & $-1335.5$ & $-1312.9$ & $-1366.9$ \\
				& (7.894) & (9.668) & (3.765) & (7.316) & (2.226) & (2.795) & (4.280) & (2.670) & (20.45) & (2.706) & (0) \\
				Accuracy & 0.729 & 0.729 & 0.727 & 0.724 & \underline{0.732} & 0.730 & 0.728 & 0.730 & 0.725 & \textbf{0.736} & 0.730 \\
				& (0.005) & (0.004) & (0.002) & (0.004) & (0.003) & (0.002) & (0.004) & (0.002) & (0.006) & (0.002) & (0) \\
				$F_1$ score & 0.727 & \underline{0.728} & 0.725 & 0.721 & \underline{0.728} & 0.726 & 0.723 & 0.726 & 0.721 & \textbf{0.733} & 0.726 \\
				& (0.004) & (0.004) & (0.002) & (0.006) & (0.003) & (0.002) & (0.004) & (0.002) & (0.006) & (0.002) & (0) \\
				Strong regularity & 0.950 & \underline{0.994} & \underline{0.994} & \textbf{0.997} & 0.942 & 0.964 & 0.987 & 0.972 & 0.933 & 0.990 & 0.993 \\
				& (0.034) & (0.004) & (0.009) & (0.003) & (0.021) & (0.013) & (0.008) & (0.012) & (0.043) & (0.002) & (0) \\
				Weak regularity & 0.969 & \underline{0.999} & 0.998 & \textbf{1.000} & 0.966 & \textbf{1.000} & \textbf{1.000} & \textbf{1.000} & \textbf{1.000} & \textbf{1.000} & \textbf{1.000} \\
				& (0.027) & (0.001) & (0.004) & (0.000) & (0.019) & (0.000) & (0.000) & (0.000) & (0.000) & (0.000) & (0) \\
				\bottomrule
			\end{tabular}
		}
	\end{table}

	First, the benchmark DNNs without gradient regularization outperform the MNL models in predictive power but underperform them in behavioral regularity, especially for the CMAP data. The benchmark DNN improves the MNL's log-likelihood by 5.2\% of its absolute value using the CMAP data and 5.5\% using the LTDS data. The empirical results show that log-likelihood is more sensitive to predictive performance than accuracy and $F_1$ score due to its probabilistic nature. Meanwhile, the difference between accuracy and $F_1$ score reflects whether the dataset is balanced: the two metrics are similar for the LTDS data but remain a gap for the car-dominated CMAP data, indicating that the former is more balanced in travel modes. Moreover, the benchmark DNNs are behaviorally irregular, as suggested by the relatively lower strong regularity (88.8\%) and weak regularity (92.2\%) using the CMAP data. This is consistent with finding illustrated in \cref{fig:sub_10k}a: the average market share of driving is non-monotonic w.r.t.\ driving cost, which is consistently presented in all local DNN models. This suggests that DNNs typically have high predictive performance but low behavioral regularity, aligning with the findings from many previous studies \citep{wang2020deeparch, wang2020deep, wong2021reslogit, xia2023random}. By contrast, the benchmark TasteNets have high behavioral regularity but relatively low predictive power, possibly due to their simpler NN architecture and linear-in-parameters specifications.
	
	Second, the DNNs with soft constraints outperform the TasteNets with hard constraints in terms of log-likelihood. By design, both ReLU and exponential constraints guarantee 100\% weak regularity, but only the latter can simultaneously guarantee high strong regularity. Despite being behaviorally regular, the TasteNets with either soft or hard constraints have lower log-likelihood than the DNNs with soft constraints. This example demonstrates the flexibility and approximation power of nonlinear DNN specifications compared with the linear-in-parameters TasteNet specifications. It might also suggest the existence of irregular behaviors for certain individuals or price ranges, highlighting the flexibility of soft constraints in capturing various behavioral mechanisms \citep{bagwell1991high, stiving2000price}. As shown in \cref{tab:10k_random,fig:sub_10k}, the exponentially constrained TasteNets resemble the MNL models in performance metrics and individual demands, reflecting the intrinsic connection between TasteNet and MNL models.
	
	Third, sum-based gradient regularization can improve the behavioral regularity of DNNs and TasteNets without sacrificing their predictive power. The demand functions of the regularized DNNs become highly monotonic, as indicated by strong and weak regularities both approaching 100\%. Meanwhile, sum-based gradient regularization is able to enhance DNNs' predictive power in both datasets. \cref{tab:10k_random} demonstrates that sum-PGR, which directly regularizes the demand functions, is slightly more effective than sum-UGR and sum-LGR, which exploit demand monotonicity through the chain rule. This finding is further elaborated by \cref{fig:sub_10k}b, where the regularized DNN has individual demand functions more consistent with the MNL: driving is less favored due to increased costs, while train and active mobility see higher demand. The regularized demand functions are more monotonic not only in the ensemble model, but also in training replications. We note that similar behavioral regularization effects of sum-based gradient regularizers are also observed for TasteNets on the LTDS data.
	
	Last but not least, norm-based gradient regularization fails to enhance behavioral regularity or predictive power. For example, norm-UGR with a small $\lambda$ can preserve accuracy and behavioral regularity for DNNs and TasteNets, but it would eventually lead to low predictive power and strong regularity as we increase $\lambda$. These results hold for all three norm-based approaches, potentially because they tend to flatten and smooth the demand curves. Therefore, the optimal DNNs and TasteNets in \cref{tab:10k_random} have small $\lambda$'s and look similar to the corresponding benchmarks (see \cref{fig:sub_10k}c for example). With strong norm-based gradient regularization, as shown in \cref{fig:sub_lambda}e, individual demand curves become almost flat and could not reflect the decision mechanism: travelers might not respond to cost changes at certain points, but are highly unlikely to be insensitive to all cost changes. Since smaller $\lambda$'s lead to better performance, we might conclude that the datasets or models do not change abruptly due to cost perturbations. In brief, although regularizing the gradient norm is a common practice in computer science \citep{drucker1991double,jakubovitz2018improving,sokolic2017robust}, it is not founded on prior beliefs in behavioral regularity and thus not effective for our purposes. For conciseness, we will not analyze norm-based gradient regularization in the next two scenarios.
	
	\subsubsection{Small sample scenario}
\begin{figure}[!htb]
		\centering
		\begin{subfigure}{.325\linewidth}
			\includegraphics[width=\linewidth]{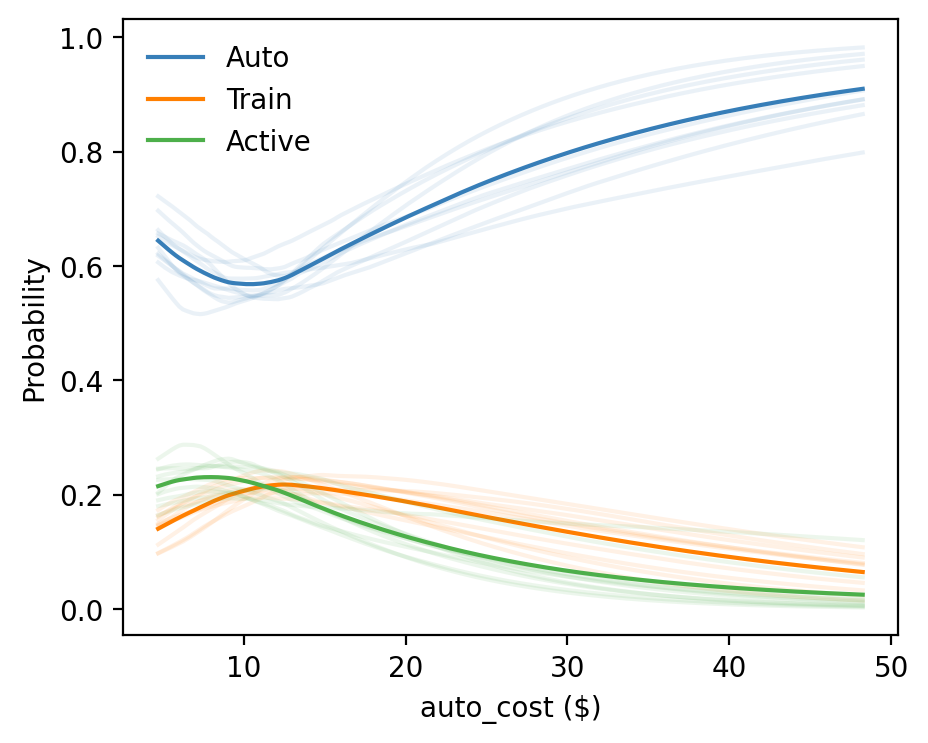}
			\subcaption{DNN}
		\end{subfigure}
		\begin{subfigure}{.325\linewidth}
			\includegraphics[width=\linewidth]{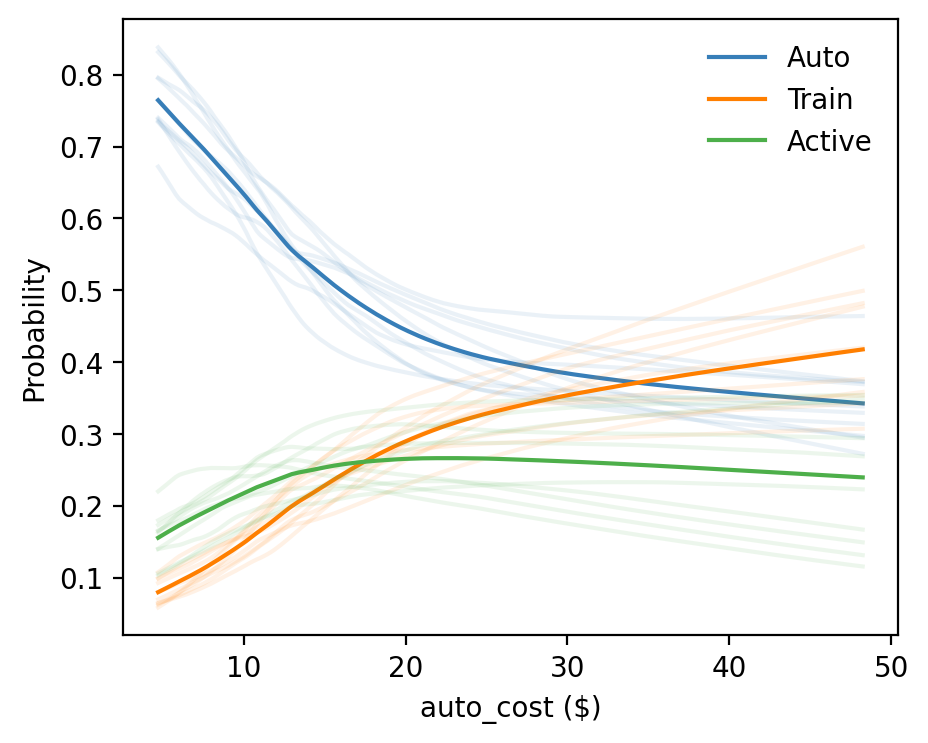}
			\subcaption{DNN, sum-PGR}
		\end{subfigure}
		\begin{subfigure}{.325\linewidth}
			\includegraphics[width=\linewidth]{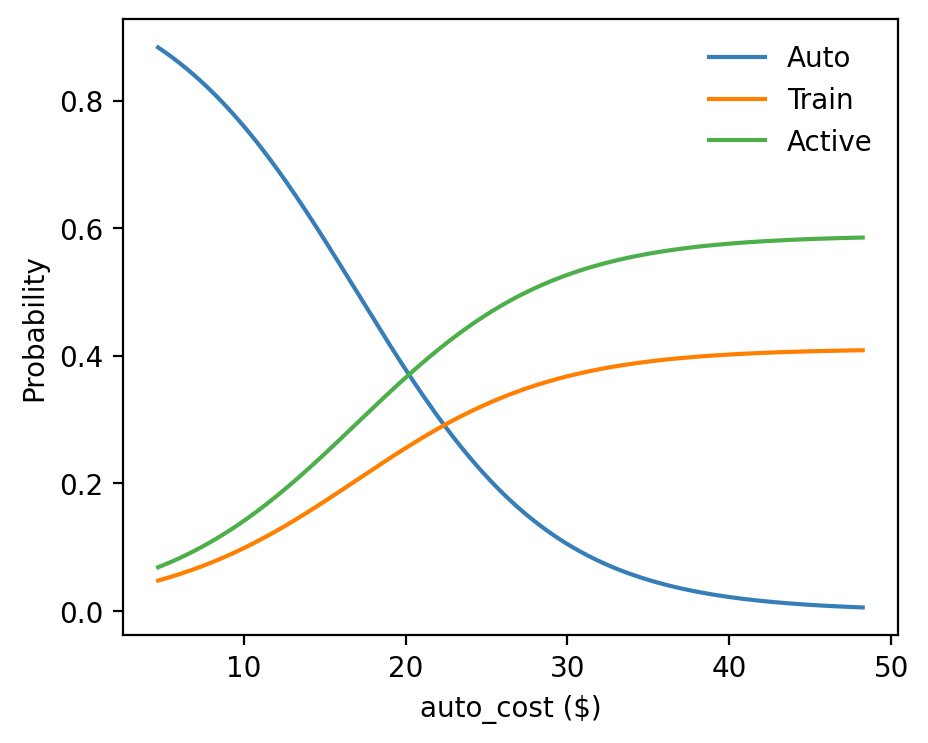}
			\subcaption{MNL}
		\end{subfigure}
		\par\smallskip
		\begin{subfigure}{.325\linewidth}
			\includegraphics[width=\linewidth]{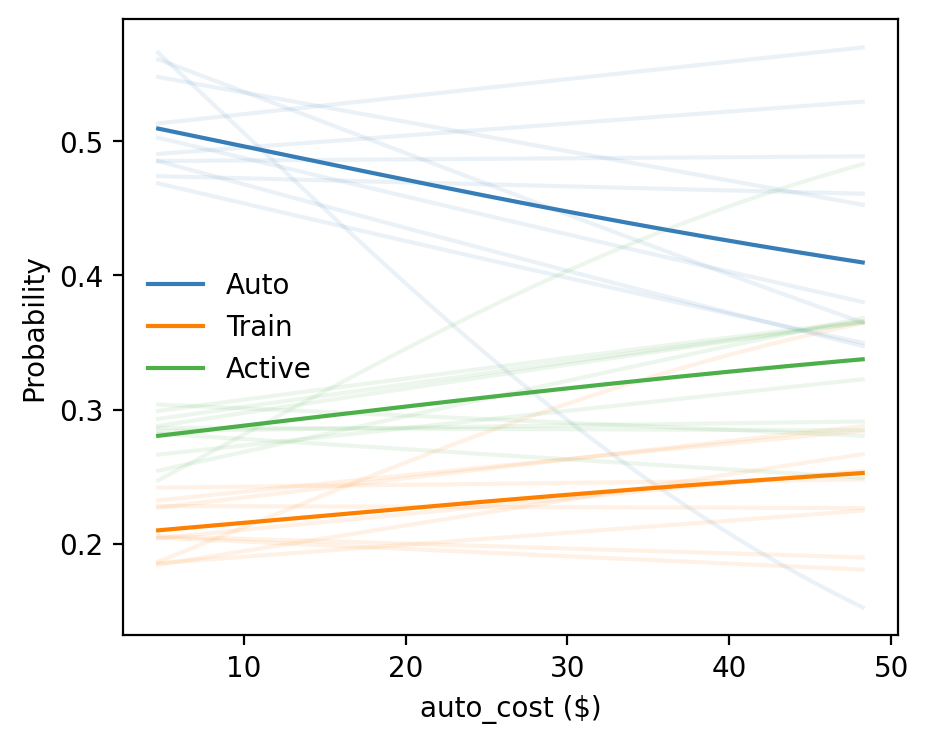}
			\subcaption{TasteNet}
		\end{subfigure}
		\begin{subfigure}{.325\linewidth}
			\includegraphics[width=\linewidth]{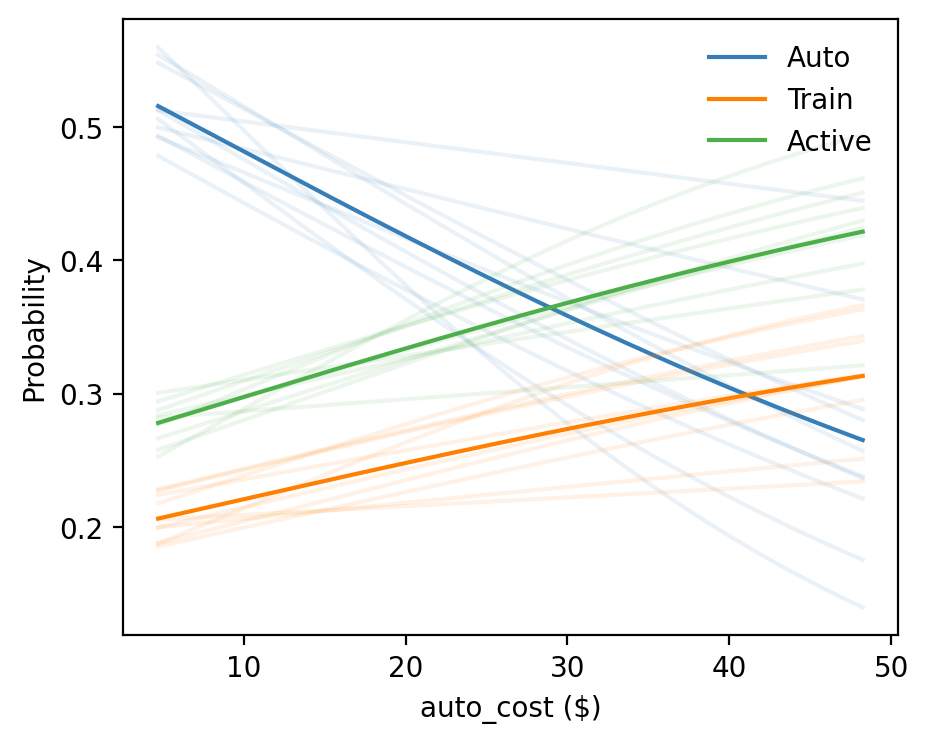}
			\subcaption{TasteNet, sum-PGR}
		\end{subfigure}
		\begin{subfigure}{.325\linewidth}
			\includegraphics[width=\linewidth]{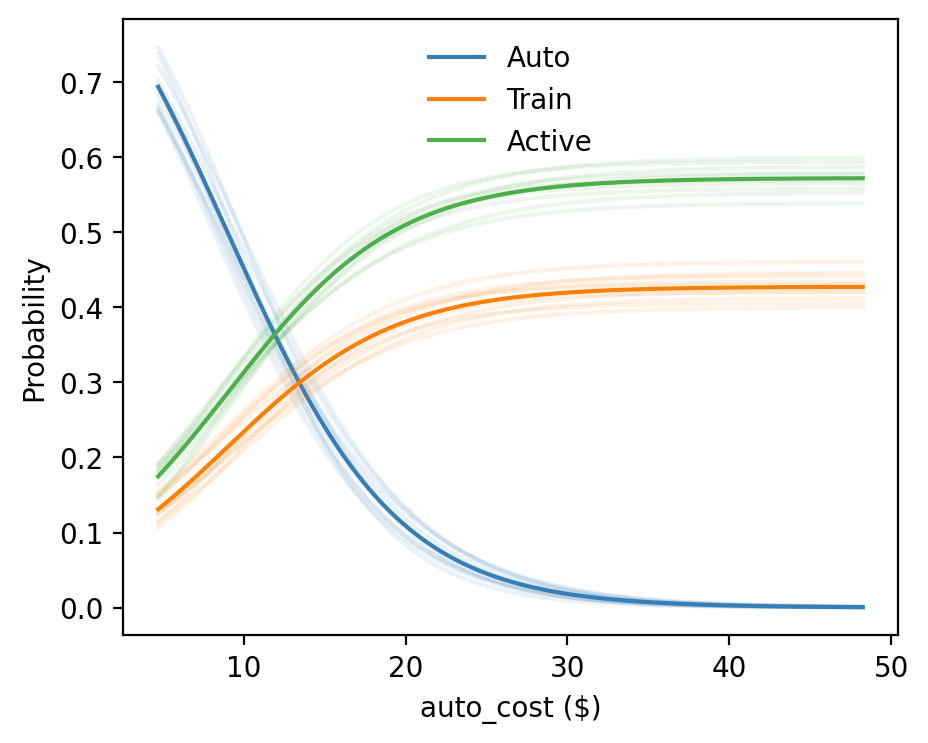}
			\subcaption{TasteNet, Exp}
		\end{subfigure}
		\caption{Individual demands as functions of driving costs (1K-Random, CMAP).}
		\label{fig:sub_1k}
	\end{figure}

	In this subsection, we focus on exemplifying the effectiveness of the proposed gradient regularizers in a typical choice modeling scenario where the number of available samples is limited due to resource restrictions or privacy concerns. The same analysis is applied to the 1K-Random CMAP and LTDS datasets. \cref{tab:1k_random} illustrates the performance of DNNs, TasteNets, and benchmark models, while \cref{fig:sub_1k} elaborates on the individual demand functions. Furthermore, \cref{sec:appendix_c.2} summarizes the training and validation performance of the models.
	
	\begin{table}[!htb]
		\caption{Model performance in the test sets of 1K-Random.}
		\label{tab:1k_random}
		\resizebox{\linewidth}{!}{
			\begin{tabular}{l|cccc|cccccc|c}
				\toprule
				Metric: & \multicolumn{4}{c|}{DNN} & \multicolumn{6}{c|}{TasteNet} & RUM \\
				Mean (SD) & No GR & PGR & UGR & LGR & No GR & PGR & UGR & LGR & ReLU & Exp & MNL \\
				\midrule
				\multicolumn{10}{l}{Panel 1: CMAP data, sum-XGR} \\
				\midrule
				Log-likelihood & $-375.1$ & $\mathbf{-367.9}$ & $-374.2$ & $\underline{-369.7}$ & $-389.3$ & $-385.2$ & $-383.1$ & $-386.6$ & $-386.3$ & $-378.0$ & $-380.2$ \\
				& (3.096) & (4.181) & (3.298) & (3.304) & (2.838) & (3.351) & (3.139) & (3.447) & (2.506) & (4.123) & (0) \\
				Accuracy & 0.705 & 0.703 & 0.676 & 0.696 & 0.700 & 0.679 & 0.686 & 0.677 & 0.696 & \underline{0.707} & \textbf{0.718} \\
				& (0.002) & (0.007) & (0.010) & (0.011) & (0.005) & (0.012) & (0.010) & (0.012) & (0.005) & (0.004) & (0) \\
				$F_1$ score & \underline{0.648} & 0.645 & 0.559 & 0.619 & 0.619 & 0.563 & 0.580 & 0.559 & 0.610 & 0.637 & \textbf{0.665} \\
				& (0.004) & (0.018) & (0.027) & (0.030) & (0.012) & (0.028) & (0.023) & (0.028) & (0.011) & (0.010) & (0) \\
				Strong regularity & 0.664 & 0.985 & 0.985 & 0.985 & 0.659 & 0.979 & 0.983 & 0.979 & 0.461 & \textbf{0.999} & \underline{0.996} \\
				& (0.173) & (0.009) & (0.011) & (0.011) & (0.129) & (0.015) & (0.013) & (0.013) & (0.195) & (0.001) & (0) \\
				Weak regularity & 0.728 & \underline{0.999} & 0.997 & \underline{0.999} & 0.685 & 0.997 & 0.995 & 0.992 & \textbf{1.000} & \textbf{1.000} & \textbf{1.000} \\
				& (0.162) & (0.001) & (0.005) & (0.002) & (0.128) & (0.004) & (0.006) & (0.009) & (0.000) & (0.000) & (0) \\
				\midrule
				\multicolumn{12}{l}{Panel 2: LTDS data, sum-XGR} \\
				\midrule
				Log-likelihood & $\underline{-322.6}$ & $-328.8$ & $\mathbf{-317.8}$ & $-335.2$ & $-345.1$ & $-339.7$ & $-343.5$ & $-340.1$ & $-341.0$ & $-335.5$ & $-331.2$ \\
				& (3.455) & (8.146) & (3.440) & (11.30) & (2.337) & (2.023) & (3.035) & (1.974) & (2.572) & (2.951) & (0) \\
				Accuracy & 0.737 & 0.720 & \underline{0.740} & 0.717 & 0.725 & 0.734 & 0.724 & 0.733 & 0.729 & 0.733 & \textbf{0.746} \\
				& (0.007) & (0.013) & (0.007) & (0.021) & (0.004) & (0.006) & (0.009) & (0.006) & (0.006) & (0.008) & (0) \\
				$F_1$ score & 0.732 & 0.703 & \underline{0.734} & 0.694 & 0.711 & 0.720 & 0.705 & 0.718 & 0.715 & 0.725 & \textbf{0.740} \\
				& (0.007) & (0.020) & (0.008) & (0.038) & (0.005) & (0.006) & (0.013) & (0.005) & (0.007) & (0.009) & (0) \\
				Strong regularity & 0.904 & \textbf{0.999} & 0.992 & 0.997 & 0.939 & 0.964 & \textbf{0.999} & 0.965 & 0.924 & \textbf{0.999} & \underline{0.998} \\
				& (0.069) & (0.002) & (0.014) & (0.008) & (0.023) & (0.013) & (0.002) & (0.01) & (0.046) & (0.002) & (0) \\
				Weak regularity & 0.913 & \textbf{1.000} & 0.993 & 0.998 & 0.943 & 0.984 & \underline{0.999} & 0.981 & \textbf{1.000} & \textbf{1.000} & \textbf{1.000} \\
				& (0.067) & (0.001) & (0.014) & (0.006) & (0.024) & (0.012) & (0.002) & (0.013) & (0.000) & (0.000) & (0) \\
				\bottomrule
			\end{tabular}
		}
	\end{table}
	
	We find that our gradient regularizers are even more effective than in the large sample scenario. First, without gradient regularization, the benchmark DNNs' behavioral regularity becomes much worse with a small sample, e.g., the strong regularity drops by 22.4 percentage points and the weak regularity drops by 19.4 percentage points using the CMAP data. Similarly, shallower NNs like the benchmark TasteNets also have worse behavioral regularity in the small sample scenario. Meanwhile, the benchmark models fail to outperform MNL in predictive power, especially for the CMAP data. As shown in \cref{fig:sub_1k}a, the benchmark DNN's individual demand curves are non-monotonic and contradictory to the law of demand. Second, sum-based gradient regularization succeeds in enhancing all metrics of DNNs and most metrics of TasteNets, as shown in \cref{tab:1k_random} and \cref{fig:sub_1k}. In other words, the regularized DNNs and TasteNets outperform their corresponding benchmark models in both predictive power and behavioral regularity. In the CMAP case, sum-PGR improves the benchmark DNN's log-likelihood by 1.9\% of its absolute value and its strong regularity by 32.1 percentage points. Moreover, from \cref{tab:1k_random,fig:sub_1k} we can draw similar findings for the hard constraints, e.g., the exponentially constrained TasteNets resemble the MNL models in performance metrics and individual demands. In addition, ensembles of regularized models have comparable overall performance to the MNL when using such a small sample size.
	
	\subsubsection{Out-of-domain generalization}
	The large and small sample scenarios above assist in examining only in-domain generalization for the random split of training and test data. Although random split is the most common practice, we are also interested in the out-of-domain generalizability of DNNs. Out-of-domain generalization is highly relevant to transportation engineering, system design, and urban planning, such as the cross-city policy transfer: when assessing whether a city should build a subway, transportation planners sometimes cite the ridership of subway systems in other cities. Our data split scheme in the 10K-Sorted datasets can be interpreted as using the patterns of individual choice behavior in one city (as in the training set) to extrapolate those in another city (as in the test set). Here we emulate such policy setting by testing the DNNs' out-of-domain generalizability using predictive power and behavioral regularity metrics.
	
	\begin{table}[!htb]
		\caption{Model performance in the test sets of 10K-Sorted.}
		\label{tab:10k_sorted}
		\resizebox{\linewidth}{!}{
			\begin{tabular}{l|cccc|cccccc|c}
				\toprule
				Metric: & \multicolumn{4}{c|}{DNN} & \multicolumn{6}{c|}{TasteNet} & RUM \\
				Mean (SD) & No GR & PGR & UGR & LGR & No GR & PGR & UGR & LGR & ReLU & Exp & MNL \\
				\midrule
				\multicolumn{12}{l}{Panel 1: CMAP data, sum-XGR} \\
				\midrule
				Log-likelihood & $-1356.1$ & $-1243.5$ & $\mathbf{-1167.9}$ & $\underline{-1232.4}$ & $-1485.9$ & $-1873.7$ & $-1901.4$ & $-2003.9$ & $-2315.4$ & $-1995.0$ & $-2025.7$ \\
				& (134.1) & (55.15) & (27.64) & (54.87) & (47.92) & (45.52) & (36.62) & (50.23) & (395.4) & (51.83) & (0) \\
				Accuracy & 0.783 & \underline{0.788} & \textbf{0.789} & \underline{0.788} & 0.750 & 0.726 & 0.725 & 0.705 & 0.578 & 0.723 & 0.722 \\
				& (0.011) & (0.002) & (0.003) & (0.002) & (0.014) & (0.008) & (0.006) & (0.008) & (0.134) & (0.007) & (0) \\
				$F_1$ score & 0.722 & \underline{0.726} & \textbf{0.727} & 0.724 & 0.707 & 0.721 & 0.720 & 0.708 & 0.603 & 0.719 & 0.721 \\
				& (0.010) & (0.006) & (0.003) & (0.005) & (0.006) & (0.005) & (0.004) & (0.006) & (0.112) & (0.005) & (0) \\
				Strong regularity & 0.317 & 0.857 & 0.923 & 0.865 & \textbf{1.000} & 0.980 & 0.978 & 0.981 & 0.933 & 0.969 & \underline{0.984} \\
				& (0.240) & (0.099) & (0.087) & (0.071) & (0.000) & (0.004) & (0.004) & (0.004) & (0.169) & (0.005) & (0) \\
				Weak regularity & 0.487 & 0.974 & \underline{0.983} & 0.977 & \textbf{1.000} & \textbf{1.000} & \textbf{1.000} & \textbf{1.000} & \textbf{1.000} & \textbf{1.000} & \textbf{1.000} \\
				& (0.230) & (0.037) & (0.040) & (0.023) & (0.000) & (0.000) & (0.000) & (0.001) & (0.000) & (0.000) & (0) \\
				\midrule
				\multicolumn{12}{l}{Panel 2: LTDS data, sum-XGR (public transit cost)} \\
				\midrule
				Log-likelihood & $-1137.9$ & $\underline{-1110.4}$ & $\mathbf{-1108.4}$ & $-1112.1$ & $-1221.1$ & $-1170.2$ & $-1171.1$ & $-1175.0$ & $-1182.3$ & $-1150.1$ & $-1301.6$ \\
				& (34.33) & (23.86) & (21.15) & (26.67) & (33.10) & (27.01) & (28.54) & (30.25) & (54.11) & (29.62) & (0) \\
				Accuracy & 0.777 & 0.784 & \textbf{0.794} & 0.782 & 0.776 & \underline{0.786} & 0.785 & 0.783 & 0.782 & 0.781 & 0.780 \\
				& (0.010) & (0.007) & (0.007) & (0.006) & (0.009) & (0.008) & (0.008) & (0.007) & (0.017) & (0.008) & (0) \\
				$F_1$ score & 0.768 & \underline{0.776} & \textbf{0.778} & \underline{0.776} & 0.765 & 0.772 & 0.772 & 0.772 & 0.769 & 0.768 & 0.770 \\
				& (0.011) & (0.006) & (0.008) & (0.006) & (0.008) & (0.008) & (0.008) & (0.007) & (0.018) & (0.008) & (0) \\
				Strong regularity & 0.185 & 0.968 & 0.979 & 0.907 & 0.313 & 0.878 & 0.872 & 0.720 & 0.248 & \textbf{0.982} & \underline{0.980} \\
				& (0.075) & (0.032) & (0.029) & (0.062) & (0.007) & (0.095) & (0.095) & (0.083) & (0.084) & (0.005) & (0) \\
				Weak regularity & 0.207 & 0.977 & 0.984 & 0.925 & 0.343 & \underline{0.993} & 0.988 & 0.900 & \textbf{1.000} & \textbf{1.000} & \textbf{1.000} \\
				& (0.080) & (0.026) & (0.025) & (0.055) & (0.008) & (0.010) & (0.012) & (0.045) & (0.000) & (0.000) & (0) \\
				\bottomrule
			\end{tabular}
		}
	\end{table}
	
	\begin{figure}[!htb]
		\centering
		\begin{subfigure}{.325\linewidth}
			\includegraphics[width=\linewidth]{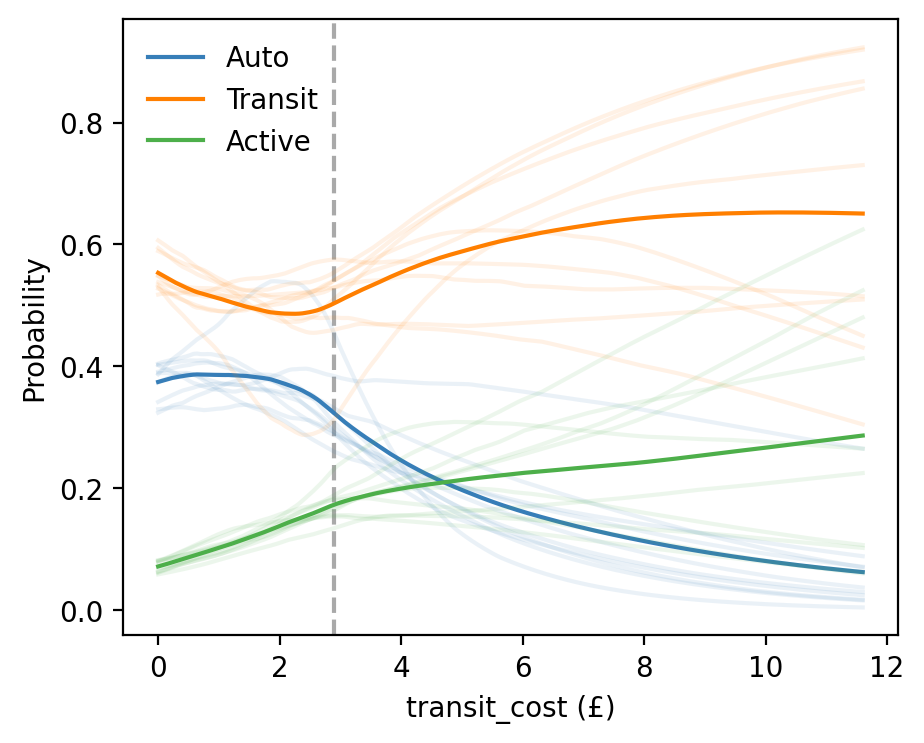}
			\subcaption{DNN}
		\end{subfigure}
		\begin{subfigure}{.325\linewidth}
			\includegraphics[width=\linewidth]{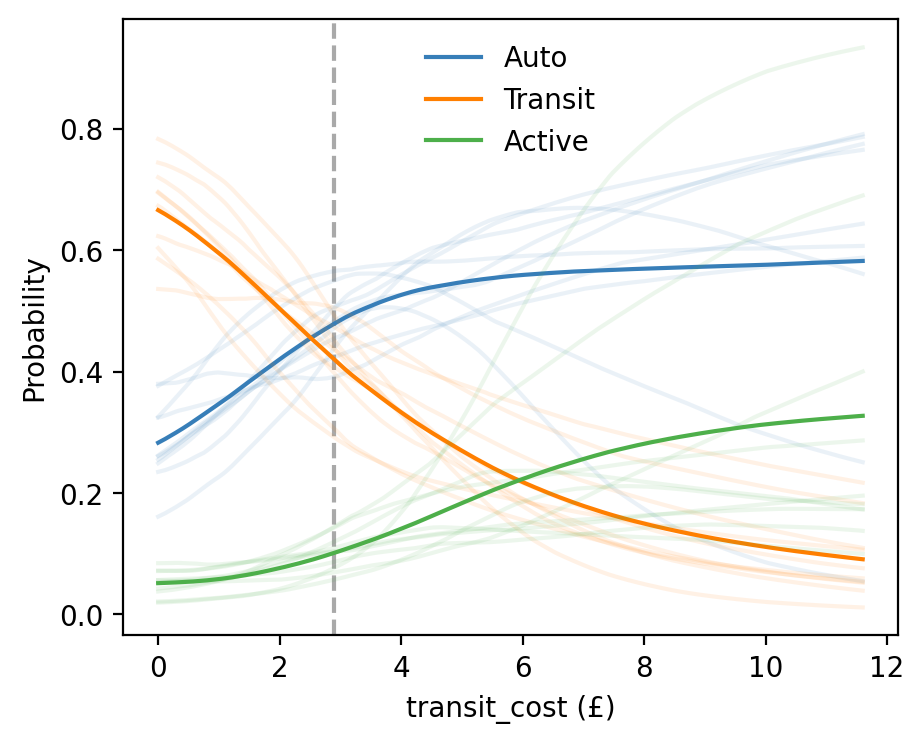}
			\subcaption{DNN, sum-UGR}
		\end{subfigure}
		\begin{subfigure}{.325\linewidth}
			\includegraphics[width=\linewidth]{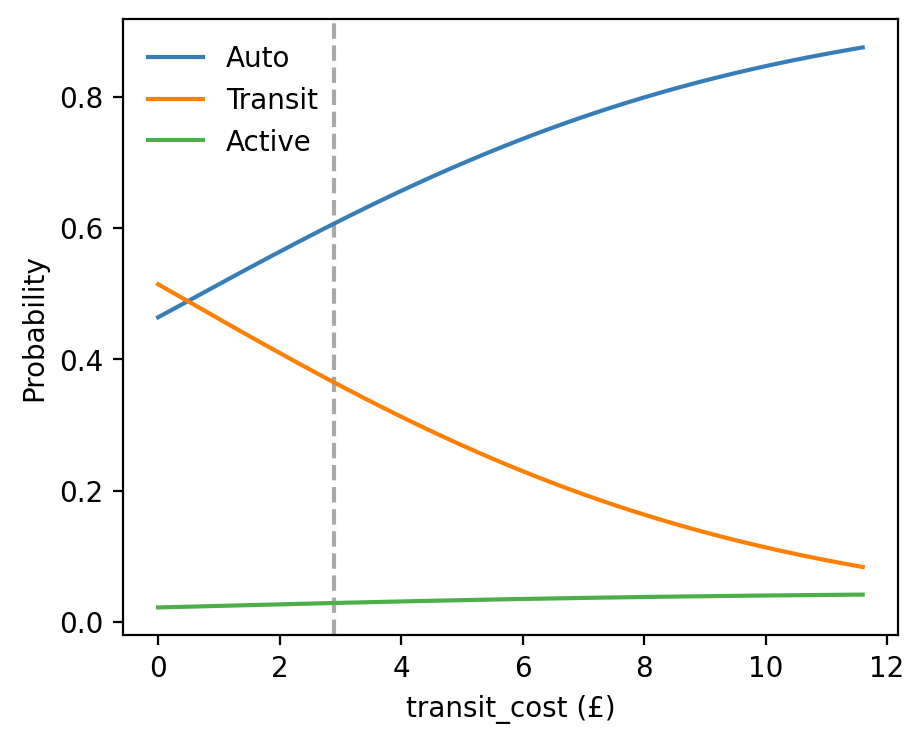}
			\subcaption{MNL}
		\end{subfigure}
		\par\smallskip
		\begin{subfigure}{.325\linewidth}
			\includegraphics[width=\linewidth]{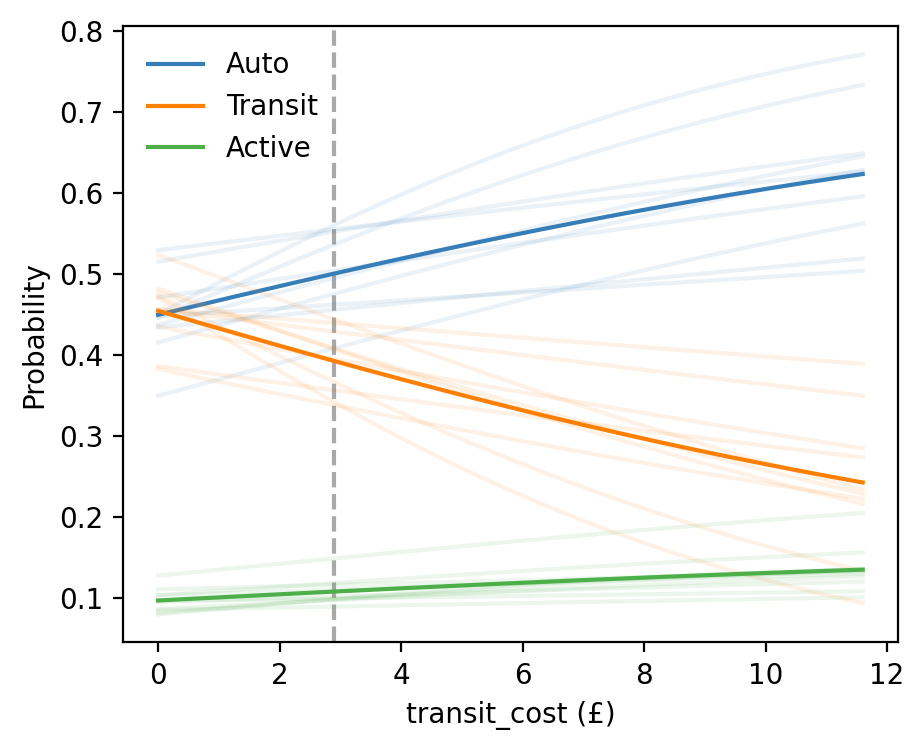}
			\subcaption{TasteNet}
		\end{subfigure}
		\begin{subfigure}{.325\linewidth}
			\includegraphics[width=\linewidth]{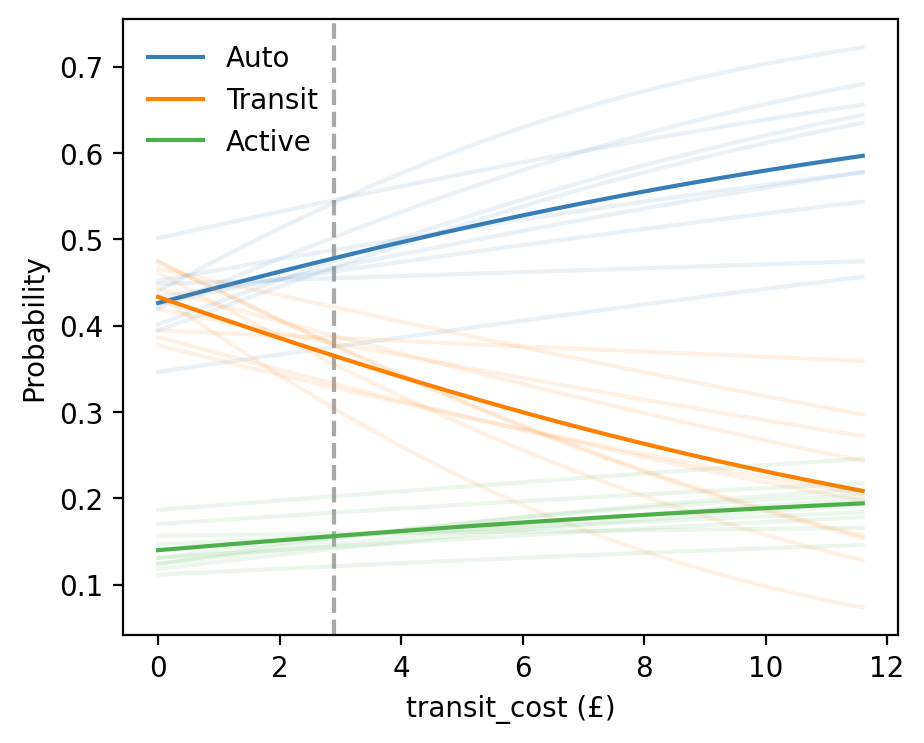}
			\subcaption{TasteNet, sum-PGR}
		\end{subfigure}
		\begin{subfigure}{.325\linewidth}
			\includegraphics[width=\linewidth]{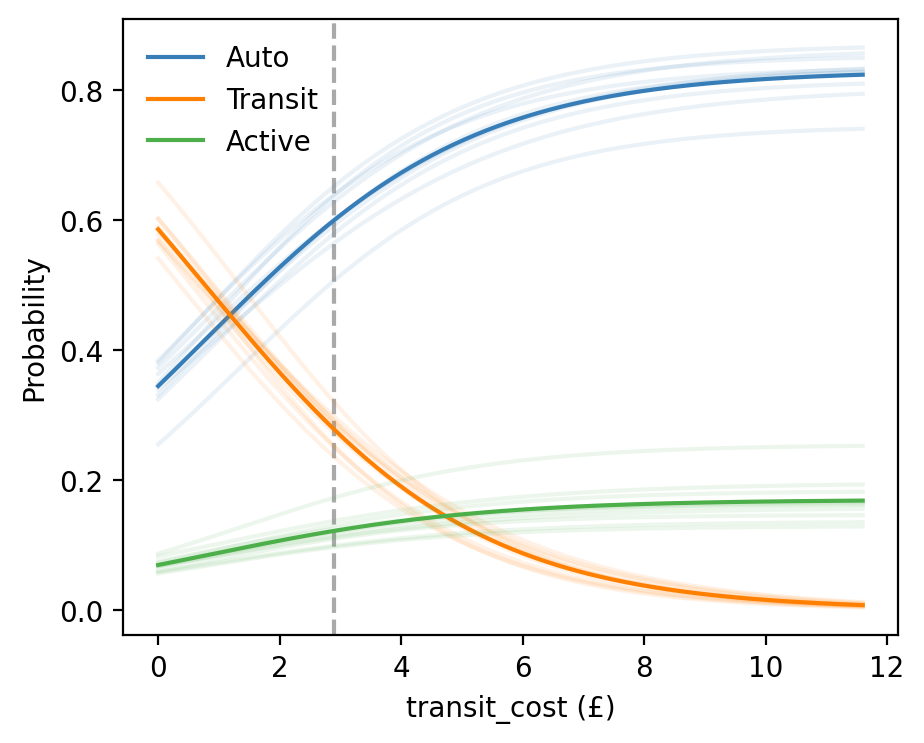}
			\subcaption{TasteNet, Exp}
		\end{subfigure}
		\caption{Individual demands as functions of public transit costs (10K-Sorted, LTDS).}
		\label{fig:sub_sorted}
	\end{figure}
	
	The results suggest that gradient regularization could drastically improve the out-of-domain generalizability of DNNs, even more effectively than improving their in-domain generalizability. \cref{tab:10k_sorted} summarizes the performance of DNNs, TasteNets, and benchmark models in the test sets, while their training and validation performance are summarized in \cref{sec:appendix_c.3}. Under this setting, DNNs and TasteNets exhibit great flexibility with higher log-likelihood than the MNL model. However, behavioral regularity could be relatively low for these NN models, as reflected in their regularity metrics. \cref{fig:sub_sorted} visualizes individual demands as functions of public transit costs using the LTDS data, in which the benchmark DNN performs unreasonably in the test set, as shown to the right of the data split threshold (dashed gray line). Second, sum-based gradient regularization has the potential to simultaneously improve the predictive power and behavioral regularity of benchmark DNNs and TasteNets. For an extreme case, sum-UGR dramatically raises the strong regularity of the benchmark DNN from 0.185 to 0.979 using the LTDS data. \cref{tab:10k_sorted} also reveals the case-dependent effects of hard constraints: they can significantly improve both behavioral regularity and predictive power of a poorly behaved benchmark TasteNet (the LTDS case), but are ineffective for a benchmark that is already highly regular (the CMAP case).
	
	\subsection{Trade-off between predictive power and behavioral regularity}
	\label{sec:tradeoff}
	\cref{sec:model_performance} demonstrates the potentials of sum-based gradient regularization in enhancing the behavioral regularity and predictive power of benchmark DNNs and TasteNets under three settings. This subsection will further investigate the trade-off between predictive power and behavioral regularity in these scenarios. Although the optimization objective in \cref{eq:loss_sum_reg} demonstrates a clear substitution effect between predictive power and behavioral regularity in the training set, it remains an open question whether this effect persists in the test set. As detailed below, we find the same substitution effect in the test sets of 10K-Random, indicating that predictive power decreases and behavioral regularity increases with stronger sum-based gradient regularization. On the other hand, interestingly, complementary effects between predictive power and behavioral regularity are observed for in-domain generalization with 1K-Random and out-of-domain generalization with 10K-Sorted. This suggests that adequate sum-based gradient regularization can enhance both predictive power and behavioral regularity, especially in the typical choice modeling setting with limited sample sizes, and for the purpose of policy evaluations.
	
	\subsubsection{Large sample scenario: substitution effects}
\begin{figure}[!htb]
		\centering
		\begin{subfigure}{.325\linewidth}
			\centering
			\includegraphics[height=4.5cm]{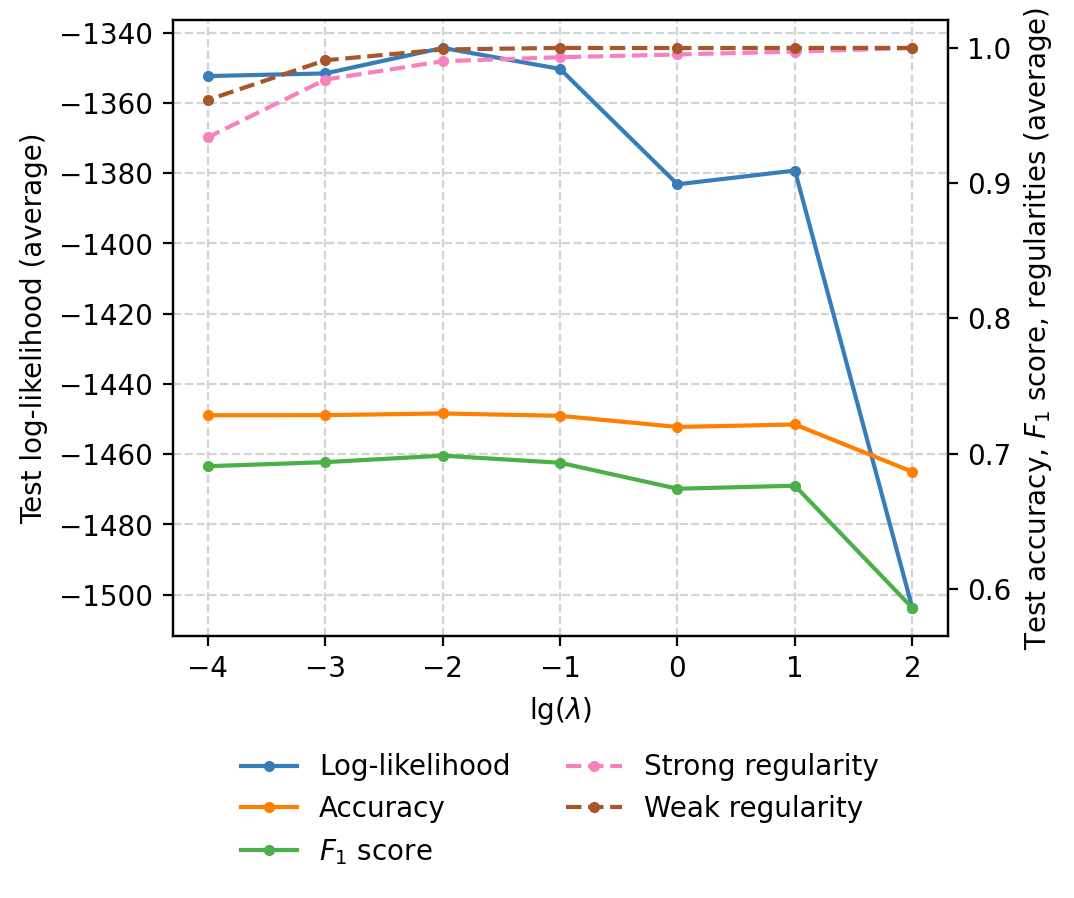}
			\subcaption{DNN, sum-PGR (CMAP)}
		\end{subfigure}
		\begin{subfigure}{.325\linewidth}
			\centering
			\includegraphics[height=4.5cm]{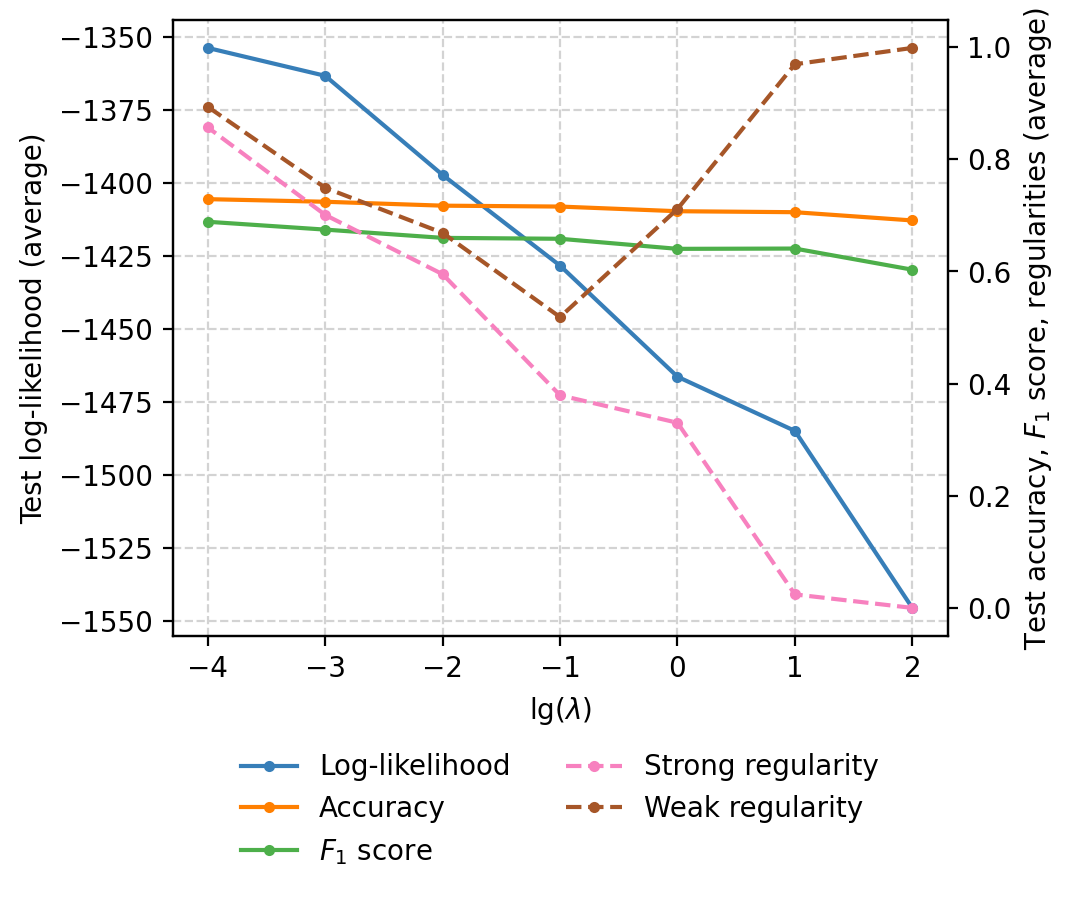}
			\subcaption{DNN, norm-PGR (CMAP)}
		\end{subfigure}
		\begin{subfigure}{.325\linewidth}
			\centering
			\includegraphics[height=4.5cm]{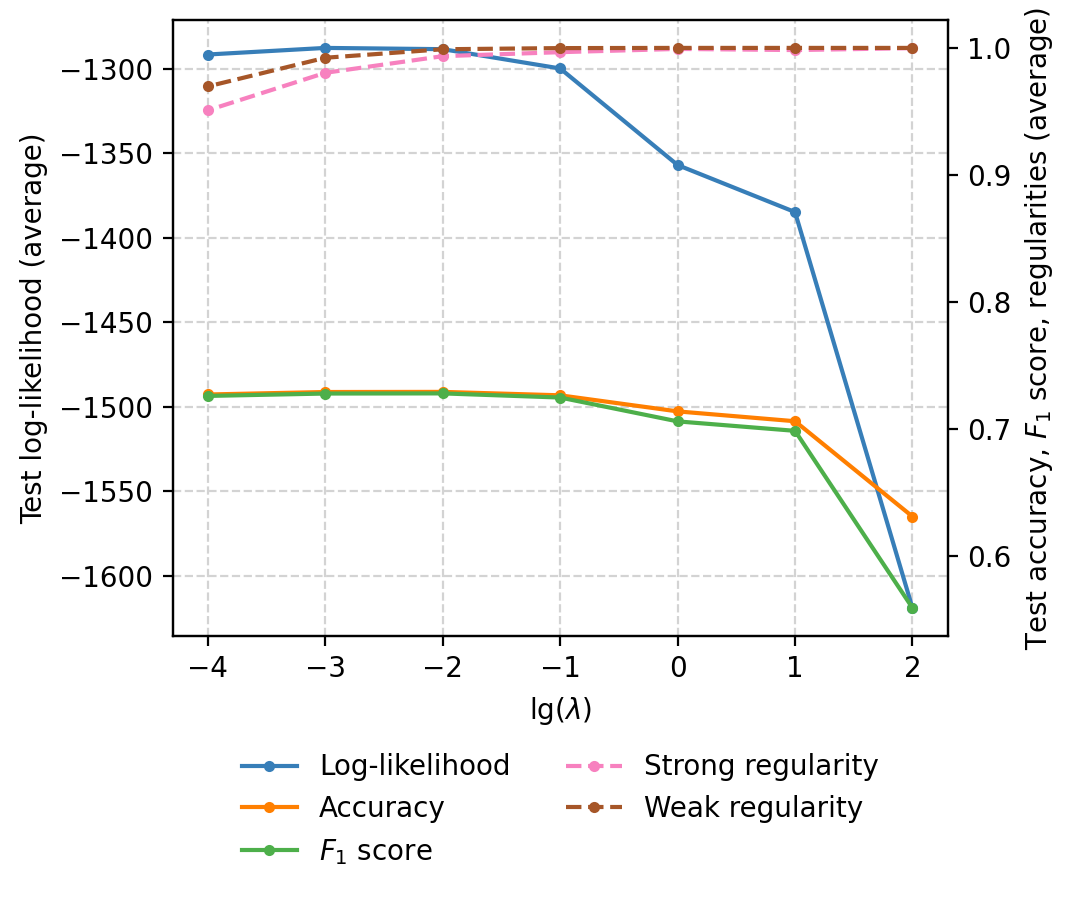}
			\subcaption{DNN, sum-PGR (LTDS)}
		\end{subfigure}
		\par\smallskip
		\begin{subfigure}{.325\linewidth}
			\centering
			\includegraphics[height=4.5cm]{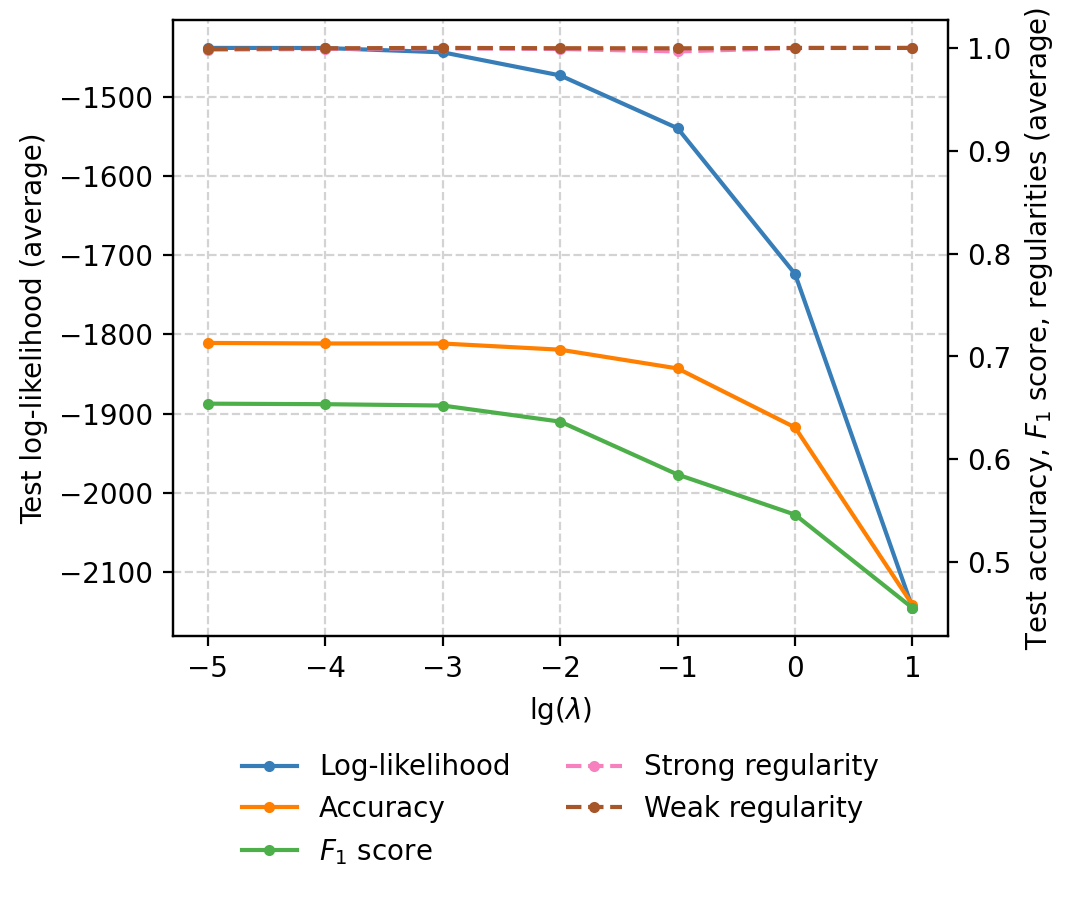}
			\subcaption{TasteNet, sum-PGR (CMAP)}
		\end{subfigure}
		\begin{subfigure}{.325\linewidth}
			\centering
			\includegraphics[height=4.5cm]{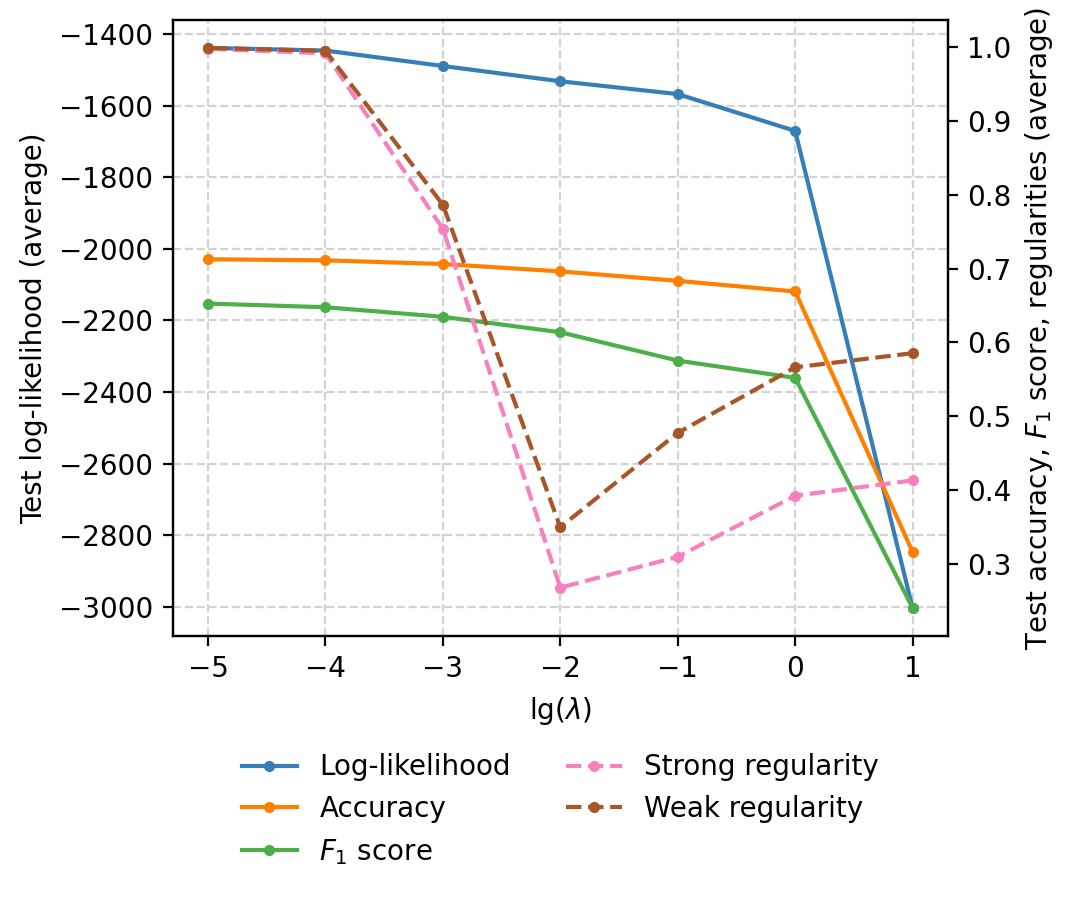}
			\subcaption{TasteNet, norm-PGR (CMAP)}
		\end{subfigure}
		\begin{subfigure}{.325\linewidth}
			\centering
			\includegraphics[height=4.5cm]{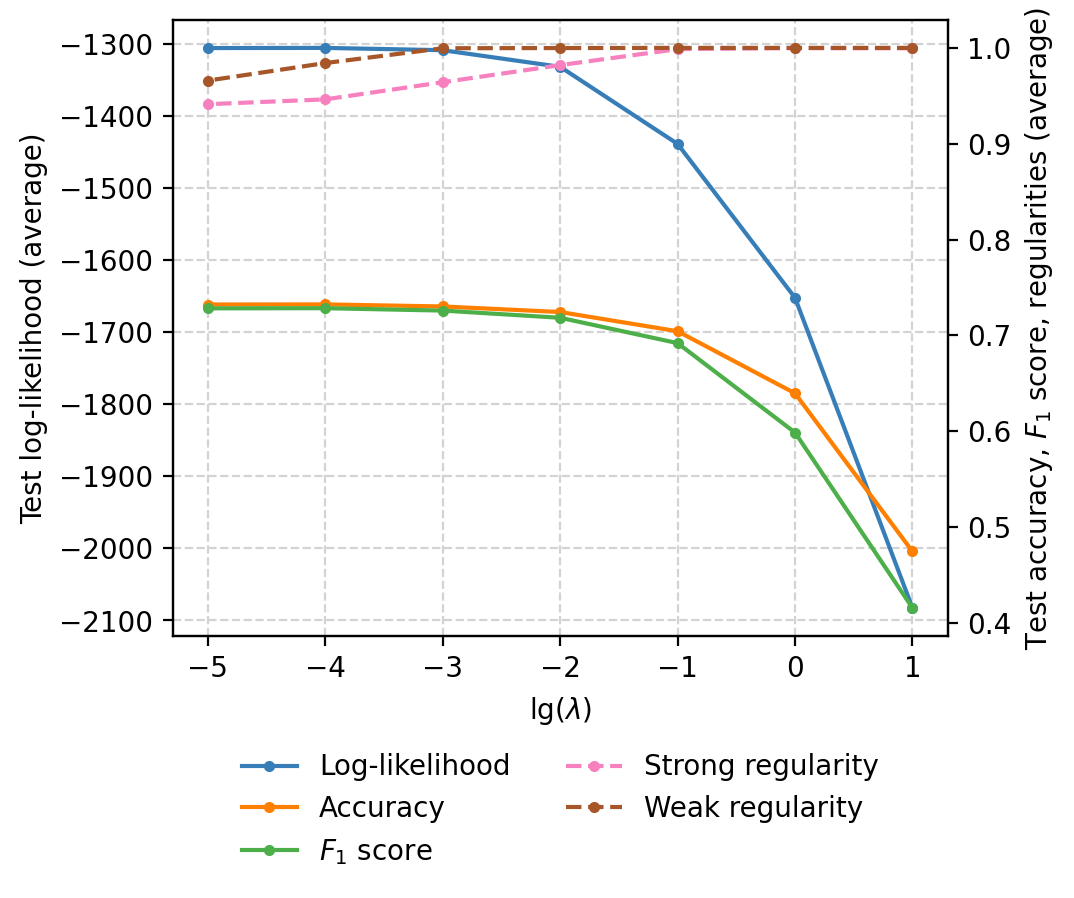}
			\subcaption{TasteNet, sum-PGR (LTDS)}
		\end{subfigure}
		\caption{Effects of regularization strength (10K-Random).}
		\label{fig:lambda_10k_random}
	\end{figure}

	In a large sample scenario, higher behavioral regularity often has a trade-off with lower predictive power. The effects of regularization strength $\lambda$ on the five metrics are illustrated in \cref{fig:lambda_10k_random} for both 10K-Random CMAP and LTDS datasets, where the horizontal axis uses logarithmic scale $\lg(\lambda)$. As shown in \cref{fig:lambda_10k_random}, DNNs' model fit declines with increasing $\lambda$, especially in terms of log-likelihood, which is more sensitive to $\lambda$ than accuracy and $F_1$ score. The decline in predictive power is particularly noticeable after a critical point, such as $\lambda = 10$ for DNNs with sum-PGR. In other words, although regularization would generally reduce predictive power, there exists a range of $\lambda$ that almost preserves predictive power while enhancing behavioral regularity, such as $\lambda \le 0.1$ for DNNs with sum-PGR, consistent with our findings in \cref{sec:performance_10k_random}.
	
	\cref{fig:lambda_10k_random} also presents two important differences between log-likelihood and accuracy or $F_1$ score, as well as between sum- and norm-based gradient regularization. First, log-likelihood is much more sensitive than accuracy or $F_1$ score in measuring predictive power, which is theoretically valid and empirically expected. Moreover, presenting accuracy and $F_1$ score together might be a good idea for imbalanced data. Therefore, we recommend that future studies use different metrics when comparing the performance of DCMs. Second, our behavioral regularity metrics can demonstrate the flattening effects of norm-based gradient regularization on individual demand curves, as indicated by weak regularity approaching 100\% and strong regularity approaching 0 for very large $\lambda$'s (see \cref{fig:lambda_10k_random}b). In brief, strong regularity might be more appropriate than the weak one, at least for describing the global declining trend of demand curves, although weak behavioral regularity metric could still be important for describing local insensitivity to cost changes.
	
	\subsubsection{Small sample scenario: complementary effects}
\begin{figure}[!htb]
		\centering
		\begin{subfigure}{.325\linewidth}
			\centering
			\includegraphics[height=4.35cm]{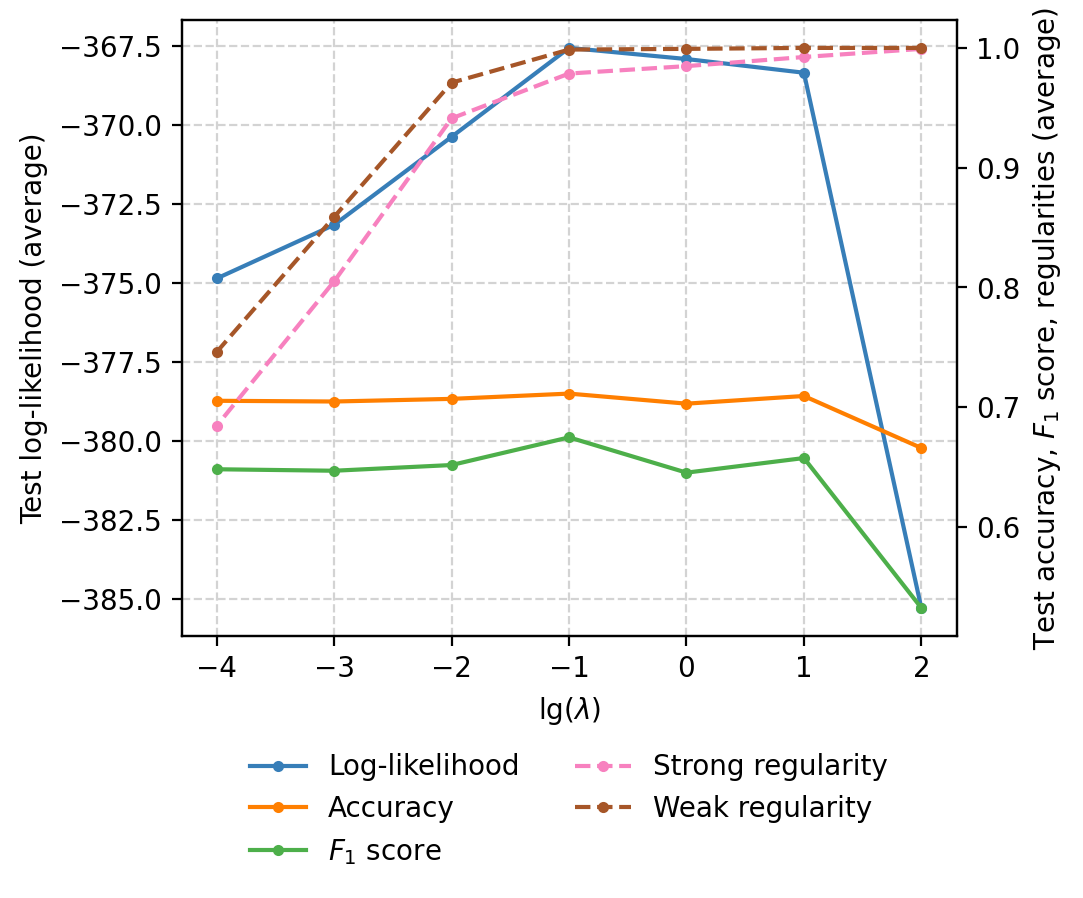}
			\subcaption{DNN, sum-PGR (CMAP)}
		\end{subfigure}
		\begin{subfigure}{.325\linewidth}
			\centering
			\includegraphics[height=4.35cm]{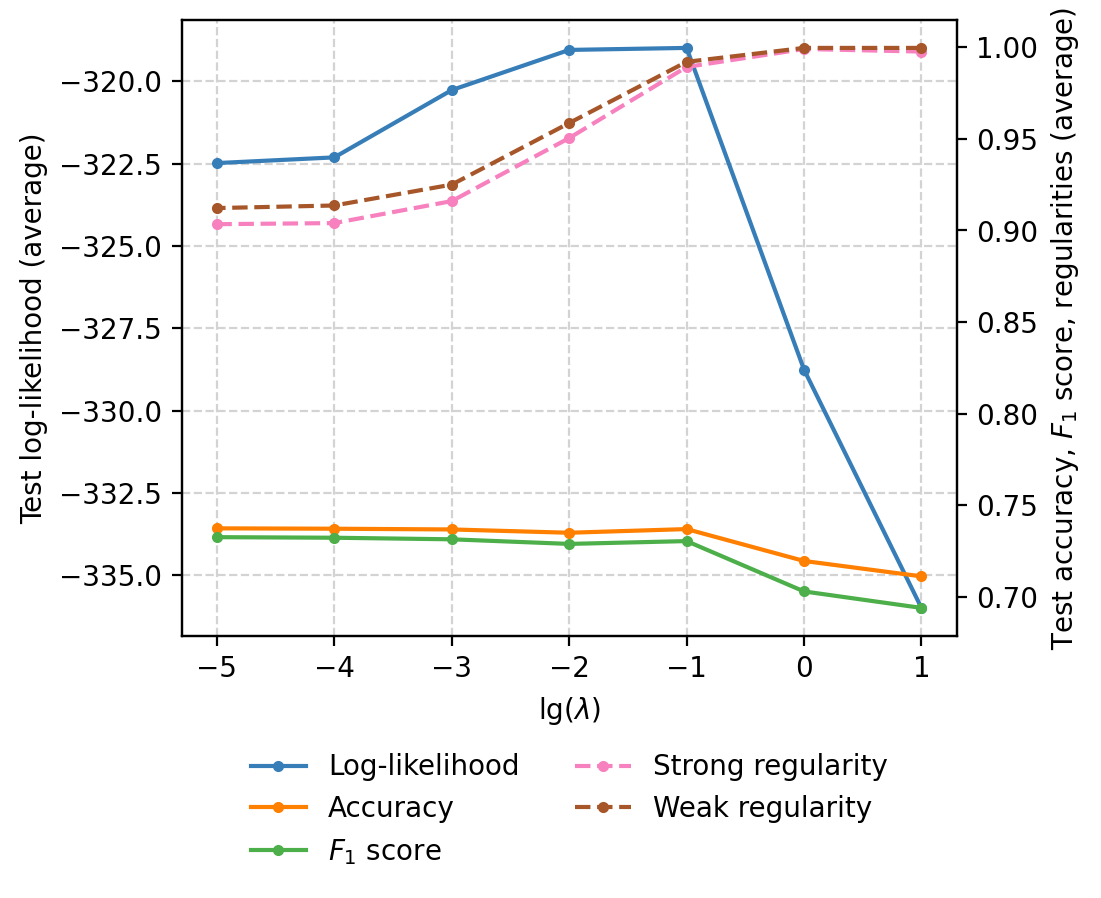}
			\subcaption{DNN, sum-PGR (LTDS)}
		\end{subfigure}
		\begin{subfigure}{.325\linewidth}
			\centering
			\includegraphics[height=4.35cm]{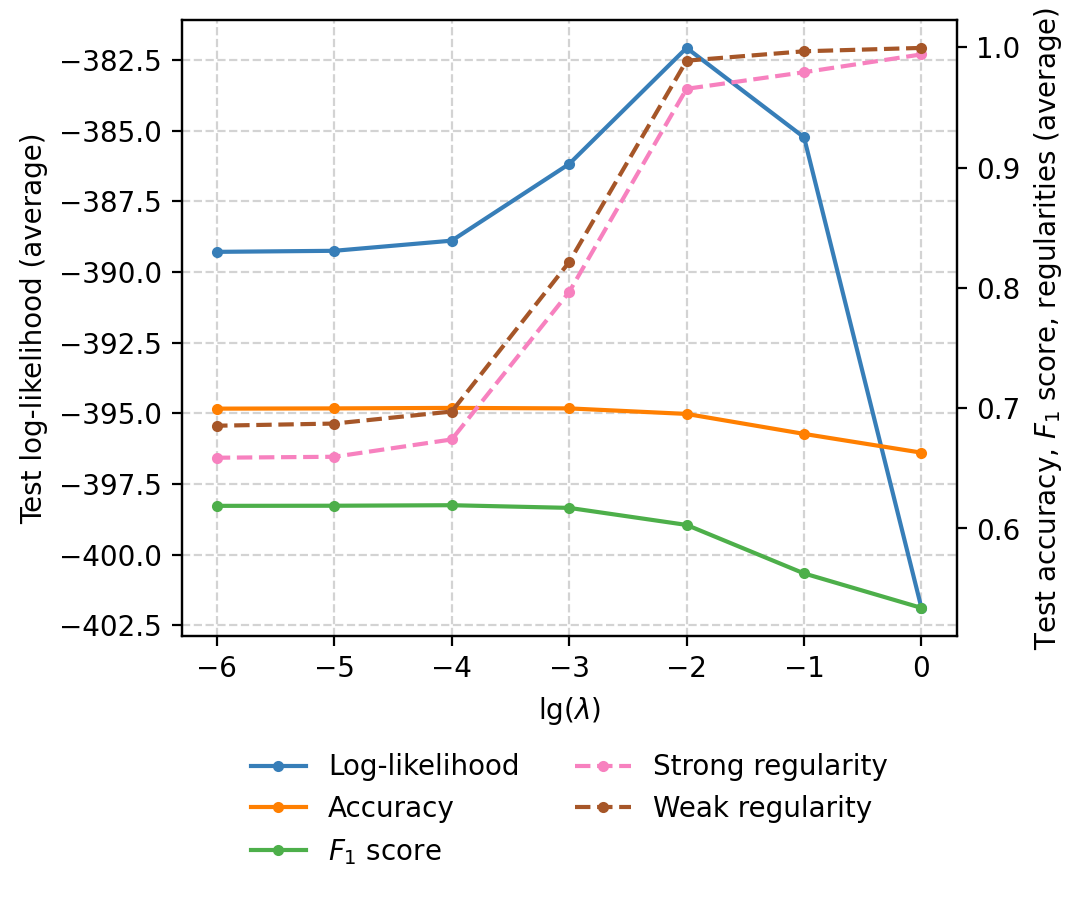}
			\subcaption{TasteNet, sum-PGR (CMAP)}
		\end{subfigure}
		\caption{Effects of regularization strength (1K-Random).}
		\label{fig:lambda_1k_random}
	\end{figure}

	Interestingly, under a setting with relatively small sample, stronger sum-based gradient regularization can simultaneously improve predictive power and behavioral regularity, thereby advancing the Pareto frontier of different performance metrics. \cref{fig:lambda_1k_random} illustrates the effects of $\lambda$ on the 1K-Random datasets, where we can find a range for each gradient regularizer such that predictive power and behavioral regularity increase together. In the CMAP case, when $\lambda$ increases from $10^{-4}$ to 0.1, sum-PGR improves the DNN's log-likelihood by 1.9\% of its absolute value and strong regularity by 29.4 percentage points. This phenomenon suggests the presence of Pareto efficiency in NN models, despite the substitution effects observed in \cref{fig:lambda_10k_random}.
	
	On the other hand, when $\lambda$ exceeds a certain critical point, we still observe substitution effects between predictive power and behavioral regularity for both DNNs and TasteNets. In the CMAP case, when $\lambda$ increases from 0.1 to 100, sum-UGR reduces the DNN's log-likelihood by 4.8\% of its absolute value, but still improves the DNN's strong regularity by 2.0 percentage points. This also demonstrates the flexibility of soft constraints by identifying the optimal $\lambda$, which reflects the alignment between the data and our behavioral assumptions.
	
	\subsubsection{Out-of-domain generalization}
\begin{figure}[!htb]
		\centering
		\begin{subfigure}{.325\linewidth}
			\centering
			\includegraphics[height=4.5cm]{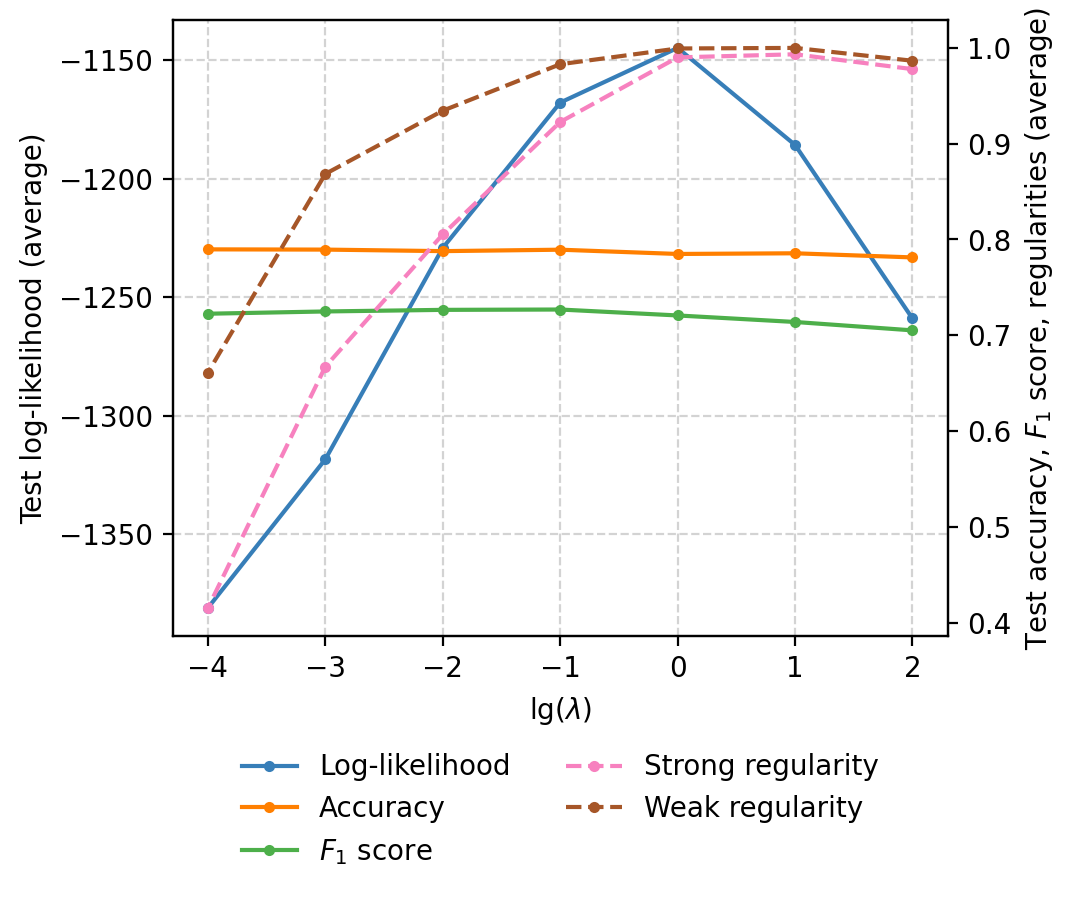}
			\subcaption{DNN, sum-UGR (CMAP)}
		\end{subfigure}
		\begin{subfigure}{.325\linewidth}
			\centering
			\includegraphics[height=4.5cm]{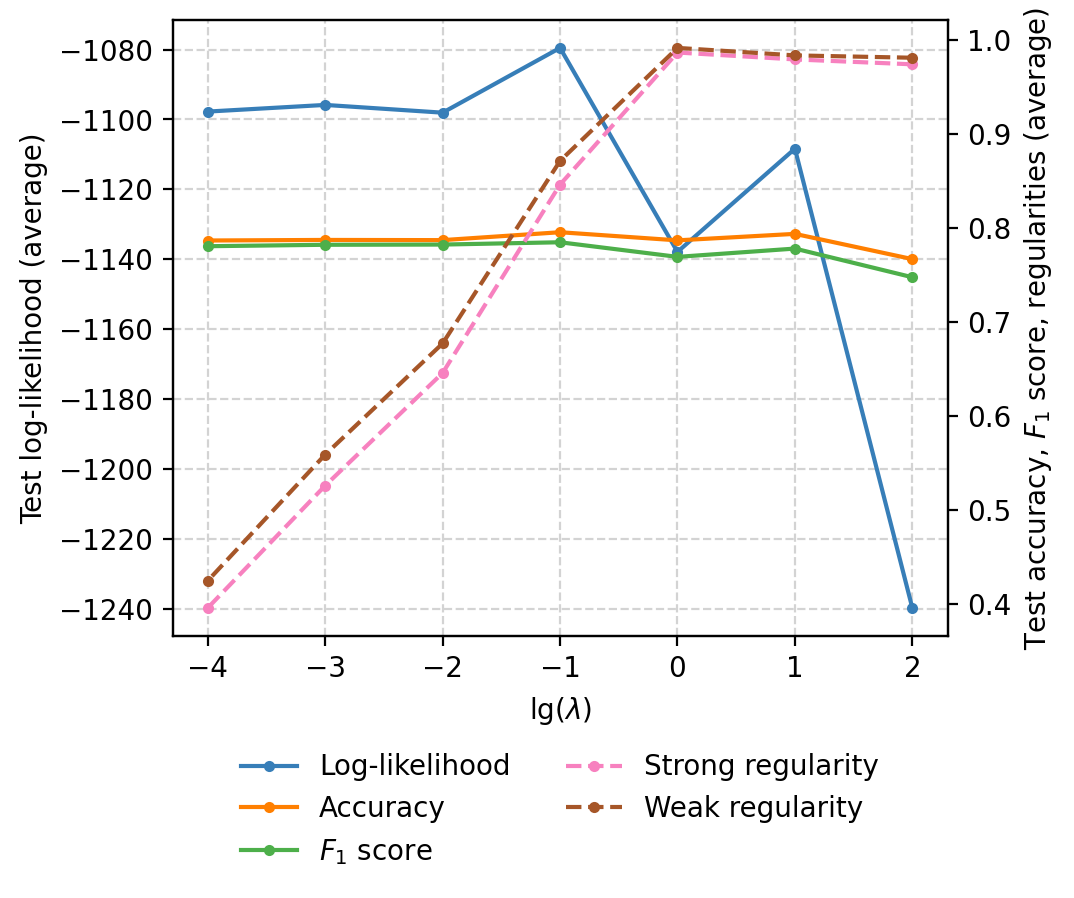}
			\subcaption{DNN, sum-UGR (LTDS)}
		\end{subfigure}
		\begin{subfigure}{.325\linewidth}
			\centering
			\includegraphics[height=4.5cm]{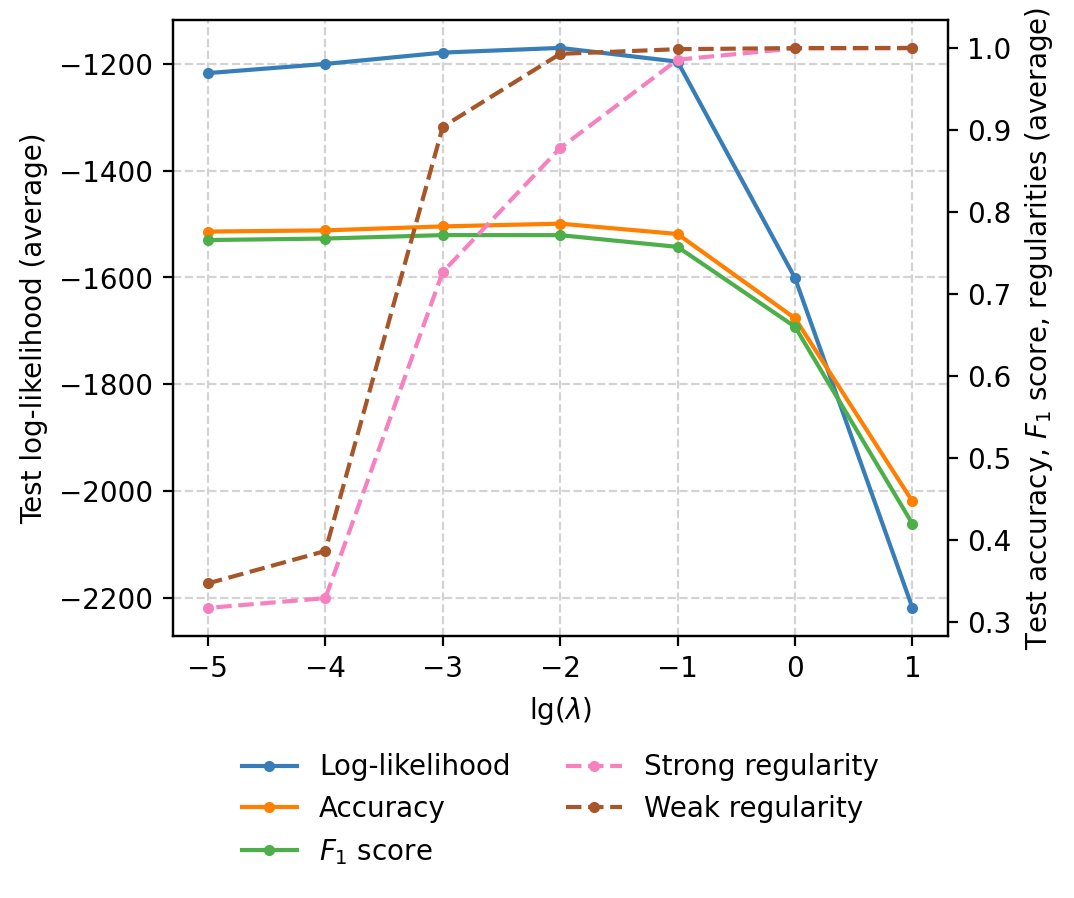}
			\subcaption{TasteNet, sum-PGR (LTDS)}
		\end{subfigure}
		\caption{Effects of regularization strength (10K-Sorted).}
		\label{fig:lambda_10k_sorted}
	\end{figure}

	To explore the out-of-domain generalizability of DNNs with sum-based gradient regularization, we visualize the effects of $\lambda$ for the 10K-Sorted datasets in \cref{fig:lambda_10k_sorted}. Interestingly, the trend of each metric combines the characteristics of the first two scenarios: we observe both complementary and substitution effects for DNNs, whereas substitution effects are more significant for TasteNets. In addition, the log-likelihood patterns for DNNs using the CMAP and LTDS data are slightly different (see \cref{fig:lambda_10k_sorted}a and b), suggesting the data-dependency nature of NN training. On the other hand, consistent with findings from previous sections, complementary effects and Pareto efficiency are observed for the sum-based gradient regularizers. These results imply the effectiveness of gradient regularization in improving behavioral regularity and predictive power simultaneously for the NN-based choice models.
	
	\subsection{Summary of empirical findings}
	\label{sec:summary_findings}
	As a summary, our findings demonstrate the conditions under which our sum-based gradient regularizers and behavioral regularity metrics are most effective. When the sample size is large, DNN models tend to have a high behavioral regularity. In this case, even a benchmark DNN architecture could be used for behavioral prediction and mobility policy analysis, and our behavioral regularity metrics and regularization methods might not be necessary. However, since it is always costly to collect travel survey data, the sample size in travel demand modeling is typically small (e.g., less than 10,000 samples). Under this context, sum-based gradient regularization should be broadly applied since it enables DNNs to generate reliable prediction and intuitive behavioral interpretation. Gradient regularization is also effective when the models are applied to future forecasting or cross-city transfer of mobility policies, which is emulated by the out-of-domain generalization scenario in our experiments. More generally, soft and hard constraints have their own advantages in improving behavioral regularity and predictive power, but neither is necessarily beneficial if the benchmark model already exhibits high behavioral regularity.
	
	\section{Conclusions}
	\label{6}
	
	DNNs often present behaviorally irregular patterns that greatly limit their practical use and theoretical appeal in travel behavior analysis, especially in applications and forecasting. Nevertheless, there is no consensus on how to measure or improve the model regularity of DNNs within the field of discrete choice modeling. This paper makes contributions by developing the behavioral regularity metrics and a gradient regularization framework. Specifically, we propose the law of demand in economics as a novel measure of DNNs' behavioral regularity w.r.t.\ generalized costs. Using a constrained optimization framework, we design six gradient regularizers to enhance the strong and weak behavioral regularities of DNNs. Empirically, these gradient regularizers are applied to two travel survey datasets collected from Chicago and London, through which we examine the trade-off between predictive power and behavioral regularity in the small versus large sample scenarios, and in-domain versus out-of-domain generalizations. Using five evaluation metrics, we demonstrate the effectiveness of our gradient regularizers on both DNNs and TasteNets.
	
	We find that sum-based gradient regularization can significantly improve the behavioral regularity of DNNs without sacrificing their predictive power in all the scenarios. There exists a substitution effect between predictive power and behavioral regularity in the large sample scenario, but a complementary effect in the small sample scenario. This is consistent with general understanding that, within the overparameterized regime, regularization can reduce variance \citep[e.g.,][]{alpaydin2014introduction}.\footnote{To be precise, this coincides with the classical viewpoint of bias--variance trade-off. Recent studies have found evidence that over-parameterization with large data samples could potentially improve model performance \citep[e.g.,][]{belkin2019reconciling}.} Utilizing the 10K-Sorted datasets, we further find that gradient regularization is also effective for out-of-domain generalization, which is critical for transferring knowledge across contexts. Our results also demonstrate how and why the gradient regularizers enhance predictive power and behavioral regularity, particularly for the small sample scenario and out-of-domain generalization. Comparing the soft and hard constraints, the DNNs with soft constraints can outperform the TasteNets with either soft or hard constraints, implying that TasteNets might be overly restrictive in capturing behavioral mechanisms. Besides the constrained optimization interpretation, the findings could be further understood by an analogy between gradient regularization and informative Bayesian prior. Specifically, gradient regularization can be seen as a deterministic prior imposed on the parameters estimates, that allows incorporating the modeler's prior belief. Such prior belief can be critical especially when the sample size is limited.
	
	This study pioneers in proposing new behavioral metrics and designing a practical regularization framework for DNNs. To address behavioral irregularity as shown in past studies \citep{wang2020deeparch, wang2020deep, wong2021reslogit, xia2023random}, we incorporate domain-knowledge into DNNs by regularizing the gradients' direction, in contrast to the norm-based (magnitude) regularization as in the computer science literature \citep{drucker1991double, jakubovitz2018improving, sokolic2017robust}. With this research, strong and weak behavioral regularities could be incorporated into future behavioral analysis using deep learning, thus evaluating the consistency of models with theories in behavioral science and microeconomics. Gradient regularization is a practical framework, which can impose any prior belief of the input--output relationship on any NN architecture. The gradient regularization method provides an alternative approach to enhance behavioral regularity, in addition to designing dedicated network architectures. It integrates the computational power of deep learning with common behavioral regularity assumptions through simple gradient regularization. Hence, it provides an opportunity to develop more plausible deep learning models for transportation policy analysis.	For example, researchers can use our approach to facilitate cross-city transfer of mobility policies when local governments seek to learn certain policies (e.g., congestion charging) from other areas. Our study demonstrates that deep learning with behavioral regularity could generate reliable insights for such policy transfer even when the source and the target cities have different data distributions in sociodemographics and pricing strategies.
	
	This study mainly adopts the market-level aggregate metric for behavioral regularity and enhances it through soft constraints, despite other possible approaches. Instead of the aggregate regularity, researchers may consider individual behavioral regularity and enhance it through hard constraints \citep{kim2024new}. We focus on the aggregate regularity because the metric is weaker than the individual one, thereby accommodating individual-level irrational behaviors \citep{bagwell1991high, stiving2000price}. Empirically, the aggregate regularity and soft constraints jointly lead to higher predictive performance, implying the existence of irregular behaviors for certain individuals or price ranges. In fact, the two issues -- individual versus aggregate regularity and soft versus hard constraints -- have been debated for a long time in the economics and optimization fields \citep{becker1962irrational}, and we avoid taking a strong stance on either side. Nevertheless, the aggregate regularity and soft constraints form the weakest assumption among all possibilities, thus serving as an acceptable common ground for the research community.
	
	Limitations still exist in this study. This work only regularizes direct partial derivatives, but not cross partial derivatives. Regularization on cross partial derivatives, on the other hand, could impose a stronger prior assumption, controlling the substitution and complementary patterns across alternatives. In addition, this research assumes exogenous penalty weight for gradient regularization, while future research could combine hyperparameter tuning to automatically learn these penalty weights. Lastly, as in typical deep learning models, parameter identification has been discussed \citep{hwang1997prediction} but largely remains an open question. In traditional DCMs, parameter identification is the foundation for statistical analysis \citep{mcfadden1980econometric}. However, without further research into DNNs' parameter identification, it could be challenging to quantify model uncertainty or design statistical tests, thus limiting the practicality of deep learning for discrete choice analysis. Future studies should investigate the necessary and sufficient conditions for model identification and statistical tests, especially in the over-parameterized deep learning models.
	
	\section*{CRediT authorship contribution statement}
	
	\textbf{Siqi Feng:} Writing -- review \& editing, Writing -- original draft, Visualization, Validation, Methodology, Investigation, Formal analysis, Data curation. \textbf{Rui Yao:} Writing -- review \& editing, Writing -- original draft, Validation, Methodology, Investigation, Conceptualization. \textbf{Stephane Hess:} Writing -- review \& editing. \textbf{Ricardo A. Daziano:} Writing -- review \& editing. \textbf{Timothy Brathwaite:} Writing -- review \& editing. \textbf{Joan Walker:} Writing -- review \& editing. \textbf{Shenhao Wang:} Writing -- review \& editing, Writing -- original draft, Validation, Supervision, Project administration, Methodology, Investigation, Formal analysis, Data curation, Conceptualization.
	
	\section*{Acknowledgments}
	
	We thank the three anonymous reviewers for their constructive criticism and valuable suggestions. We also thank Qingyi Wang, Yunhan Zheng, and Dingyi Zhuang from MIT for their helpful comments. Stephane Hess acknowledges support from the European Research Council through the advanced grant 101020940-SYNERGY.
	
	\begin{appendices}
		\crefalias{section}{appendix}
		\crefalias{subsection}{appendix}
		
		\section{Summary statistics of datasets}
		
		\subsection{Full datasets}
		See \cref{tab:stat_full,tab:stat_full_london}.
		
		\begin{table}[!htb]
			\centering\small
			\caption{Summary statistics of the full CMAP dataset.}
			\label{tab:stat_full}
			\begin{tabular}{l*{2}{S[table-format = 2.2]}S[table-format = 1.2]*{4}{S[table-format = 2.2]}}
				\toprule
				Variable & {Mean} & {SD} & {Min.} & {25\%} & {50\%} & {75\%} & {Max.} \\
				\midrule
				Age (year) & 38.95 & 13.30 & 6 & 29 & 37 & 47 & 84 \\
				Household size & 2.50 & 1.36 & 1 & 1 & 2 & 3 & 12 \\
				Number of cars in the household & 1.42 & 0.99 & 0 & 1 & 1 & 2 & 8 \\
				Time by car (h) & 0.22 & 0.16 & 0.00 & 0.11 & 0.17 & 0.29 & 1.66 \\
				Cost by car (\$) & 8.77 & 4.11 & 1.18 & 5.84 & 6.57 & 11.06 & 48.24 \\
				Time by train (h) & 0.82 & 0.67 & 0.05 & 0.38 & 0.64 & 0.99 & 4.95 \\
				Cost by train (\$) & 2.55 & 0.39 & 0.00 & 2.33 & 2.48 & 2.61 & 10.00 \\
				Time by active mobility (h) & 1.14 & 1.34 & 0.03 & 0.33 & 0.64 & 1.33 & 21.82 \\
				\midrule
				Male & \multicolumn{3}{l}{11,821 (1: yes);} & \multicolumn{3}{l}{14,278 (0: no)} \\
				Bachelor's or graduate degree & \multicolumn{3}{l}{18,853 (1: yes);} & \multicolumn{3}{l}{7246 (0: no)} \\
				One-person household & \multicolumn{3}{l}{6697 (1: yes);} & \multicolumn{3}{l}{19,402 (0: no)} \\
				One-car household & \multicolumn{3}{l}{10,004 (1: yes);} & \multicolumn{3}{l}{16,095 (0: no)} \\
				Household income $\ge$ \$75K & \multicolumn{3}{l}{15,520 (1: yes);} & \multicolumn{3}{l}{10,579 (0: no)} \\
				\bottomrule
			\end{tabular}
		\end{table}
		
		\begin{table}[H]
			\centering\small
			\caption{Summary statistics of the full LTDS dataset.}
			\label{tab:stat_full_london}
			\begin{tabular}{l*{6}{S[table-format = 1.2]}S[table-format = 2.2]}
				\toprule
				Variable & {Mean} & {SD} & {Min.} & {25\%} & {50\%} & {75\%} & {Max.} \\
				\midrule
				Number of cars in the household & 0.98 & 0.75 & 0 & 0 & 1 & 2 & 2 \\
				Number of public transit transfers & 0.37 & 0.62 & 0 & 0 & 0 & 1 & 4 \\
				Time by car (h) & 0.28 & 0.25 & 0.00 & 0.11 & 0.19 & 0.37 & 2.06 \\
				Cost by car (\pounds) & 1.90 & 3.49 & 0.00 & 0.29 & 0.57 & 1.29 & 17.16 \\
				Time by public transit (h) & 0.47 & 0.31 & 0.01 & 0.23 & 0.39 & 0.64 & 2.73 \\
				Cost by public transit (\pounds) & 1.56 & 1.54 & 0.00 & 0.00 & 1.50 & 2.40 & 13.49 \\
				Time by active mobility (h) & 0.75 & 0.73 & 0.02 & 0.23 & 0.48 & 1.00 & 5.98 \\
				\midrule
				Youth (age $<$ 25) & \multicolumn{3}{l}{18,917 (1: yes);} & \multicolumn{3}{l}{62,169 (0: no)} \\
				Senior (age $>$ 55) & \multicolumn{3}{l}{17,212 (1: yes);} & \multicolumn{3}{l}{63,874 (0: no)} \\
				Male & \multicolumn{3}{l}{38,396 (1: yes);} & \multicolumn{3}{l}{42,690 (0: no)} \\
				Driving license & \multicolumn{3}{l}{50,035 (1: yes);} & \multicolumn{3}{l}{31,051 (0: no)} \\
				\bottomrule
			\end{tabular}
		\end{table}
		
	%	\clearpage
		\subsection{10K-Sorted datasets}
		\label{sec:appendix_a.2}
		See \cref{tab:stat_10k_sorted_train,tab:stat_10k_sorted_test,tab:stat_10k_sorted_london_train,tab:stat_10k_sorted_london_test}.
		
		\begin{table}[!htb]
			\centering\small
			\caption{Summary statistics of the training set of 10K-Sorted CMAP.}
			\label{tab:stat_10k_sorted_train}
			\begin{tabular}{l*{7}{S[table-format = 2.2]}}
				\toprule
				Variable & {Mean} & {SD} & {Min.} & {25\%} & {50\%} & {75\%} & {Max.} \\
				\midrule
				Age (year) & 38.44 & 13.26 & 13 & 28 & 36 & 46 & 83 \\
				Household size & 2.49 & 1.36 & 1 & 1 & 2 & 3 & 8 \\
				Number of cars in the household & 1.37 & 1.01 & 0 & 1 & 1 & 2 & 8 \\
				Time by car (h) & 0.16 & 0.08 & 0.01 & 0.10 & 0.15 & 0.21 & 0.76 \\
				\rowcolor{gray!30}
				Cost by car (\$) & 6.98 & 1.86 & 1.33 & 5.73 & 6.57 & 7.76 & 11.41 \\
				Time by train (h) & 0.60 & 0.37 & 0.05 & 0.34 & 0.52 & 0.78 & 4.62 \\
				Cost by train (\$) & 2.41 & 0.23 & 0.00 & 2.31 & 2.41 & 2.52 & 10.00 \\
				Time by active mobility (h) & 0.60 & 0.40 & 0.04 & 0.28 & 0.51 & 0.84 & 5.93 \\
				\midrule
				Male & \multicolumn{3}{l}{3104 (1: yes);} & \multicolumn{3}{l}{3896 (0: no)} \\
				Bachelor's or graduate degree & \multicolumn{3}{l}{5087 (1: yes);} & \multicolumn{3}{l}{1913 (0: no)} \\
				One-person household & \multicolumn{3}{l}{1830 (1: yes);} & \multicolumn{3}{l}{5170 (0: no)} \\
				One-car household & \multicolumn{3}{l}{2678 (1: yes);} & \multicolumn{3}{l}{4322 (0: no)} \\
				Household income $\ge$ \$75K & \multicolumn{3}{l}{4079 (1: yes);} & \multicolumn{3}{l}{2921 (0: no)} \\
				\bottomrule
			\end{tabular}
		\end{table}
		
		\begin{table}[!htb]
			\centering\small
			\caption{Summary statistics of the test set of 10K-Sorted CMAP.}
			\label{tab:stat_10k_sorted_test}
			\begin{tabular}{l*{7}{S[table-format = 2.2]}}
				\toprule
				Variable & {Mean} & {SD} & {Min.} & {25\%} & {50\%} & {75\%} & {Max.} \\
				\midrule
				Age (year) & 41.30 & 13.62 & 6 & 31 & 39 & 50 & 84 \\
				Household size & 2.61 & 1.35 & 1 & 2 & 2 & 4 & 8 \\
				Number of cars in the household & 1.67 & 0.96 & 0 & 1 & 2 & 2 & 7 \\
				Time by car (h) & 0.46 & 0.15 & 0.09 & 0.34 & 0.44 & 0.55 & 1.51 \\
				\rowcolor{gray!30}
				Cost by car (\$) & 15.95 & 2.33 & 11.41 & 14.44 & 16.54 & 17.44 & 48.24 \\
				Time by train (h) & 1.77 & 0.81 & 0.22 & 1.16 & 1.59 & 2.18 & 4.94 \\
				Cost by train (\$) & 3.10 & 0.41 & 2.00 & 2.83 & 3.15 & 3.33 & 6.65 \\
				Time by active mobility (h) & 3.22 & 1.58 & 0.16 & 2.03 & 2.76 & 4.03 & 18.37 \\
				\midrule
				Male & \multicolumn{3}{l}{963 (1: yes);} & \multicolumn{3}{l}{1037 (0: no)} \\
				Bachelor's or graduate degree & \multicolumn{3}{l}{1439 (1: yes);} & \multicolumn{3}{l}{561 (0: no)} \\
				One-person household & \multicolumn{3}{l}{436 (1: yes);} & \multicolumn{3}{l}{1564 (0: no)} \\
				One-car household & \multicolumn{3}{l}{669 (1: yes);} & \multicolumn{3}{l}{1331 (0: no)} \\
				Household income $\ge$ \$75K & \multicolumn{3}{l}{1309 (1: yes;} & \multicolumn{3}{l}{691 (0: no)} \\
				\bottomrule
			\end{tabular}
		\end{table}
		
		\begin{table}[!htb]
			\centering\small
			\caption{Summary statistics of the training set of 10K-Sorted LTDS.}
			\label{tab:stat_10k_sorted_london_train}
			\begin{tabular}{l*{6}{S[table-format = 1.2]}S[table-format = 2.2]}
				\toprule
				Variable & {Mean} & {SD} & {Min.} & {25\%} & {50\%} & {75\%} & {Max.} \\
				\midrule
				Number of cars in the household & 0.98 & 0.75 & 0 & 0 & 1 & 2 & 2 \\
				Number of public transit transfers & 0.16 & 0.43 & 0 & 0 & 0 & 0 & 3 \\
				Time by car (h) & 0.21 & 0.18 & 0.01 & 0.09 & 0.15 & 0.27 & 1.45 \\
				Cost by car (\pounds) & 1.44 & 3.00 & 0.03 & 0.25 & 0.44 & 0.87 & 16.11 \\
				Time by public transit (h) & 0.38 & 0.25 & 0.02 & 0.20 & 0.32 & 0.50 & 2.00 \\
				\rowcolor{gray!30}
				Cost by public transit (\pounds) & 0.98 & 0.85 & 0.00 & 0.00 & 1.50 & 1.50 & 2.90 \\
				Time by active mobility & 0.54 & 0.52 & 0.02 & 0.20 & 0.36 & 0.69 & 5.67 \\
				\midrule
				Youth (age $<$ 25) & \multicolumn{3}{l}{1746 (1: yes);} & \multicolumn{3}{l}{5254 (0: no)} \\
				Senior (age $>$ 55) & \multicolumn{3}{l}{1729 (1: yes);} & \multicolumn{3}{l}{5271 (0: no)}  \\
				Male & \multicolumn{3}{l}{3296 (1: yes);} & \multicolumn{3}{l}{3704 (0: no)} \\
				Driving license & \multicolumn{3}{l}{4140 (1: yes);} & \multicolumn{3}{l}{2860 (0: no)} \\
				\bottomrule
			\end{tabular}
		\end{table}
		
		\begin{table}[!htb]
			\centering\small
			\caption{Summary statistics of the test set of 10K-Sorted LTDS.}
			\label{tab:stat_10k_sorted_london_test}
			\begin{tabular}{l*{6}{S[table-format = 1.2]}S[table-format = 2.2]}
				\toprule
				Variable & {Mean} & {SD} & {Min.} & {25\%} & {50\%} & {75\%} & {Max.} \\
				\midrule
				Number of cars in the household & 1.04 & 0.72 & 0 & 1 & 1 & 2 & 2 \\
				Number of public transit transfers & 1.20 & 0.57 & 0 & 1 & 1 & 1 & 4 \\
				Time by car (h) & 0.55 & 0.30 & 0.07 & 0.31 & 0.49 & 0.75 & 1.58 \\
				Cost by car (\pounds) & 3.89 & 4.64 & 0.21 & 0.97 & 1.63 & 3.47 & 16.28 \\
				Time by public transit (h) & 0.82 & 0.29 & 0.22 & 0.61 & 0.78 & 0.98 & 2.73 \\
				\rowcolor{gray!30}
				Cost by public transit (\pounds) & 3.97 & 1.28 & 2.90 & 3.00 & 3.40 & 4.50 & 11.60 \\
				Time by active mobility & 1.55 & 0.86 & 0.22 & 0.87 & 1.36 & 2.05 & 4.87 \\
				\midrule
				Youth (age $<$ 25) & \multicolumn{3}{l}{212 (1: yes);} & \multicolumn{3}{l}{1788 (0: no)} \\
				Senior (age $>$ 55) & \multicolumn{3}{l}{152 (1: yes);} & \multicolumn{3}{l}{1848 (0: no)} \\
				Male & \multicolumn{3}{l}{1028 (1: yes);} & \multicolumn{3}{l}{972 (0: no)} \\
				Driving license & \multicolumn{3}{l}{1561 (1: yes);} & \multicolumn{3}{l}{439 (0: no)} \\
				\bottomrule
			\end{tabular}
		\end{table}
		
		\clearpage
		\section{Impacts of optimization algorithms}
		\label{sec:appendix_b}
		
\begin{figure}[!htb]
			\centering
			\begin{subfigure}{.325\linewidth}
				\includegraphics[width=\linewidth]{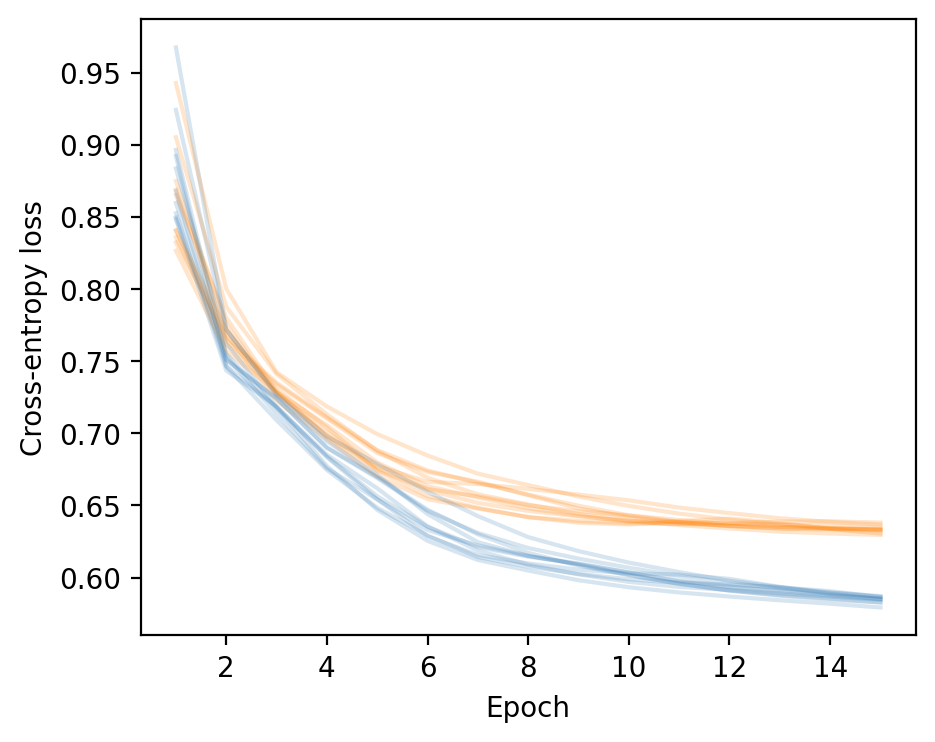}
				\subcaption{Adam}
			\end{subfigure}
			\begin{subfigure}{.325\linewidth}
				\includegraphics[width=\linewidth]{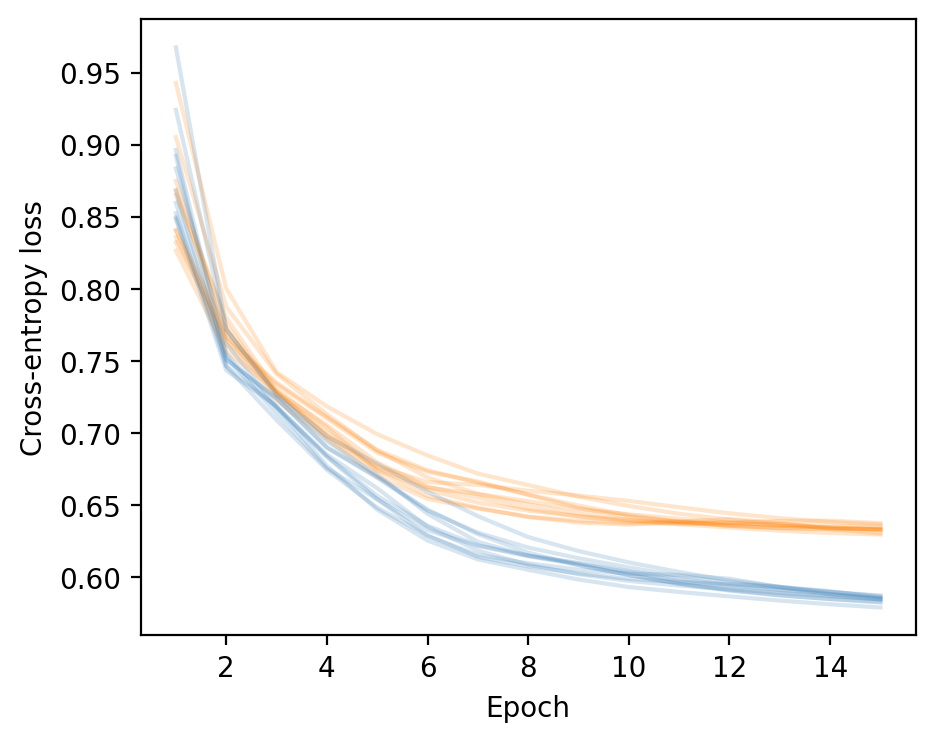}
				\subcaption{AdamW}
			\end{subfigure}
			\begin{subfigure}{.325\linewidth}
				\includegraphics[width=\linewidth]{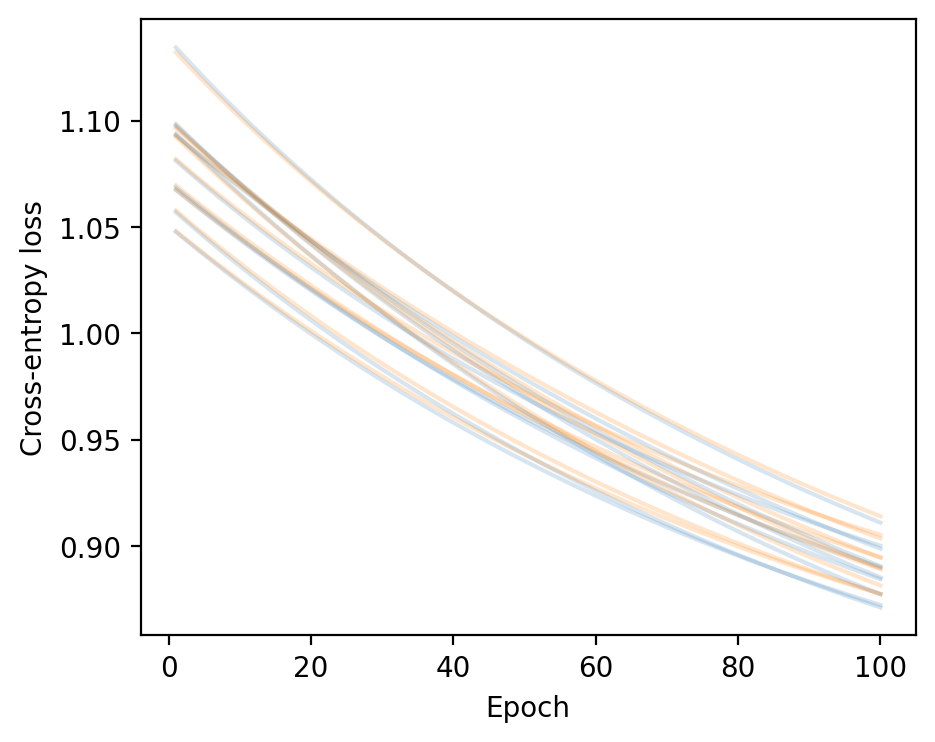}
				\subcaption{SGD}
			\end{subfigure}
			\caption{Training loss (blue) and validation loss (orange) per epoch for different algorithms.}
			\label{fig:losses_per_epoch}
		\end{figure}

\begin{figure}[!htb]
			\centering
			\begin{subfigure}{.325\linewidth}
				\centering
				\includegraphics[height=4.2cm]{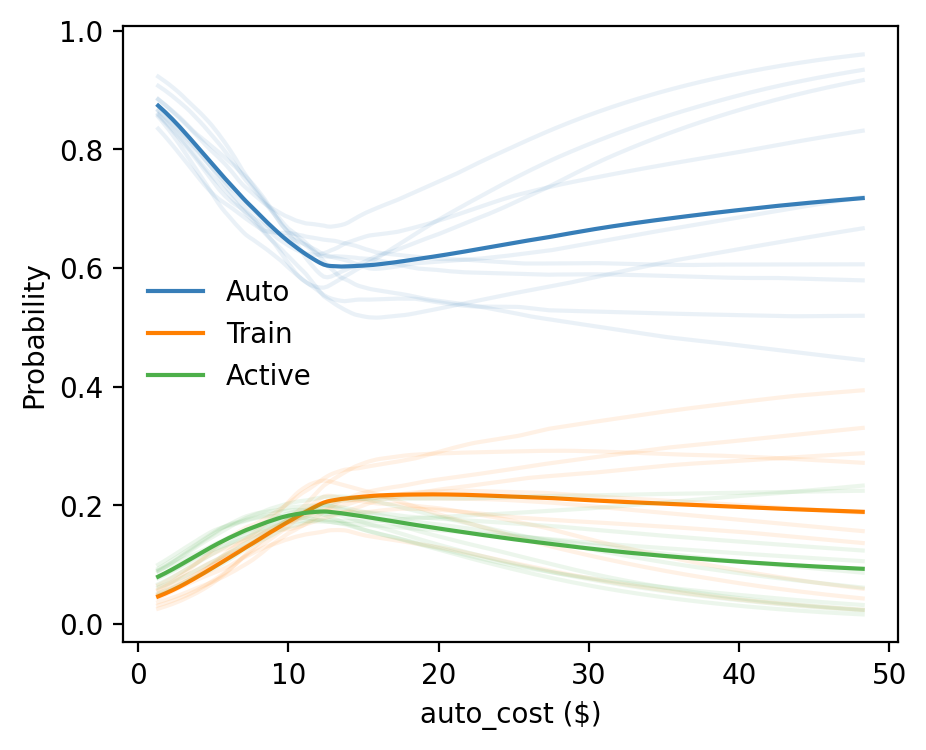}
				\subcaption{Adam}
			\end{subfigure}
			\begin{subfigure}{.325\linewidth}
				\centering
				\includegraphics[height=4.2cm]{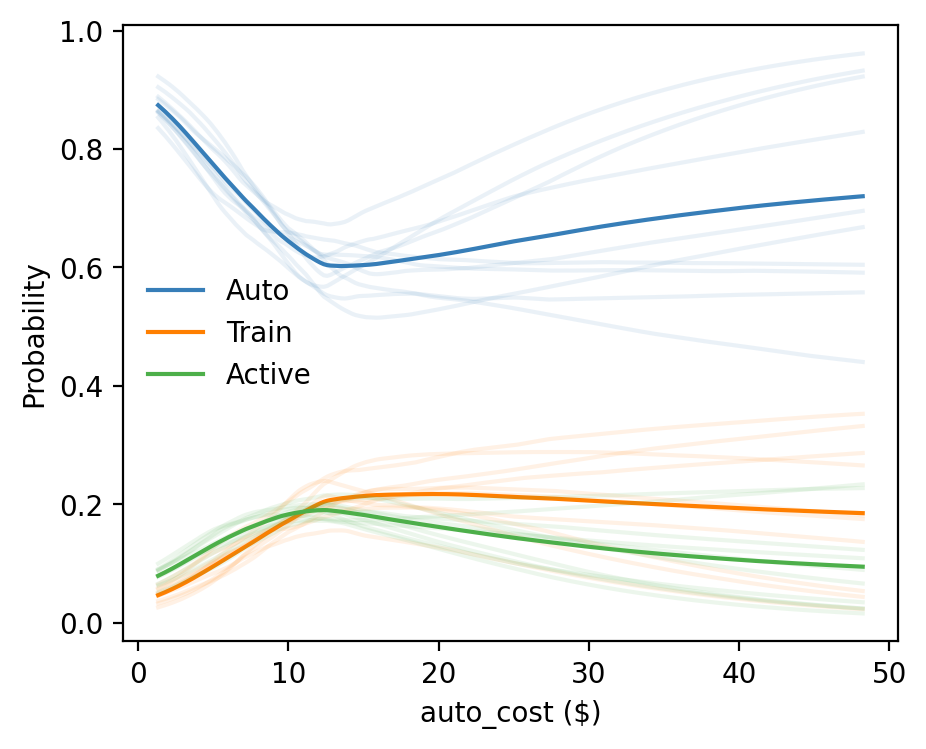}
				\subcaption{AdamW}
			\end{subfigure}
			\begin{subfigure}{.325\linewidth}
				\centering
				\includegraphics[height=4.2cm]{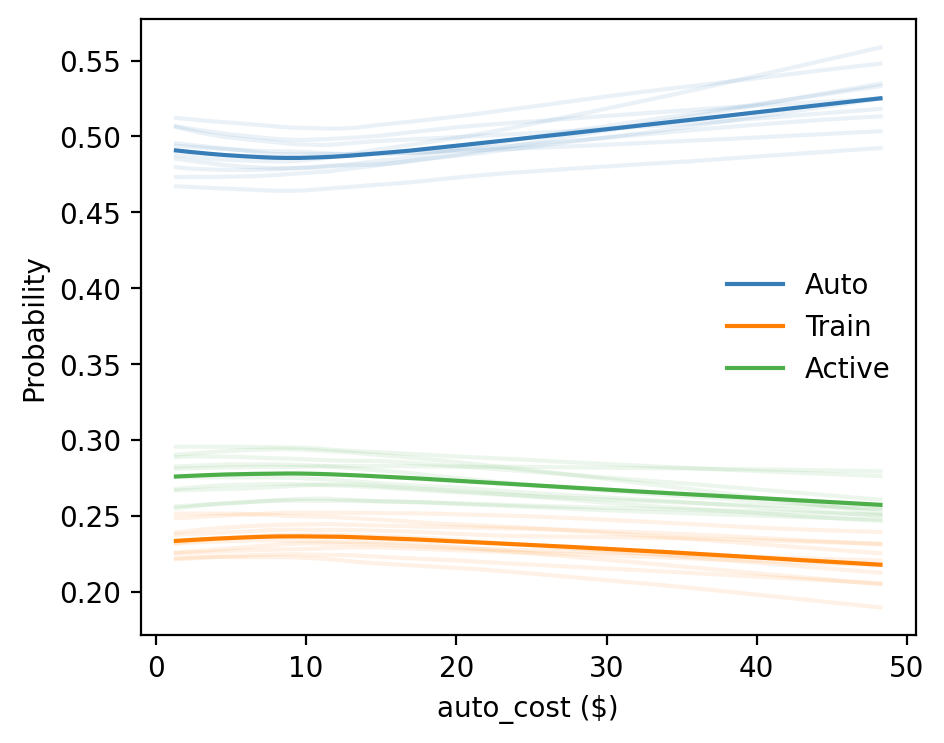}
				\subcaption{SGD}
			\end{subfigure}
			\caption{Individual demands as functions of driving costs for different algorithms.}
			\label{fig:sub_algorithms}
		\end{figure}

		Using the 10K-Random CMAP dataset, we visualize the training and validation losses (\cref{fig:losses_per_epoch}) as well as the corresponding individual demand functions (\cref{fig:sub_algorithms}) for the Adam, AdamW, and SGD algorithms. Based on default settings of all three algorithms, both Adam and AdamW lead to fast convergence and reasonable demand functions, whereas SGD does not converge within 100 epochs and leads to less reasonable demand functions. It is worth noting that weight decay of AdamW is set as zero to avoid further complexity in regularization, thus identifying the direct impacts of our gradient regularization on behavioral regularity.
		
		\clearpage
		\section{Training and validation performance}
		\label{sec:appendix_c}
		
		\subsection{Large sample scenario}
		\label{sec:appendix_c.1}
		See \cref{tab:10k_random_train,tab:10k_random_val}.
		
		\begin{table}[!htb]
			\caption{Model performance in the training sets of 10K-Random.}
			\label{tab:10k_random_train}
			\resizebox{\linewidth}{!}{
				\begin{tabular}{l|cccc|cccccc|c}
					\toprule
					Metric: & \multicolumn{4}{c|}{DNN} & \multicolumn{6}{c|}{TasteNet} & RUM \\
					Mean (SD) & No GR & PGR & UGR & LGR & No GR & PGR & UGR & LGR & ReLU & Exp & MNL \\
					\midrule
					\multicolumn{12}{l}{Panel 1: CMAP data, sum-XGR} \\
					\midrule
					Log-likelihood & $-4272.7$ & $\mathbf{-4216.5}$ & $-4240.1$ & $\underline{-4221.9}$ & $-4677.8$ & $-4675.8$ & $-4676.0$ & $-4675.7$ & $-4737.2$ & $-4675.7$ & $-4710.2$ \\
					& (12.97) & (11.74) & (15.40) & (17.58) & (19.89) & (19.38) & (19.57) & (19.24) & (51.18) & (18.97) & (0) \\
					Accuracy & 0.751 & \textbf{0.753} & \underline{0.752} & \textbf{0.753} & 0.732 & 0.732 & 0.733 & 0.732 & 0.729 & 0.733 & 0.731 \\
					& (0.002) & (0.002) & (0.002) & (0.002) & (0.002) & (0.002) & (0.002) & (0.002) & (0.002) & (0.002) & (0) \\
					$F_1$ score & 0.717 & \textbf{0.726} & 0.721 & \underline{0.724} & 0.675 & 0.675 & 0.675 & 0.675 & 0.671 & 0.681 & 0.681 \\
					& (0.004) & (0.004) & (0.003) & (0.004) & (0.004) & (0.004) & (0.004) & (0.004) & (0.004) & (0.003) & (0) \\
					Strong regularity & 0.889 & 0.987 & 0.980 & 0.988 & 0.997 & \underline{0.998} & \underline{0.998} & \underline{0.998} & 0.423 & \textbf{0.999} & \underline{0.998} \\
					& (0.061) & (0.003) & (0.011) & (0.003) & (0.003) & (0.002) & (0.002) & (0.002) & (0.315) & (0.000) & (0) \\
					Weak regularity & 0.927 & \underline{0.999} & 0.997 & \underline{0.999} & \underline{0.999} & \underline{0.999} & \underline{0.999} & \underline{0.999} & \textbf{1.000} & \textbf{1.000} & \textbf{1.000} \\
					& (0.058) & (0.001) & (0.004) & (0.001) & (0.002) & (0.001) & (0.002) & (0.001) & (0.000) & (0.000) & (0) \\
					\midrule
					\multicolumn{12}{l}{Panel 2: CMAP data, norm-XGR} \\
					\midrule
					Log-likelihood & $\mathbf{-4272.7}$ & $\underline{-4282.9}$ & $-4333.2$ & $-4303.7$ & $-4677.8$ & $-4679.7$ & $-4763.8$ & $-4684.8$ & $-4737.2$ & $-4675.7$ & $-4710.2$ \\
					& (13.22) & (12.23) & (14.33) & (11.96) & (19.89) & (19.88) & (17.51) & (19.68) & (51.18) & (18.97) & (0) \\
					Accuracy & \textbf{0.751} & \underline{0.750} & 0.745 & 0.748 & 0.732 & 0.732 & 0.729 & 0.732 & 0.729 & 0.733 & 0.731 \\
					& (0.002) & (0.002) & (0.002) & (0.001) & (0.002) & (0.002) & (0.002) & (0.002) & (0.002) & (0.002) & (0) \\
					$F_1$ score & \textbf{0.717} & \underline{0.713} & 0.700 & 0.706 & 0.675 & 0.674 & 0.665 & 0.673 & 0.671 & 0.681 & 0.681 \\
					& (0.004) & (0.004) & (0.005) & (0.005) & (0.004) & (0.004) & (0.004) & (0.004) & (0.004) & (0.003) & (0) \\
					Strong regularity & 0.889 & 0.856 & 0.704 & 0.813 & 0.997 & 0.997 & 0.922 & 0.996 & 0.423 & \textbf{0.999} & \underline{0.998} \\
					& (0.061) & (0.063) & (0.044) & (0.063) & (0.003) & (0.003) & (0.032) & (0.005) & (0.315) & (0.000) & (0) \\
					Weak regularity & 0.927 & 0.898 & 0.761 & 0.855 & \underline{0.999} & \underline{0.999} & 0.935 & 0.998 & \textbf{1.000} & \textbf{1.000} & \textbf{1.000} \\
					& (0.058) & (0.062) & (0.046) & (0.065) & (0.002) & (0.002) & (0.028) & (0.003) & 1.000 & 1.000 & (0) \\
					\midrule
					\multicolumn{12}{l}{Panel 3: LTDS data, sum-XGR} \\
					\midrule
					Log-likelihood & $\underline{-4427.9}$ & $\mathbf{-4416.1}$ & $-4434.8$ & $-4499.3$ & $-4500.2$ & $-4517.9$ & $-4567.1$ & $-4518.7$ & $-4589.4$ & $-4498.1$ & $-4834.7$ \\
					& (20.61) & (39.09) & (24.41) & (16.00) & (10.52) & (12.03) & (24.14) & (11.93) & (68.97) & (9.146) & (0) \\
					Accuracy & 0.732 & 0.729 & 0.731 & 0.729 & \underline{0.736} & 0.734 & 0.728 & 0.734 & 0.728 & \textbf{0.738} & 0.718 \\
					& (0.003) & (0.004) & (0.005) & (0.004) & (0.001) & (0.002) & (0.004) & (0.002) & (0.005) & (0.001) & (0) \\
					$F_1$ score & 0.731 & 0.728 & 0.730 & 0.726 & \underline{0.732} & 0.730 & 0.723 & 0.730 & 0.724 & \textbf{0.734} & 0.714 \\
					& (0.002) & (0.003) & (0.004) & (0.004) & (0.001) & (0.002) & (0.004) & (0.002) & (0.005) & (0.001) & (0) \\
					Strong regularity & 0.947 & 0.993 & 0.994 & \textbf{0.997} & 0.936 & 0.964 & 0.988 & 0.975 & 0.927 & 0.990 & \underline{0.996} \\
					& (0.032) & (0.004) & (0.009) & (0.003) & (0.023) & (0.012) & (0.009) & (0.011) & (0.046) & (0.002) & (0) \\
					Weak regularity & 0.968 & \underline{0.999} & 0.998 & \textbf{1.000} & 0.960 & \textbf{1.000} & \textbf{1.000} & \textbf{1.000} & \textbf{1.000} & \textbf{1.000} & \textbf{1.000} \\
					& (0.025) & (0.001) & (0.005) & (0.000) & (0.019) & (0.000) & (0.000) & (0.000) & (0.000) & (0.000) & (0) \\
					\bottomrule
				\end{tabular}
			}
		\end{table}
		
		\begin{table}
			\caption{Model performance in the validation sets of 10K-Random.}
			\label{tab:10k_random_val}
			\resizebox{\linewidth}{!}{
				\begin{tabular}{l|cccc|cccccc|c}
					\toprule
					Metric: & \multicolumn{4}{c|}{DNN} & \multicolumn{6}{c|}{TasteNet} & RUM \\
					Mean (SD) & No GR & PGR & UGR & LGR & No GR & PGR & UGR & LGR & ReLU & Exp & MNL \\
					\midrule
					\multicolumn{12}{l}{Panel 1: CMAP data, sum-XGR} \\
					\midrule
					Log-likelihood & $\underline{-633.6}$ & $\mathbf{-633.0}$ & $-634.1$ & $-634.2$ & $-683.0$ & $-683.1$ & $-683.1$ & $-683.1$ & $-689.2$ & $-687.1$ & $-693.8$ \\
					& (2.507) & (2.474) & (3.594) & (2.794) & (2.442) & (2.470) & (2.482) & (2.467) & (6.598) & (2.722) & (0) \\
					Accuracy & \textbf{0.739} & \textbf{0.739} & \textbf{0.739} & \textbf{0.739} & 0.726 & 0.727 & 0.727 & 0.727 & 0.724 & 0.728 & \underline{0.737} \\
					& (0.004) & (0.004) & (0.003) & (0.003) & (0.004) & (0.003) & (0.003) & (0.003) & (0.003) & (0.004) & (0) \\
					$F_1$ score & 0.702 & \textbf{0.710} & 0.705 & \underline{0.708} & 0.668 & 0.669 & 0.670 & 0.669 & 0.664 & 0.679 & 0.690 \\
					& (0.005) & (0.004) & (0.005) & (0.003) & (0.007) & (0.006) & (0.006) & (0.006) & (0.003) & (0.004) & (0) \\
					Strong regularity & 0.879 & 0.981 & 0.973 & 0.982 & 0.997 & \underline{0.999} & 0.998 & 0.998 & 0.427 & \textbf{1.000} & \textbf{1.000} \\
					& (0.058) & (0.005) & (0.014) & (0.005) & (0.005) & (0.002) & (0.003) & (0.002) & (0.319) & (0.001) & (0) \\
					Weak regularity & 0.928 & \underline{0.999} & 0.997 & \underline{0.999} & 0.998 & \underline{0.999} & \underline{0.999} & \underline{0.999} & \textbf{1.000} & \textbf{1.000} & \textbf{1.000} \\
					& (0.057) & (0.001) & (0.004) & (0.001) & (0.003) & (0.002) & (0.002) & (0.002) & (0.000) & (0.000) & (0) \\
					\midrule
					\multicolumn{12}{l}{Panel 2: CMAP data, norm-XGR} \\
					\midrule
					Log-likelihood & $\underline{-633.6}$ & $\mathbf{-633.3}$ & $-635.9$ & $\mathbf{-633.3}$ & $-683.0$ & $-683.0$ & $-687.4$ & $-682.9$ & $-689.2$ & $-687.1$ & $-693.8$ \\
					& (2.507) & (2.414) & (2.676) & (2.989) & (2.442) & (2.421) & (2.236) & (2.366) & (6.598) & (2.722) & (0) \\
					Accuracy & \underline{0.739} & \textbf{0.741} & \textbf{0.741} & \textbf{0.741} & 0.726 & 0.725 & 0.720 & 0.725 & 0.724 & 0.728 & 0.737 \\
					& (0.004) & (0.004) & (0.003) & (0.003) & (0.004) & (0.004) & (0.001) & (0.004) & (0.003) & (0.004) & (0) \\
					$F_1$ score & \textbf{0.702} & \textbf{0.702} & 0.696 & \underline{0.699} & 0.668 & 0.666 & 0.655 & 0.665 & 0.664 & 0.679 & 0.690 \\
					& (0.005) & (0.005) & (0.006) & (0.005) & (0.007) & (0.007) & (0.003) & (0.006) & (0.003) & (0.004) & (0) \\
					Strong regularity & 0.879 & 0.846 & 0.701 & 0.803 & \underline{0.997} & 0.996 & 0.924 & 0.995 & 0.427 & \textbf{1.000} & \textbf{1.000} \\
					& (0.058) & (0.058) & (0.044) & (0.061) & (0.005) & (0.005) & (0.03) & (0.007) & (0.319) & (0.001) & (0) \\
					Weak regularity & 0.928 & 0.900 & 0.766 & 0.858 & \underline{0.998} & \underline{0.998} & 0.935 & 0.997 & \textbf{1.000} & \textbf{1.000} & \textbf{1.000} \\
					& (0.057) & (0.057) & (0.046) & (0.062) & (0.003) & (0.003) & (0.026) & (0.006) & (0.000) & (0.000) & (0) \\
					\midrule
					\multicolumn{12}{l}{Panel 3: LTDS data, sum-XGR} \\
					\midrule
					Log-likelihood & $-645.5$ & $-642.8$ & $-644.2$ & $-647.8$ & $\mathbf{-638.8}$ & $\underline{-639.9}$ & $-645.3$ & $\underline{-639.9}$ & $-654.1$ & $-640.6$ & $-670.0$ \\
					& (4.399) & (5.709) & (4.345) & (3.915) & (1.949) & (2.182) & (3.254) & (1.985) & (11.75) & (2.157) & (0) \\
					Accuracy & 0.721 & 0.722 & 0.725 & 0.723 & \underline{0.729} & 0.726 & 0.726 & 0.726 & 0.720 & 0.726 & \textbf{0.735} \\
					& (0.002) & (0.003) & (0.002) & (0.005) & (0.003) & (0.002) & (0.002) & (0.002) & (0.006) & (0.003) & (0) \\
					$F_1$ score & 0.719 & 0.721 & 0.722 & 0.718 & \underline{0.723} & 0.720 & 0.719 & 0.720 & 0.714 & 0.721 & \textbf{0.730} \\
					& (0.003) & (0.003) & (0.002) & (0.004) & (0.003) & (0.002) & (0.003) & (0.002) & (0.007) & (0.003) & (0) \\
					Strong regularity & 0.947 & \underline{0.992} & 0.991 & \textbf{0.996} & 0.924 & 0.953 & 0.982 & 0.964 & 0.916 & 0.982 & 0.991 \\
					& (0.032) & (0.005) & (0.012) & (0.004) & (0.026) & (0.014) & (0.011) & (0.014) & (0.050) & (0.003) & (0) \\
					Weak regularity & 0.971 & \underline{0.999} & 0.997 & \textbf{1.000} & 0.957 & \textbf{1.000} & \textbf{1.000} & \textbf{1.000} & \textbf{1.000} & \textbf{1.000} & \textbf{1.000} \\
					& (0.025) & (0.001) & (0.007) & (0.001) & (0.023) & (0.001) & (0.000) & (0.001) & (0.000) & (0.000) & (0) \\
					\bottomrule
				\end{tabular}
			}
		\end{table}
		
		\clearpage
		\subsection{Small sample scenario}
		\label{sec:appendix_c.2}
		See \cref{tab:1k_random_train,tab:1k_random_val}.
		
		\begin{table}[!htb]
			\centering
			\caption{Model performance in the training sets of 1K-Random.}
			\label{tab:1k_random_train}
			\resizebox{.97\linewidth}{!}{
				\begin{tabular}{l|cccc|cccccc|c}
					\toprule
					Metric: & \multicolumn{4}{c|}{DNN} & \multicolumn{6}{c|}{TasteNet} & RUM \\
					Mean (SD) & No GR & PGR & UGR & LGR & No GR & PGR & UGR & LGR & ReLU & Exp & MNL \\
					\midrule
					\multicolumn{12}{l}{Panel 1: CMAP data, sum-XGR} \\
					\midrule
					Log-likelihood & $\mathbf{-494.8}$ & $\underline{-511.0}$ & $-546.4$ & $-526.2$ & $-555.6$ & $-580.2$ & $-570.9$ & $-583.1$ & $-563.0$ & $-558.1$ & $-548.1$ \\
					& (6.074) & (8.266) & (9.391) & (8.871) & (4.071) & (6.789) & (5.316) & (9.196) & (4.542) & (6.232) & (0) \\
					Accuracy & \textbf{0.740} & \underline{0.732} & 0.691 & 0.718 & 0.711 & 0.691 & 0.696 & 0.690 & 0.708 & 0.718 & 0.715 \\
					& (0.005) & (0.011) & (0.008) & (0.01) & (0.006) & (0.007) & (0.008) & (0.007) & (0.007) & (0.006) & (0) \\
					$F_1$ score & \textbf{0.690} & \underline{0.682} & 0.578 & 0.648 & 0.634 & 0.575 & 0.588 & 0.573 & 0.623 & 0.652 & 0.663 \\
					& (0.005) & (0.019) & (0.024) & (0.029) & (0.013) & (0.018) & (0.020) & (0.018) & (0.012) & (0.008) & (0) \\
					Strong regularity & 0.649 & 0.983 & 0.986 & 0.984 & 0.624 & 0.981 & 0.981 & 0.984 & 0.450 & \textbf{1.000} & \underline{0.999} \\
					& (0.175) & (0.006) & (0.013) & (0.008) & (0.120) & (0.015) & (0.013) & (0.011) & (0.185) & (0.000) & (0) \\
					Weak regularity & 0.718 & \underline{0.999} & 0.995 & 0.998 & 0.657 & \underline{0.999} & 0.997 & 0.996 & \textbf{1.000} & \textbf{1.000} & \textbf{1.000} \\
					& (0.165) & (0.002) & (0.007) & (0.003) & (0.125) & (0.001) & (0.003) & (0.004) & (0.000) & (0.000) & (0) \\
					\midrule
					\multicolumn{12}{l}{Panel 2: LTDS data, sum-XGR} \\
					\midrule
					Log-likelihood & $\mathbf{-506.5}$ & $-542.9$ & $\underline{-517.8}$ & $-560.6$ & $-548.0$ & $-553.0$ & $-565.9$ & $-553.5$ & $-562.1$ & $-541.0$ & $-570.2$ \\
					& (4.002) & (11.91) & (6.636) & (13.342) & (3.359) & (3.389) & (5.310) & (3.395) & (10.07) & (3.794) & (0) \\
					Accuracy & \textbf{0.732} & 0.707 & \underline{0.726} & 0.693 & 0.705 & 0.703 & 0.691 & 0.702 & 0.703 & 0.720 & 0.709 \\
					& (0.005) & (0.010) & (0.006) & (0.02) & (0.005) & (0.006) & (0.007) & (0.005) & (0.007) & (0.007) & (0) \\
					$F_1$ score & \textbf{0.727} & 0.694 & \underline{0.720} & 0.673 & 0.691 & 0.689 & 0.672 & 0.688 & 0.690 & 0.714 & 0.701 \\
					& (0.005) & (0.014) & (0.007) & (0.032) & (0.006) & (0.007) & (0.008) & (0.006) & (0.007) & (0.007) & (0) \\
					Strong regularity & 0.898 & \underline{0.999} & 0.991 & \underline{0.999} & 0.946 & 0.968 & 0.998 & 0.970 & 0.931 & 0.995 & \textbf{1.000} \\
					& (0.072) & (0.002) & (0.015) & (0.003) & (0.023) & (0.013) & (0.002) & (0.009) & (0.047) & (0.003) & (0) \\
					Weak regularity & 0.906 & \textbf{1.000} & 0.994 & \underline{0.999} & 0.950 & 0.988 & \underline{0.999} & 0.986 & \textbf{1.000} & \textbf{1.000} & \textbf{1.000} \\
					& (0.072) & (0.001) & (0.012) & (0.003) & (0.021) & (0.011) & (0.001) & (0.01) & (0.000) & (0.000) & (0) \\
					\bottomrule
				\end{tabular}
			}
		\end{table}
		
		\begin{table}[!htb]
			\centering
			\caption{Model performance in the validation sets of 1K-Random.}
			\label{tab:1k_random_val}
			\resizebox{.97\linewidth}{!}{
				\begin{tabular}{l|cccc|cccccc|c}
					\toprule
					Metric: & \multicolumn{4}{c|}{DNN} & \multicolumn{6}{c|}{TasteNet} & RUM \\
					Mean (SD) & No GR & PGR & UGR & LGR & No GR & PGR & UGR & LGR & ReLU & Exp & MNL \\
					\midrule
					\multicolumn{12}{l}{Panel 1: CMAP data, sum-XGR} \\
					\midrule
					Log-likelihood & $-151.9$ & $\mathbf{-147.1}$ & $-149.2$ & $\underline{-147.6}$ & $-159.2$ & $-155.3$ & $-155.5$ & $-155.6$ & $-157.8$ & $-160.4$ & $-163.0$ \\
					& (1.033) & (1.993) & (1.876) & (2.548) & (1.725) & (2.055) & (1.920) & (2.188) & (2.613) & (2.203) & (0) \\
					Accuracy & 0.680 & 0.692 & 0.677 & 0.686 & 0.680 & 0.686 & 0.682 & 0.684 & 0.692 & \underline{0.697} & \textbf{0.700} \\
					& (0.006) & (0.008) & (0.008) & (0.008) & (0.011) & (0.012) & (0.012) & (0.014) & (0.010) & (0.012) & (0) \\
					$F_1$ score & 0.628 & \underline{0.638} & 0.561 & 0.610 & 0.599 & 0.574 & 0.579 & 0.571 & 0.606 & 0.637 & \textbf{0.654} \\
					& (0.006) & (0.018) & (0.016) & (0.024) & (0.019) & (0.027) & (0.026) & (0.030) & (0.017) & (0.017) & (0) \\
					Strong regularity & 0.671 & 0.984 & 0.990 & 0.988 & 0.692 & 0.982 & 0.980 & 0.985 & 0.468 & \underline{0.994} & \textbf{0.995} \\
					& (0.169) & (0.009) & (0.015) & (0.015) & (0.119) & (0.018) & (0.010) & (0.012) & (0.208) & (0.003) & (0) \\
					Weak regularity & 0.746 & \textbf{1.000} & 0.996 & \underline{0.999} & 0.721 & 0.997 & 0.996 & 0.996 & \textbf{1.000} & \textbf{1.000} & \textbf{1.000} \\
					& (0.154) & (0.002) & (0.007) & (0.003) & (0.119) & (0.003) & (0.005) & (0.004) & (0.000) & (0.000) & (0) \\
					\midrule
					\multicolumn{12}{l}{Panel 2: LTDS data, sum-XGR} \\
					\midrule
					Log-likelihood & $-143.3$ & $\underline{-138.9}$ & $-139.2$ & $-140.5$ & $-144.5$ & $-141.7$ & $-143.4$ & $-141.9$ & $-145.3$ & $-139.1$ & $\mathbf{-138.7}$ \\
					& (2.016) & (1.676) & (1.338) & (3.175) & (0.741) & (0.958) & (1.105) & (1.085) & (2.642) & (1.596) & (0) \\
					Accuracy & 0.688 & 0.681 & 0.693 & 0.680 & 0.697 & \underline{0.705} & 0.698 & \underline{0.705} & 0.694 & \textbf{0.727} & 0.700 \\
					& (0.008) & (0.015) & (0.015) & (0.019) & (0.009) & (0.009) & (0.010) & (0.009) & (0.013) & (0.014) & (0) \\
					$F_1$ score & 0.683 & 0.663 & 0.685 & 0.653 & 0.676 & 0.684 & 0.676 & 0.684 & 0.674 & \textbf{0.719} & \underline{0.689} \\
					& (0.009) & (0.019) & (0.015) & (0.031) & (0.009) & (0.010) & (0.011) & (0.01) & (0.013) & (0.014) & (0) \\
					Strong regularity & 0.890 & \underline{0.998} & 0.988 & 0.997 & 0.949 & 0.970 & \underline{0.998} & 0.972 & 0.937 & \textbf{1.000} & \textbf{1.000} \\
					& (0.073) & (0.003) & (0.021) & (0.007) & (0.027) & (0.015) & (0.003) & (0.007) & (0.044) & (0.000) & (0) \\
					Weak regularity & 0.896 & \underline{0.999} & 0.992 & 0.997 & 0.951 & 0.984 & 0.998 & 0.982 & \textbf{1.000} & \textbf{1.000} & \textbf{1.000} \\
					& (0.073) & (0.002) & (0.016) & (0.006) & (0.025) & (0.012) & (0.003) & (0.012) & (0.000) & (0.000) & (0) \\
					\bottomrule
				\end{tabular}
			}
		\end{table}
		
		\subsection{Out-of-domain generalization}
		\label{sec:appendix_c.3}
		See \cref{tab:10k_sorted_train,tab:10k_sorted_val}.
		
		\begin{table}[!htb]
			\caption{Model performance in the training sets of 10K-Sorted.}
			\label{tab:10k_sorted_train}
			\resizebox{\linewidth}{!}{
				\begin{tabular}{l|cccc|cccccc|c}
					\toprule
					Metric: & \multicolumn{4}{c|}{DNN} & \multicolumn{6}{c|}{TasteNet} & RUM \\
					Mean (SD) & No GR & PGR & UGR & LGR & No GR & PGR & UGR & LGR & ReLU & Exp & MNL \\
					\midrule
					\multicolumn{12}{l}{Panel 1: CMAP data, sum-XGR} \\
					\midrule
					Log-likelihood & $\mathbf{-4292.7}$ & $\underline{-4325.0}$ & $-4398.3$ & $-4350.1$ & $-4847.9$ & $-4600.0$ & $-4595.5$ & $-4590.9$ & $-4664.2$ & $-4572.7$ & $-4720.0$ \\
					& (21.52) & (17.68) & (13.50) & (20.77) & (28.64) & (9.302) & (9.634) & (11.26) & (60.61) & (8.131) & (0) \\
					Accuracy & \textbf{0.745} & \underline{0.742} & 0.738 & 0.741 & 0.715 & 0.726 & 0.727 & 0.727 & 0.723 & 0.730 & 0.726 \\
					& (0.002) & (0.002) & (0.002) & (0.002) & (0.001) & (0.001) & (0.001) & (0.001) & (0.003) & (0.001) & (0) \\
					$F_1$ score & \textbf{0.717} & \underline{0.714} & 0.707 & 0.713 & 0.661 & 0.691 & 0.692 & 0.692 & 0.684 & 0.698 & 0.693 \\
					& (0.003) & (0.003) & (0.002) & (0.003) & (0.002) & (0.002) & (0.002) & (0.001) & (0.006) & (0.001) & (0) \\
					Strong regularity & 0.881 & 0.998 & 0.998 & 0.998 & \textbf{1.000} & \textbf{1.000} & \underline{0.999} & \underline{0.999} & 0.952 & \textbf{1.000} & \textbf{1.000} \\
					& (0.045) & (0.001) & (0.001) & (0.001) & (0.000) & (0.001) & (0.001) & (0.001) & (0.141) & (0.000) & (0) \\
					Weak regularity & 0.898 & \textbf{1.000} & \textbf{1.000} & \textbf{1.000} & \textbf{1.000} & \textbf{1.000} & \textbf{1.000} & \underline{0.999} & \textbf{1.000} & \textbf{1.000} & \textbf{1.000} \\
					& (0.042) & (0.000) & (0.000) & (0.000) & (0.000) & (0.001) & (0.001) & (0.001) & (0.000) & (0.000) & (0) \\
					\midrule
					\multicolumn{12}{l}{Panel 2: LTDS data, sum-XGR} \\
					\midrule
					Log-likelihood & $\underline{-4554.5}$ & $-4589.4$ & $-4806.0$ & $\mathbf{-4530.7}$ & $-4661.6$ & $-4785.8$ & $-4724.6$ & $-4676.5$ & $-4752.8$ & $-4663.7$ & $-4919.2$ \\
					& (11.10) & (25.82) & (71.16) & (6.477) & (7.997) & (41.15) & (21.08) & (8.192) & (75.76) & (8.416) & (0) \\
					Accuracy & \textbf{0.725} & 0.722 & 0.710 & \textbf{0.725} & \underline{0.724} & 0.712 & 0.717 & 0.721 & 0.716 & \underline{0.724} & 0.712 \\
					& (0.002) & (0.002) & (0.005) & (0.002) & (0.001) & (0.005) & (0.003) & (0.001) & (0.004) & (0.001) & (0) \\
					$F_1$ score & \textbf{0.722} & 0.719 & 0.706 & \textbf{0.722} & \underline{0.720} & 0.706 & 0.713 & 0.717 & 0.712 & \underline{0.720} & 0.708 \\
					& (0.002) & (0.002) & (0.006) & (0.002) & (0.001) & (0.006) & (0.003) & (0.001) & (0.005) & (0.001) & (0) \\
					Strong regularity & 0.736 & \textbf{0.999} & 0.991 & 0.991 & 0.645 & 0.991 & 0.991 & 0.968 & 0.430 & \underline{0.998} & 0.997 \\
					& (0.06) & (0.001) & (0.019) & (0.005) & (0.07) & (0.008) & (0.006) & (0.023) & (0.172) & (0.000) & (0) \\
					Weak regularity & 0.751 & \underline{0.999} & 0.993 & 0.994 & 0.676 & \textbf{1.000} & \underline{0.999} & 0.989 & \textbf{1.000} & \textbf{1.000} & \textbf{1.000} \\
					& (0.054) & (0.0) & (0.016) & (0.002) & (0.067) & (0.000) & (0.001) & (0.005) & (0.000) & (0.000) & (0) \\
					\bottomrule
				\end{tabular}
			}
		\end{table}
		
		\begin{table}[!htb]
			\caption{Model performance in the validation sets of 10K-Sorted.}
			\label{tab:10k_sorted_val}
			\resizebox{\linewidth}{!}{
				\begin{tabular}{l|cccc|cccccc|c}
					\toprule
					Metric: & \multicolumn{4}{c|}{DNN} & \multicolumn{6}{c|}{TasteNet} & RUM \\
					Mean (SD) & No GR & PGR & UGR & LGR & No GR & PGR & UGR & LGR & ReLU & Exp & MNL \\
					\midrule
					\multicolumn{12}{l}{Panel 1: CMAP data, sum-XGR} \\
					\midrule
					Log-likelihood & $-656.6$ & $\underline{-647.9}$ & $-648.5$ & $\mathbf{-646.9}$ & $-683.0$ & $-660.1$ & $-660.0$ & $-660.2$ & $-665.1$ & $-671.2$ & $-673.2$ \\
					& (4.636) & (3.430) & (2.176) & (2.397) & (3.344) & (1.171) & (1.316) & (1.320) & (5.144) & (1.098) & (0) \\
					Accuracy & 0.737 & 0.739 & 0.737 & 0.739 & 0.729 & \underline{0.745} & \textbf{0.746} & \underline{0.745} & 0.744 & 0.741 & \underline{0.745} \\
					& (0.003) & (0.002) & (0.004) & (0.003) & (0.004) & (0.003) & (0.002) & (0.002) & (0.003) & (0.004) & (0) \\
					$F_1$ score & 0.703 & 0.707 & 0.704 & 0.707 & 0.675 & 0.709 & \underline{0.710} & 0.709 & 0.704 & 0.707 & \textbf{0.711} \\
					& (0.004) & (0.002) & (0.005) & (0.005) & (0.005) & (0.004) & (0.003) & (0.003) & (0.004) & (0.005) & (0) \\
					Strong regularity & 0.883 & 0.998 & 0.998 & 0.998 & \textbf{1.000} & \textbf{1.000} & \textbf{1.000} & \textbf{1.000} & 0.952 & \underline{0.999} & \textbf{1.000} \\
					& (0.045) & (0.001) & (0.001) & (0.001) & (0.000) & (0.001) & (0.001) & (0.001) & (0.142) & (0.000) & (0) \\
					Weak regularity & 0.898 & \underline{0.999} & \textbf{1.000} & \textbf{1.000} & \textbf{1.000} & \textbf{1.000} & \textbf{1.000} & \textbf{1.000} & \textbf{1.000} & \textbf{1.000} & \textbf{1.000} \\
					& (0.042) & (0.001) & (0.000) & (0.000) & (0.000) & (0.001) & (0.001) & (0.001) & (0.000) & (0.000) & (0) \\
					\midrule
					\multicolumn{12}{l}{Panel 2: LTDS data, sum-XGR} \\
					\midrule
					Log-likelihood & $\underline{-698.5}$ & $-699.7$ & $-713.6$ & $\mathbf{-696.3}$ & $-720.3$ & $-726.6$ & $-722.7$ & $-720.4$ & $-721.9$ & $-719.4$ & $-730.2$ \\
					& (2.201) & (2.208) & (4.791) & (2.877) & (2.160) & (4.988) & (3.051) & (2.274) & (2.608) & (1.861) & (0) \\
					Accuracy & 0.691 & 0.693 & 0.691 & 0.694 & \textbf{0.701} & 0.686 & 0.694 & \underline{0.699} & 0.698 & 0.697 & 0.687 \\
					& (0.004) & (0.005) & (0.004) & (0.004) & (0.003) & (0.008) & (0.005) & (0.003) & (0.004) & (0.002) & (0) \\
					F1 score & 0.688 & 0.690 & 0.688 & 0.692 & \textbf{0.697} & 0.680 & 0.689 & \underline{0.695} & 0.693 & 0.693 & 0.683 \\
					& (0.004) & (0.005) & (0.004) & (0.004) & (0.003) & (0.009) & (0.005) & (0.003) & (0.004) & (0.002) & (0) \\
					Strong regularity & 0.724 & \textbf{0.998} & 0.991 & 0.990 & 0.637 & 0.992 & 0.992 & 0.967 & 0.420 & \underline{0.997} & 0.996 \\
					& (0.062) & (0.001) & (0.020) & (0.004) & (0.070) & (0.009) & (0.006) & (0.026) & (0.168) & (0.001) & (0) \\
					Weak regularity & 0.739 & \underline{0.999} & 0.992 & 0.992 & 0.668 & \textbf{1.000} & \underline{0.999} & 0.988 & \textbf{1.000} & \textbf{1.000} & \textbf{1.000} \\
					& (0.054) & (0.001) & (0.018) & (0.003) & (0.066) & (0.001) & (0.001) & (0.007) & (0.000) & (0.000) & (0) \\
					\bottomrule
				\end{tabular}
			}
		\end{table}
		
		\clearpage
		\section{Effects of regularization strength}
		\label{sec:appendix_d}
		
\begin{figure}[!htb]
			\centering
			\begin{subfigure}{.245\linewidth}
				\includegraphics[width=\linewidth]{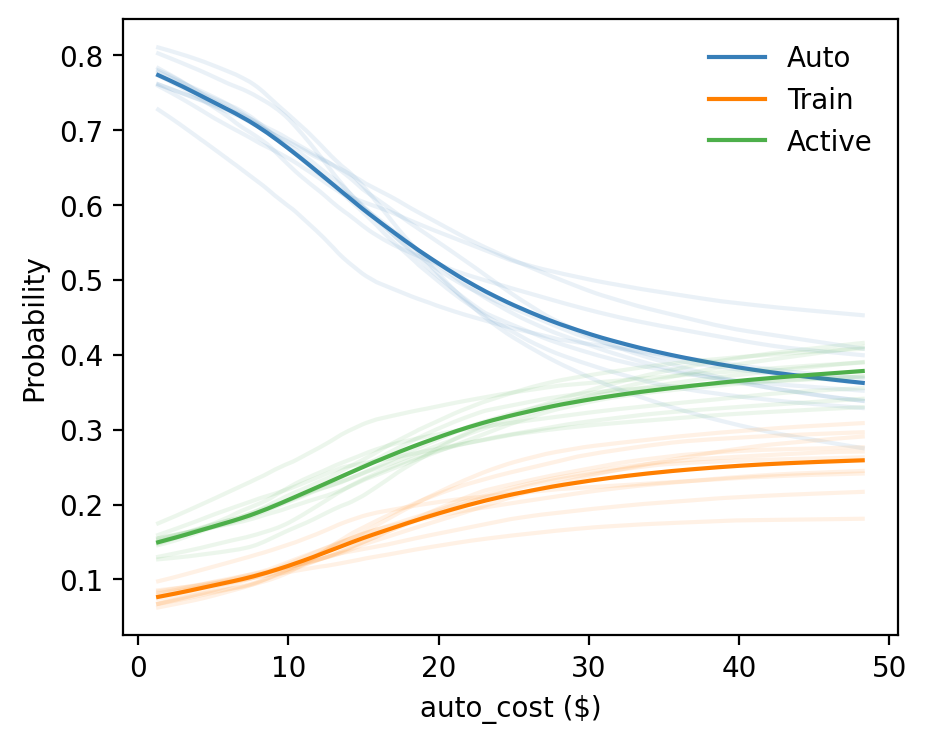}
				\subcaption{$\lambda=100$, sum-PGR}
			\end{subfigure}
			\begin{subfigure}{.245\linewidth}
				\includegraphics[width=\linewidth]{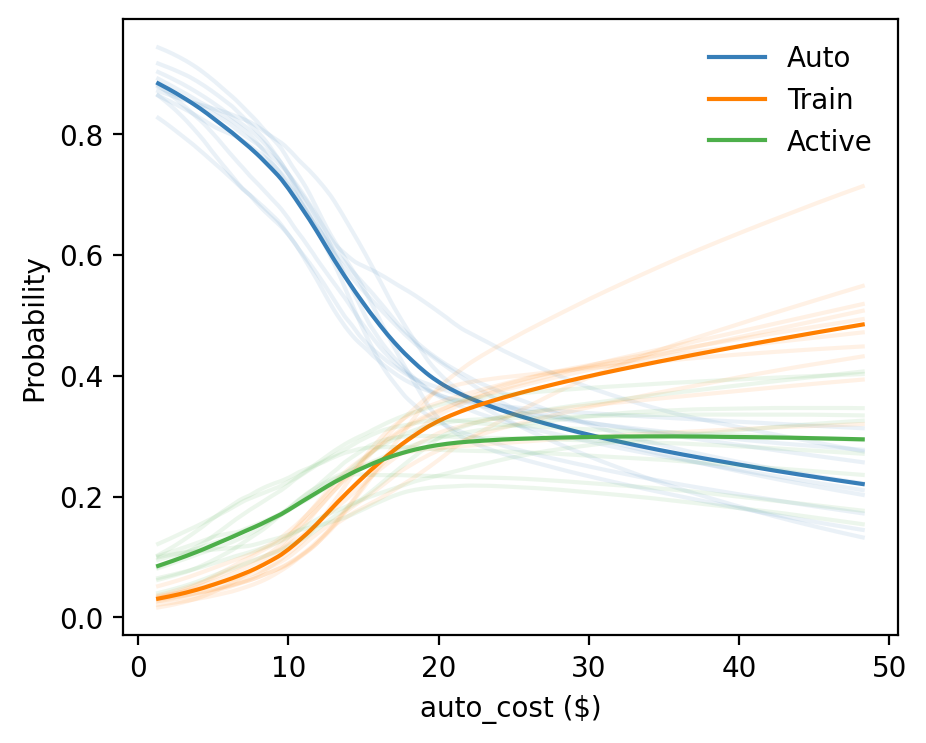}
				\subcaption{$\lambda=1$, sum-PGR}
			\end{subfigure}
			\begin{subfigure}{.245\linewidth}
				\includegraphics[width=\linewidth]{fig/DNN_sum_PGR_0.01_10K.png}
				\subcaption{$\lambda=10^{-2}$, sum-PGR}
			\end{subfigure}
			\begin{subfigure}{.245\linewidth}
				\includegraphics[width=\linewidth]{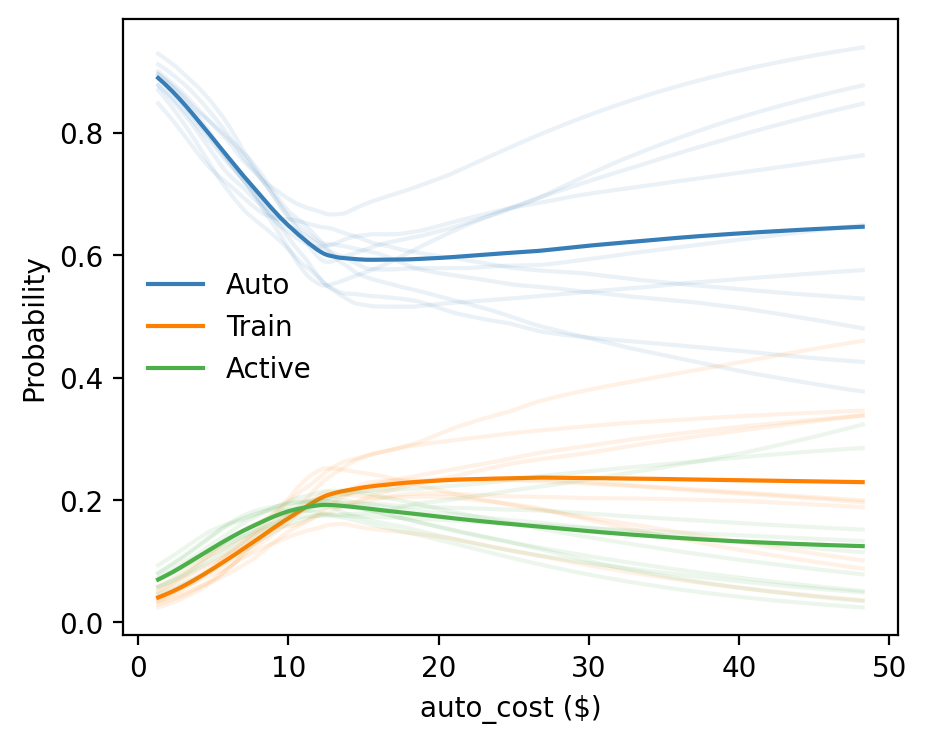}
				\subcaption{$\lambda=10^{-4}$, sum-PGR}
			\end{subfigure}
			\par\smallskip
			\begin{subfigure}{.245\linewidth}
				\includegraphics[width=\linewidth]{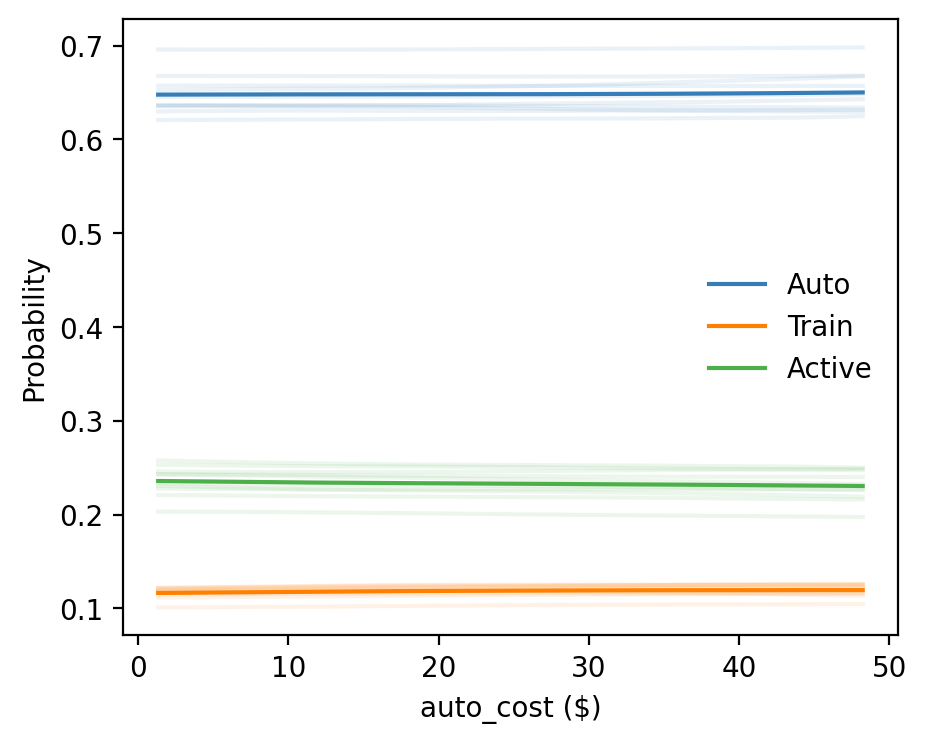}
				\subcaption{$\lambda=100$, norm-PGR}
			\end{subfigure}
			\begin{subfigure}{.245\linewidth}
				\includegraphics[width=\linewidth]{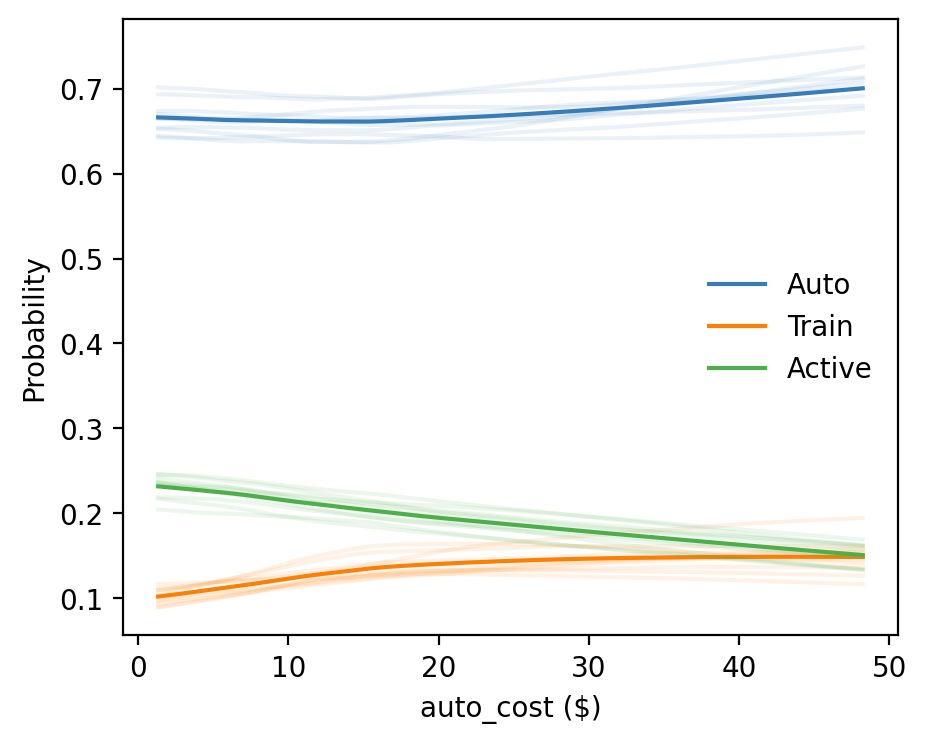}
				\subcaption{$\lambda=1$, norm-PGR}
			\end{subfigure}
			\begin{subfigure}{.245\linewidth}
				\includegraphics[width=\linewidth]{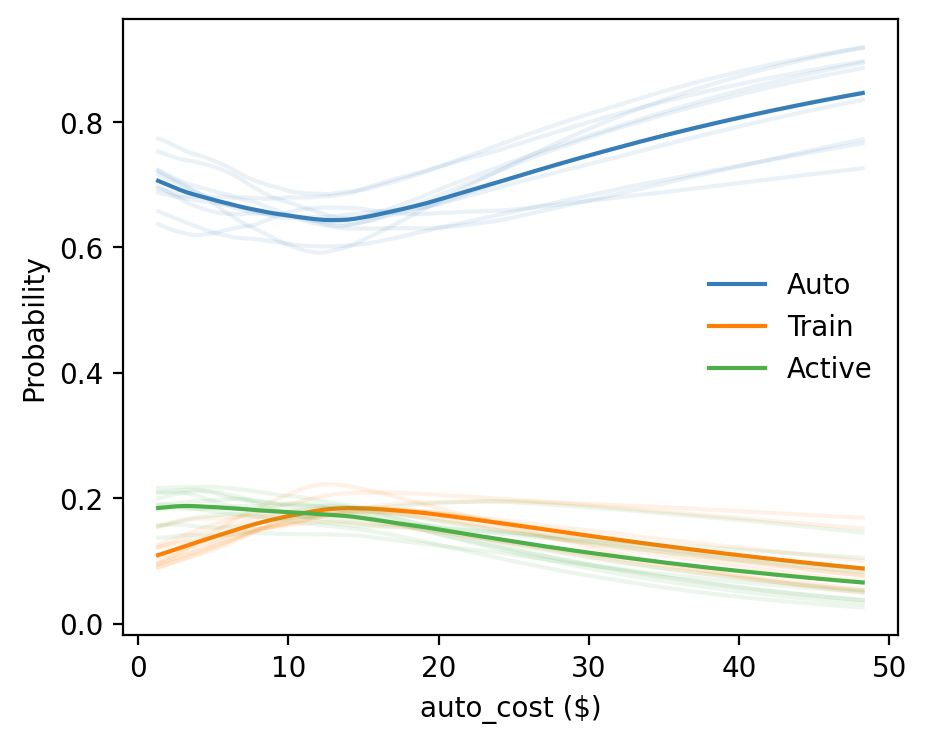}
				\subcaption{$\lambda=10^{-2}$, norm-PGR}
			\end{subfigure}
			\begin{subfigure}{.245\linewidth}
				\includegraphics[width=\linewidth]{fig/DNN_norm_PGR_0.0001_10K.png}
				\subcaption{$\lambda=10^{-4}$, norm-PGR}
			\end{subfigure}
			\caption{Individual demands as functions of driving costs (10K-Random, CMAP).}
			\label{fig:sub_lambda}
		\end{figure}

		\cref{fig:sub_lambda} visualizes the effects of regularization strength $\lambda$ on individual demand functions.
		
		\section{Sensitivity test of the behavioral regularity metric}
		\label{sec:appendix_e}
		
\begin{figure}[!htb]
			\centering
			\begin{subfigure}{.325\linewidth}
				\includegraphics[width=\linewidth]{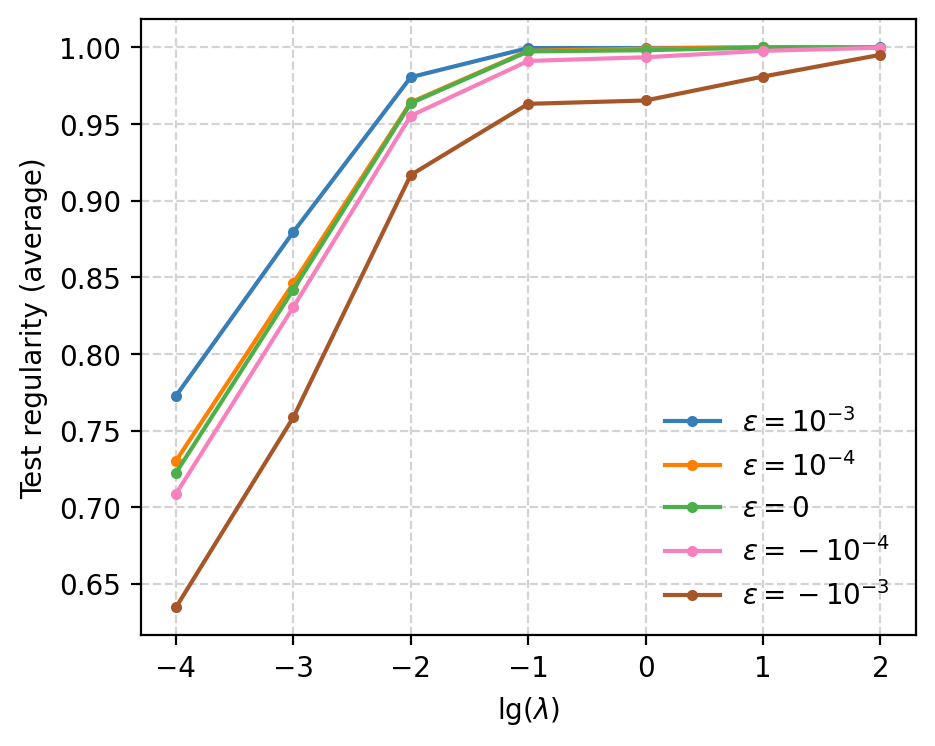}
				\subcaption{DNN}
			\end{subfigure}
			\begin{subfigure}{.325\linewidth}
				\includegraphics[width=\linewidth]{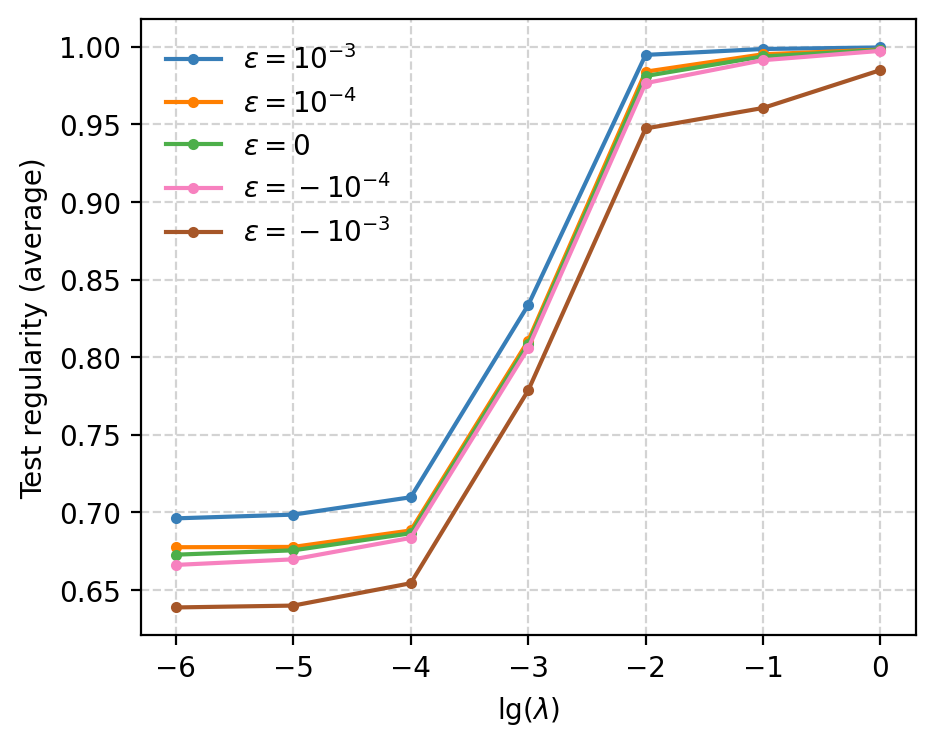}
				\subcaption{TasteNet}
			\end{subfigure}
			\caption{Behavioral regularity as a function of $\varepsilon$ (1K-Random, CMAP).}
			\label{fig:sensitivity_test}
		\end{figure}

		The trend of behavioral regularity remains stable when $\varepsilon$ changes. As shown in \cref{fig:sensitivity_test}, a sensitivity test shows that the regularity metric increases with $\varepsilon$. This is because a smaller $\varepsilon$ represents a tighter upper bound of being behaviorally regular, which leads to lower behavioral regularity. The trend of the metric with $\lambda$ is generally not affected by the value of $\varepsilon$, unless we choose a negative number with a relatively large absolute value, indicating that the modeler's assumption on behavioral regularity is too strong. Since the metric trend is of more concern, we can manually set suitable $\varepsilon$'s to distinguish between strong and weak regularities.
		
	\end{appendices}
	
	\bibliography{bibfile.bib}
\end{document}